\documentclass[10pt,twocolumn,letterpaper]{article}

\usepackage{cvpr} 

\usepackage[utf8]{inputenc}
\usepackage{graphicx}
\usepackage{booktabs}
\usepackage{multirow}
\usepackage{float}
\usepackage{adjustbox}
\usepackage{stfloats}
\usepackage{amssymb}
\usepackage{caption}

\usepackage{colortbl}
\usepackage[accsupp]{axessibility}
\usepackage{tikz}
\usepackage{hyperref}

\def\checkmark{\tikz\fill[scale=0.4]
(0,.35) -- (.25,0) -- (1,.7) -- (.25,.15) -- cycle;}

\usepackage{adjustbox}
\usepackage{stfloats}
\usepackage[most]{tcolorbox}  
\usepackage{xcolor} 
\usepackage{graphicx} 
\usepackage{amssymb}
\usepackage{caption} 
\usepackage[utf8]{inputenc}
\usepackage{multirow}
\usepackage{listings}
\usepackage{tikz}
\def\checkmark{\tikz\fill[scale=0.4](0,.35) -- (.25,0) -- (1,.7) -- (.25,.15) -- cycle;} 

\usepackage{subcaption}

\definecolor{cvprblue}{rgb}{0.21,0.49,0.74}

\def\paperID{9365}
\def\confName{CVPR}
\def\confYear{2026}
\title{\textit{UDVideoQA}: A Traffic Video Question Answering Dataset for Multi-Object Spatio-Temporal Reasoning in Urban Dynamics}

\author{Joseph Raj Vishal \quad Nagasiri Poluri \quad 
Katha Naik \quad 
  Rutuja Patil \quad 
 Kashyap Hegde Kota \\ Krishna Vinod  \quad Prithvi Jai Ramesh \quad  Mohammad Farhadi \quad  Yezhou Yang \quad  Bharatesh Chakravarthi
\\
Arizona State University\\
{\tt\small \{jnolas77, mfarhadi, yz.yang, bshettah\}@asu.edu}
}

\newcommand{\mywatermark}{%
    \begin{minipage}{\textwidth}
        \centering
        \fontsize{10}{10}\selectfont 
       This paper has been accepted for publication at the IEEE Conference on Computer Vision and Pattern Recognition \\ (CVPR) Findings, Denver, 2026.
    \end{minipage}%
}
\usepackage{background}
\backgroundsetup{
    scale=1,
    color=gray,
    angle=0,
    opacity=0.5,
    position=current page.north,
    vshift=-1cm, 
    contents={\mywatermark}
}

\begin{document}

\twocolumn[{
    \renewcommand\twocolumn[1][]{#1}%
    \maketitle
    \vspace{-0.8cm}

    \begin{center}
        \includegraphics[width=0.88\linewidth]{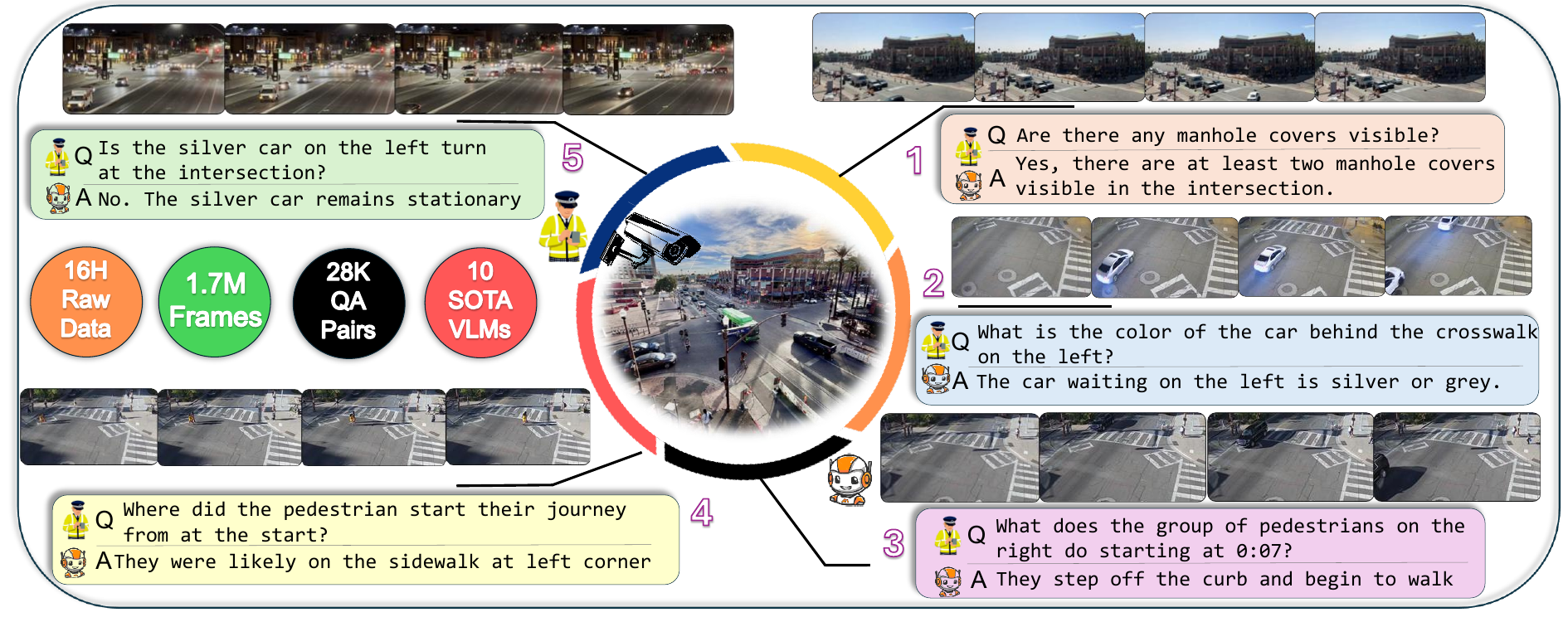}
        \vspace{0.3cm}

        \captionof{figure}{
        The \textit{UDVideoQA} benchmark challenges a \textit{VideoQA} model to analyze a busy intersection using \textit{question-answer} pairs across diverse urban traffic conditions. The model's capabilities are demonstrated across five reasoning categories: attribution, basic understanding, event reasoning, reverse reasoning, and counterfactual inference.
        }
        \label{fig:teaser}
    \end{center}

    \vspace{0.1cm}
}]

\begin{abstract}

Understanding the complex, multi-agent dynamics of urban traffic remains a fundamental challenge for video-language models (VideoLMs). This paper introduces \textbf{U}rban \textbf{D}ynamics VideoQA (UDVideoQA), a benchmark dataset that captures the unscripted real-world behavior of dynamic urban scenes. UDVideoQA is curated from $16$ hours ($1.7$M frames) of traffic footage recorded at multiple city intersections under diverse traffic, weather, and lighting conditions. It employs an event-driven dynamic blur technique to ensure privacy preservation without compromising scene fidelity. Using a unified annotation pipeline, the dataset contains $\sim$ $28,000$ question–answer pairs generated across $8$ hours of densely annotated video, averaging one question per second. Its taxonomy follows a hierarchical reasoning level, spanning basic understanding and attribution to event reasoning, reverse reasoning, and counterfactual inference, enabling systematic evaluation of both visual grounding and causal reasoning. Comprehensive experiments benchmark $10$ state-of-the-art VideoLMs on UDVideoQA and $8$  models on a complementary video question generation (VideoQGen) benchmark. Results reveal a persistent perception-reasoning gap, showing models that excel in abstract inference often fail with fundamental visual grounding. While models like Gemini Pro achieve the highest zero-shot accuracy, fine-tuning the smaller Qwen$2.5$-VL 7B model on UDVideoQA bridges this gap, achieving performance comparable to proprietary systems. In VideoQGen, Gemini 2.5 Pro, and Qwen3 Max generate the most relevant and complex questions, though all models exhibit limited linguistic diversity, underscoring the need for human-centric evaluation. The UDVideoQA suite, including the dataset, annotation tools, and benchmarks for both VideoQA and VideoQGen, provides a foundation for advancing robust, privacy-aware, and real-world multimodal reasoning. UDVideoQA is available at \url{https://ud-videoqa.github.io/UD-VideoQA/UD-VideoQA/}.

\end{abstract}
 \section{Introduction}
\label{sec:intro}

Urban traffic scenes are inherently dynamic, characterized by simultaneous interactions among multiple agents, including vehicles, pedestrians, micro-mobility, and their surroundings. This complexity results in unpredictable motion patterns and dense visual dependencies, posing major challenges for robust visual-language reasoning, particularly in domains such as automated surveillance and intelligent transportation systems.
With the rise of edge-based AI systems and continuous visual monitoring, there is a growing demand for \textit{VideoLMs} that can understand long-term, untrimmed video streams. Achieving such understanding requires spatio-temporally dense datasets that reflect real-world conditions rather than short, curated clips. Despite the proliferation of urban surveillance infrastructure recording thousands of streams daily, the data remains largely unexploited and uncurated. As a result, a critical gap persists between the abundance of data and the needs of \textit{VideoLM} benchmarking, as most existing datasets rely on web-sourced or simulation-based videos that lack temporal continuity and scene realism~\cite{Zhao2021hacs}.

To bridge this gap,  \textit{UDVideoQA} is introduced as a large-scale, open-source \textbf{\textit{U}}rban \textbf{\textit{D}}ynamics \textbf{\textit{V}}ideo \textbf{\textit{Q}}uestion \textbf{\textit{A}}nswering benchmark for evaluating multimodal reasoning in complex urban scenes. \textit{UDVideoQA} provides an end-to-end solution for data curation, annotation, and benchmarking, encompassing $16$ hours of surveillance footage ($1.7M$ frames) and $28,800$ question-answer (\textit{QA}) pairs over $8$ hours of densely annotated video as summarized in Figure~\ref {fig:teaser}. The dataset spans a hierarchy of reasoning skills, from basic perception and attribute recognition to causal and counterfactual inference, which remain challenging even for state-of-the-art (\textit{SOTA}) models.

\textit{UDVideoQA} introduces key contributions that set it apart from existing \textit{VideoQA} datasets.
First, a unified annotation framework integrates a novel event-driven dynamic blurring mechanism that generates object-level motion masks for privacy-preserving processes. This approach maintains contextual integrity while safeguarding individual privacy, advancing beyond traditional detector-based redaction.
Second, a semi-automated annotation tool supports scalable generation and validation of free-form \textit{QA} pairs, moving beyond the constrained, template-based formats common in prior benchmarks~\cite{Marcu2024LingoQA, Zhou2025TUMTrafficVideoQA}.
Finally, to complement standard \textit{VideoQA} evaluation, \textit{UDVideoQA} introduces the \textit{first-of-its-kind} benchmark for video question generation (\textit{VideoQGen}) in the urban traffic domain, designed to evaluate linguistic diversity, contextual grounding, and reasoning generalization. All resources, including the dataset, annotation toolkit, and both \textit{VideoQA} and \textit{VideoQGen} benchmarks, are released as part of an open, institutional reviewer board (IRB) approved work to promote ethically aligned research in real-world multimodal reasoning.
The key contributions of this work are summarized as follows.

\begin{itemize}
\item \textbf{A highly dense, large-scale dataset:} $16$ hours of urban traffic videos ($\sim$ $1.7M$ frames) with $28,800$ \textit{QA} pairs at a density of $3.6$K \textit{QA/hour} across diverse weather, lighting, and traffic conditions.
\item \textbf{A scalable and ethical annotation toolkit:} A privacy-preserving event-driven framework compatible with any traffic dataset~\cite{Xu2021SUTDTrafficQA,RoadSocial2025}, incorporating motion-based blurring, human validation, and model-ready templates.
\item \textbf{Comprehensive benchmarking:} Evaluation of $10$ \textit{SOTA} \textit{VideoLMs} for \textit{VideoQA} and $8$ models for \textit{VideoQGen}, revealing consistent perception–reasoning misalignment.
\item \textbf{Empirical insight:} Smaller models demonstrated superior performance in low-level visual tasks (e.g., \textit{color} and \textit{attribute recognition}) compared to larger models, highlighting challenges in large-scale reasoning architectures.
\end{itemize}

The remainder of this paper is organized as follows. Section~\ref{sec:related_works} reviews related work in traffic \textit{VideoQA} and multimodal reasoning, Section~\ref{dataset} presents the \textit{UDVideoQA} dataset, including data collection, privacy methods, and \textit{QA} taxonomy, Section~\ref{sec:experiments} details the experimental setup and benchmarking results, and Section~\ref{sec:conclusion} concludes with key findings and future directions.

\begin{table*}[!t]
\centering

\resizebox{0.85\textwidth}{!}{%
\begin{tabular}{@{}lccccccccc@{}}
\toprule 
\textbf{Dataset} &
  \textbf{Year} &
  \textbf{Domain} &
  \textbf{Video Frames} &
  \textbf{QA density} &
  \textbf{Curated/Internet Sourced} &
  \textbf{Hallucination Robustness} &
  \textbf{Multi-Events} &
  \textbf{Annotation Tool} \\ \midrule
  BDD-OIA~\cite{Rasouli2019BDDOIA} &
  2019 & 
  Driving &
  0.02M &
  - &
  Curated &
  {\color[HTML]{FE0000} ×} &
  {\color[HTML]{FE0000} ×} &
  {\color[HTML]{FE0000} ×}
  \\ \midrule
  BDD-X~\cite{Kim2020BDDX} &
  2020 &
  Driving &
  8.4M &
  0.79 QA/h &
  Curated &
  {\color[HTML]{FE0000} ×} &
  {\color[HTML]{FE0000} ×} &
  {\color[HTML]{FE0000} ×}
 \\ \midrule
  MM-AU~\cite{fang2024abductiveegoviewaccidentvideo} &
  2021 & 
  Driving &
  2.2M &
  2.9k QA/h &
  Internet Sourced &
  {\color[HTML]{FE0000} ×} &
  {\color[HTML]{FE0000} ×} &
  {\color[HTML]{FE0000} ×}
  \\ \midrule
  ROAD~\cite{Singh2021ROAD} &
  2021 &
  Driving &
  0.02M &
  15 QA/h &
  Curated &
  {\color[HTML]{FE0000} ×} &
  {\color[HTML]{FE0000} ×} &
  {\color[HTML]{FE0000} ×} \\ \midrule
 SUTD-TrafficQA~\cite{Xu2021SUTDTrafficQA} &
  2021 &
  Wild Traffic &
  10M &
  560 QAs/h &
  Internet Sourced &
  {\color[HTML]{FE0000} ×} &
  {\color[HTML]{FE0000} ×} &
  {\color[HTML]{FE0000} ×}  \\ \midrule
DRAMA~\cite{Malla2023DRAMA} &
  2023 &
  Driving &
  0.02M &
  10k QAs/h &
  Curated &
  {\color[HTML]{FE0000} ×} &
  {\color[HTML]{FE0000} ×} &
  {\color[HTML]{FE0000} ×} \\ \midrule
Rank2Tell~\cite{Jamaludin2023Rank2Tell} &
  2023 &
  Driving &
  0.02M &
  181k QA/h &
  Curated &
  {\color[HTML]{FE0000} ×} &
  {\color[HTML]{FE0000} ×} &
  {\color[HTML]{FE0000} ×} \\ \midrule
  Drive-LM~\cite{Wang2024DriveLM} &
  2024 &
  Driving &
  0.03M &
  - &
  Curated &
  {\color[HTML]{FE0000} ×} &
  {\color[HTML]{FE0000} ×} &
  {\color[HTML]{FE0000} ×} 
 \\ \midrule
  NuScenes-QA~\cite{Qian2024NuScenesQA} &
  2024 &
  Driving &
  34K &
  13.5 QAs/h &
  Curated &
  {\color[HTML]{FE0000} ×} &
  {\color[HTML]{FE0000} ×} &
  {\color[HTML]{FE0000} ×}
 \\ \midrule
  LingoQA~\cite{Marcu2024LingoQA} &
  2024 &
  Driving &
  0.1M &
  13k QAs/h &
  Curated &
  {\color[HTML]{FE0000} ×} &
  {\color[HTML]{FE0000} ×} &
  {\color[HTML]{FE0000} ×} 
 \\ \midrule
  TUMTraffic-VideoQA~\cite{Zhou2025TUMTrafficVideoQA} &
  2024 &
  Roadside &
  8.9K &
  85 QAs/h &
  Curated &
  {\color[HTML]{FE0000} ×} &
  {\color[HTML]{FE0000} ×} &
  {\color[HTML]{FE0000} ×}
 \\ \midrule
  RoadSocial~\cite{RoadSocial2025} &
  2025 &
  Wild Traffic &
  14M &
  1.98k QAs/h &
  Internet Sourced &
  {\color[HTML]{036400} \checkmark} &
  {\color[HTML]{FE0000} ×} &
  {\color[HTML]{FE0000} ×} 
 \\ \midrule 
  \rowcolor[HTML]{FFFFC7} 
\textbf{UDVideoQA (Ours)} &
  \textbf{2026} &
  \textbf{Roadside} &
 \textbf{1.7M} &
  \textbf{3.6k QAs/h} &
  \textbf{Curated} &
  \textbf{\color[HTML]{036400} \checkmark} &
  \textbf{\color[HTML]{036400} \checkmark} &
  \textbf{\color[HTML]{036400} \checkmark} \\ \bottomrule
\end{tabular}%

}

\caption{Comparison of \textit{UDVideoQA} with existing traffic \textit{VideoQA} datasets. It highlights the key features of various datasets, showcasing \textit{UDVideoQA}'s advancements in data scale, annotation density, and its unique focus on hallucination robustness and multi-event scenarios from a roadside perspective.}
\label{tab:related_works}
\end{table*}

\section{Related Work}
\label{sec:related_works}

Recent progress in video-language understanding has inspired the development of several benchmarks and models targeting complex reasoning across visual and temporal dimensions. This section positions \textit{UDVideoQA} within the urban traffic landscape by reviewing prior efforts in traffic-focused \textit{VideoQA}, examining advances in \textit{VideoLMs}, and highlighting how existing datasets and evaluation practices fall short in capturing the continuous, multi-agent nature of real-world urban traffic scenes.

\subsection{Traffic VideoQA and Urban Reasoning}

Recent traffic \textit{VideoQA} benchmarks have advanced domain-specific reasoning in areas such as causality~\cite{Xu2021SUTDTrafficQA}, risk localization~\cite{Malla2023DRAMA}, and text-aware scene understanding~\cite{Marcu2024LingoQA}. Efforts to scale data collection have similarly expanded, leveraging social media videos and multi-sensor data from autonomous driving platforms~\cite{RoadSocial2025, Qian2024NuScenesQA}.

However, as summarized in Table~\ref{tab:related_works}, existing benchmarks face key limitations. Most are curated and event-centric, capturing discrete driving incidents from a single, vehicle-mounted perspective. While useful for structured perception tasks, this setup sparsely represents the continuous multi-agent (pedestrians and vehicles) interactions found in dense real-world urban intersections. Furthermore, many current datasets do not adequately test for hallucination robustness by asking negative \textit{QA} pairs on objects not present in the scene. This gap highlights the need for benchmarks that capture the true spatio-temporal density and social complexity of traffic environments to support more robust multimodal reasoning.

\subsection{Video-Language Models and Benchmarking}
Recent surveys on \textit{VideoLMs}~\cite{vishal2024eyes} have outlined the evolution of multimodal reasoning architectures, highlighting both scaling success and persistent grounding limitations. In this study, the benchmarking framework includes a diverse mix of proprietary and open-source models to provide a comprehensive evaluation of video-language understanding capabilities.
The proprietary set includes \textit{Gemini} series~\cite{Google2024Gemini, Google2025Gemini} and \textit{GPT} series~\cite{OpenAI2024GPT4o,OpenAI2025GPT5}, While the open-source models cover \textit{Qwen2.5VL}~\cite{Qwen2_2024}, \textit{VideoLLaMA 3}~\cite{zhang2025videollama3frontiermultimodal}, \textit{InternVL3}~\cite{zhu2025internvl3exploringadvancedtraining}, \textit{NVILA}~\cite{Lin2024VILA}, and \textit{LLaVA-NeXT-Video}~\cite{Liu2024LLaVANeXT}.
This selection spans multiple architectures, parameter scales, and training paradigms, offering a balanced assessment of the current \textit{VideoLM} landscape. Despite impressive reasoning capabilities, these models often struggle with fine-grained perception, temporal consistency, and contextual grounding when applied to real-world traffic interactions.

\subsection{Positioning of UDVideoQA}
\begin{figure*}[ht]
  \centering
\includegraphics[width=0.83\linewidth]{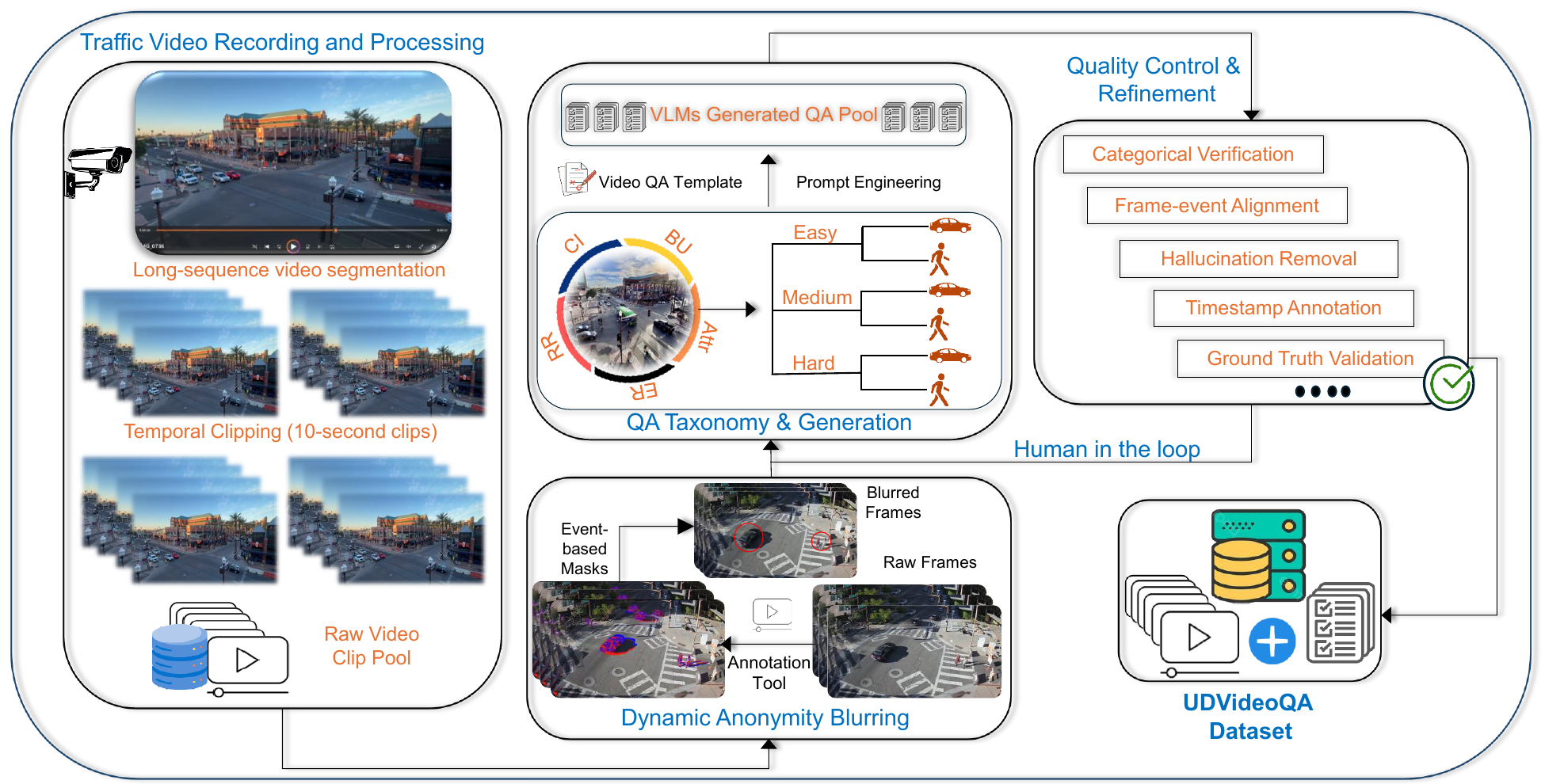}
\caption{Illustrates the pipeline for creating the \textit{UDVideoQA} dataset. The process begins with traffic video recording, which is segmented and temporally clipped into $10$s clips. These clips undergo dynamic anonymity blurring. The \textit{QA} taxonomy and generation module then uses model based on \textit{VideoQGen} benchmark and prompt engineering to create a question pool, which undergoes rigorous \textit{human in the loop} quality control and refinement.}
\label{fig:overview}
\end{figure*}

In response to these limitations, \textit{UDVideoQA} introduces a continuous, surveillance-based dataset designed to capture multi-agent dynamics across varied urban conditions. The accompanying annotation framework supports ethical and scalable data handling through motion-based dynamic blurring and facilitates \textit{QA} generation (see Figure~\ref{fig:overview}), surpassing the rigid, template structures of earlier datasets~\cite{Marcu2024LingoQA}.
As summarized in Table~\ref{tab:related_works}, \textit{UDVideoQA} uniquely incorporates dense multi-event interactions, adverse weather, and nighttime scenarios, while introducing a \textit{VideoQGen} benchmark for assessing linguistic diversity and reasoning robustness.
Together, this establishes \textit{UDVideoQA} as a comprehensive platform for advancing multimodal reasoning in complex, real-world urban traffic scenes.

\section{\textit{UDVideoQA} Dataset}
\label{dataset}

\subsection{Data Collection and Composition}
The \textit{UDVideoQA} dataset comprises $16$ hours of varied traffic footage with over $1.7$ million frames, collected from multiple metropolitan intersections using deployed smart traffic camera systems (details in Supplementary). The cameras captured video at $30$ fps, providing sufficient spatio-temporal resolution to observe complex pedestrian-pedestrian, vehicle-vehicle, and vehicle-pedestrian interactions, as well as their evolving spatial relationships within the scene. To reflect the diversity of real-world conditions, recordings span different times of day, morning rush hours ($7$-$9$ AM), midday ($11$ AM-$1$ PM), evening rush hours ($5$-$7$ PM), and nighttime ($8$ PM-midnight), and encompass a wide range of weather variations, including clear, rainy, and dust-storm conditions. Together, these recordings form a dataset that captures the full temporal, environmental, and behavioral variability inherent to urban traffic dynamics.

\subsection{Data Preprocessing and Anonymity}

\paragraph{\textit{Video Segmentation.}}
The raw footage was segmented into $10$s clips, following prior studies that identify this duration as sufficient to capture complete human and vehicular actions within a coherent temporal context~\cite{Carreira2017Kinetics}. Longer clips tend to introduce redundancy and increase modeling complexity, while shorter clips, less than $10$s, often miss key temporal dependencies~\cite{Yan2024LLaMAVID, Wu2019LongTerm}. For \textit{VideoQA}, adopting $10$s intervals aligns with the \textit{“less is more”} principle~\cite{Lei2021ClipBERT}, enabling denser annotations ($1$s of frame, there is $1$ \textit{QA} pair) and maintaining sufficient temporal information for complex reasoning tasks. This design choice ensures that models focus on concise, context-rich segments that support fine-grained spatio-temporal understanding.
\\
\textbf{\textit{Ethics and Dynamic Blurring.}}
Ethical data handling was prioritized during dataset curation as it possessed personally identifiable information. The project received approval from the university IRB under protocol \textit{STUDY00018234}. To preserve privacy, \textit{UDVideoQA} employs an event-driven dynamic blurring technique that generates motion-based masks for anonymization~\cite{Marziliano2002BlurMetric, Crete2007BlurEffect}. This method operates on the principle of event-based vision \cite{chakravarthi2024recent}, applying localized, time-dependent blurring that protects identities while retaining contextual integrity within each frame.
For benchmarking, it was compared against a detector–segmenter baseline that combines \textit{YOLOv11} for object detection~\cite{Jocher2023YOLOv8} with \textit{SAM2} for segmentation~\cite{Kirillov2024SAM2}. Quantitative and qualitative results (details in Supplementary) demonstrate that the event-driven method provides superior robustness and temporal consistency under low-light or cluttered conditions.

\subsection{QA Taxonomy and Generation Benchmark}

\paragraph{\textit{QA Taxonomy.}}
To systematically evaluate model reasoning in traffic scenarios, \textit{UDVideoQA} defines a hierarchical taxonomy of questions that spans both perceptual and inferential skills~\cite{Xu2021SUTDTrafficQA}. This taxonomy enables structured assessment of model performance across diverse reasoning types, reflecting the layered complexity of urban traffic perception.
Each model is evaluated on its ability to: \textbf{\textit{(1)}} interpret environmental context such as lighting, weather, and background conditions, \textbf{\textit{(2)}} recognize relevant objects and their attributes to establish spatial grounding, \textbf{\textit{(3)}} reason over temporal interactions among multiple agents, and \textbf{\textit{(4)}} enable higher-order inference via causal and counterfactual reasoning while adding out-of-context questions to reduce hallucination. The taxonomy consists of five primary reasoning categories:

\begin{itemize}
    \item \textbf{\textit{Attribution:}} Focuses on basic perception and recognition of objects, attributes, and text. \textit{Example:} \textit{“Is the crosswalk signal visible and displaying a ‘walk’ icon for pedestrians?”}
    
    \item \textbf{\textit{Basic Understanding:}} Evaluates global scene comprehension, such as identifying ongoing activities or summarizing environmental context.
     \textit{Example:} \textit{“What is the lighting condition in the background?”}
    \item \textbf{\textit{Event Reasoning:}} Captures cause-and-effect reasoning over temporal sequences.    \textit{Example:}
    \textit{“Why did the white sedan brake after the blue truck changed lanes?”}
    
    \item \textbf{\textit{Reverse Reasoning:}} Requires inference of prior states that explain the observed event.
   \textit{Example:}
   \textit{“Given the pedestrian is halfway across the street, what was the likely state of the traffic light just before?”}
    
    \item \textbf{\textit{Counterfactual Inference:}} Tests model robustness to hallucination through hypothetical scenarios.
    \textit{Example:} \textit{“If the traffic light had been green, would the motorcycle have cleared the intersection before the pedestrian?”}
\end{itemize}

Each question is further labeled by \textit{difficulty} (easy, medium, hard) and \textit{interaction-focused} (pedestrian-centric or vehicular-centric), resulting in $30$ taxonomic tags. A structured annotation template ensures consistency across annotators, standardizing question format, difficulty calibration, and coverage across reasoning types (details in Supplementary).

\begin{table*}[t!]
\centering
\resizebox{\textwidth}{!}{%
\begin{tabular}{@{}cccccclccccclcccccl|ccccclcccccl|ccccclcccccl@{}}
\toprule
\textbf{Model Name} &
 \multicolumn{6}{c}{\textbf{Relevance}} &
 \multicolumn{6}{c}{\textbf{Answerability}} &
 \multicolumn{6}{c|}{\textbf{Diversity}} &
 \multicolumn{6}{c}{\textbf{Pedestrian Centric}} &
 \multicolumn{6}{c|}{\textbf{Vehicle Centric}} &
 \multicolumn{6}{c}{\textbf{Specific (Grounded)}} &
 \multicolumn{6}{c}{\textbf{Generic}} \\ \midrule
&
 \textbf{BU} &
 \textbf{Atr} &
 \textbf{ER} &
 \textbf{RR} &
 \multicolumn{2}{c|}{\textbf{CI}} &
 \textbf{BU} &
 \textbf{Atr} &
 \textbf{ER} &
 \textbf{RR} &
 \multicolumn{2}{c|}{\textbf{CI}} &
 \textbf{BU} &
 \textbf{Atr} &
 \textbf{ER} &
 \textbf{RR} &
 \multicolumn{2}{c|}{\textbf{CI}} &
 \textbf{BU} &
 \textbf{Atr} &
 \textbf{ER} &
 \textbf{RR} &
 \multicolumn{2}{c|}{\textbf{CI}} &
 \textbf{BU} &
 \textbf{Atr} &
 \textbf{ER} &
 \textbf{RR} &
 \multicolumn{2}{c|}{\textbf{CI}} &
 \textbf{BU} &
 \textbf{Atr} &
 \textbf{ER} &
 \textbf{RR} &
 \multicolumn{2}{c|}{\textbf{CI}} &
 \textbf{BU} &
 \textbf{Atr} &
 \textbf{ER} &
 \textbf{RR} &
 \multicolumn{2}{c}{\textbf{CI}} \\ \midrule
\multicolumn{1}{c|}{\textbf{Gemini 2.5 Pro}} &
 \cellcolor[HTML]{DC143C}92 &
 \cellcolor[HTML]{FD6864}84 &
 \cellcolor[HTML]{FD6864}82 &
 \cellcolor[HTML]{FD6864}86 &
 \multicolumn{2}{c|}{\cellcolor[HTML]{FD6864}80} &
 \cellcolor[HTML]{FD6864}90 &
 \cellcolor[HTML]{FD6864}80 &
 \cellcolor[HTML]{FD6864}80 &
 \cellcolor[HTML]{FD6864}84 &
 \multicolumn{2}{c|}{\cellcolor[HTML]{FD6864}82} &
 \cellcolor[HTML]{FFA500}64 &
 \cellcolor[HTML]{FFA500}62 &
 \cellcolor[HTML]{FFA500}66 &
 \cellcolor[HTML]{FFA500}64 &
 \multicolumn{2}{c|}{\cellcolor[HTML]{FFA500}60} &
 \cellcolor[HTML]{E0FFFF}35 &
 \cellcolor[HTML]{E0FFFF}30 &
 \cellcolor[HTML]{FFFFE0}45 &
 \cellcolor[HTML]{FFFFE0}40 &
 \multicolumn{2}{c|}{\cellcolor[HTML]{FFD700}50} &
 \cellcolor[HTML]{FFD700}55 &
 \cellcolor[HTML]{FFA500}60 &
 \cellcolor[HTML]{FFD700}50 &
 \cellcolor[HTML]{FFD700}55 &
 \multicolumn{2}{c|}{\cellcolor[HTML]{FFFFE0}45} &
 \cellcolor[HTML]{FF8C00}70 &
 \cellcolor[HTML]{FD6864}80 &
 \cellcolor[HTML]{FFA500}65 &
 \cellcolor[HTML]{FFA500}60 &
 \multicolumn{2}{c|}{\cellcolor[HTML]{FFD700}55} &
 \cellcolor[HTML]{E0FFFF}30 &
 \cellcolor[HTML]{ADD8E6}20 &
 \cellcolor[HTML]{E0FFFF}35 &
 \cellcolor[HTML]{FFFFE0}40 &
 \multicolumn{2}{c}{\cellcolor[HTML]{FFFFE0}45} \\ \midrule
\multicolumn{1}{c|}{\textbf{Qwen3 Max}} &
 \cellcolor[HTML]{FD6864}82 &
 \cellcolor[HTML]{FD6864}84 &
 \cellcolor[HTML]{FD6864}84 &
 \cellcolor[HTML]{FD6864}85 &
 \multicolumn{2}{c|}{\cellcolor[HTML]{FD6864}85} &
 \cellcolor[HTML]{DC143C}92 &
 \cellcolor[HTML]{FD6864}88 &
 \cellcolor[HTML]{FD6864}88 &
 \cellcolor[HTML]{FD6864}88 &
 \multicolumn{2}{c|}{\cellcolor[HTML]{FD6864}88} &
 \cellcolor[HTML]{FFD700}50 &
 \cellcolor[HTML]{FFD700}50 &
 \cellcolor[HTML]{FFFFE0}48 &
 \cellcolor[HTML]{FFD700}50 &
 \multicolumn{2}{c|}{\cellcolor[HTML]{FFD700}50} &
 \cellcolor[HTML]{E0FFFF}33 &
 \cellcolor[HTML]{FFFFE0}41 &
 \cellcolor[HTML]{FFFFE0}49 &
 \cellcolor[HTML]{FFFFE0}46 &
 \multicolumn{2}{c|}{\cellcolor[HTML]{FFFFE0}43} &
 \cellcolor[HTML]{FFA500}67 &
 \cellcolor[HTML]{FFD700}59 &
 \cellcolor[HTML]{FFD700}51 &
 \cellcolor[HTML]{FFD700}54 &
 \multicolumn{2}{c|}{\cellcolor[HTML]{FFD700}57} &
 \cellcolor[HTML]{E0FFFF}38 &
 \cellcolor[HTML]{FFFFE0}45 &
 \cellcolor[HTML]{FFD700}56 &
 \cellcolor[HTML]{FFD700}57 &
 \multicolumn{2}{c|}{\cellcolor[HTML]{FFFFE0}49} &
 \cellcolor[HTML]{FFA500}62 &
 \cellcolor[HTML]{FFD700}55 &
 \cellcolor[HTML]{FFFFE0}44 &
 \cellcolor[HTML]{FFFFE0}43 &
 \multicolumn{2}{c}{\cellcolor[HTML]{FFD700}51} \\ \midrule
\multicolumn{1}{c|}{\textbf{Gemini Flash}} &
 \cellcolor[HTML]{FF8C00}78 &
 \cellcolor[HTML]{FD6864}84 &
 \cellcolor[HTML]{FD6864}82 &
 \cellcolor[HTML]{FD6864}84 &
 \multicolumn{2}{c|}{\cellcolor[HTML]{FD6864}80} &
 \cellcolor[HTML]{FF8C00}77 &
 \cellcolor[HTML]{FD6864}81 &
 \cellcolor[HTML]{FD6864}80 &
 \cellcolor[HTML]{FD6864}83 &
 \multicolumn{2}{c|}{\cellcolor[HTML]{FD6864}82} &
 \cellcolor[HTML]{FFA500}64 &
 \cellcolor[HTML]{FFFFE0}43 &
 \cellcolor[HTML]{FFD700}50 &
 \cellcolor[HTML]{FFFFE0}47 &
 \multicolumn{2}{c|}{\cellcolor[HTML]{FFFFE0}41} &
 \cellcolor[HTML]{E0FFFF}36 &
 \cellcolor[HTML]{FFFFE0}43 &
 \cellcolor[HTML]{FFD700}50 &
 \cellcolor[HTML]{FFFFE0}47 &
 \multicolumn{2}{c|}{\cellcolor[HTML]{FFFFE0}41} &
 \cellcolor[HTML]{FFA500}64 &
 \cellcolor[HTML]{FFD700}57 &
 \cellcolor[HTML]{FFD700}50 &
 \cellcolor[HTML]{FFD700}53 &
 \multicolumn{2}{c|}{\cellcolor[HTML]{FFD700}59} &
 \cellcolor[HTML]{E0FFFF}39 &
 \cellcolor[HTML]{FFFFE0}45 &
 \cellcolor[HTML]{FFD700}58 &
 \cellcolor[HTML]{FFD700}56 &
 \multicolumn{2}{c|}{\cellcolor[HTML]{FFD700}50} &
 \cellcolor[HTML]{FFA500}61 &
 \cellcolor[HTML]{FFD700}55 &
 \cellcolor[HTML]{FFFFE0}42 &
 \cellcolor[HTML]{FFFFE0}44 &
 \multicolumn{2}{c}{\cellcolor[HTML]{FFD700}51} \\ \midrule
\multicolumn{1}{c|}{\textbf{Qwen3-VL-235B}} &
 \cellcolor[HTML]{FF8C00}70 &
 \cellcolor[HTML]{FF8C00}76 &
 \cellcolor[HTML]{FF8C00}79 &
 \cellcolor[HTML]{FD6864}81 &
 \multicolumn{2}{c|}{\cellcolor[HTML]{FF8C00}74} &
 \cellcolor[HTML]{FF8C00}76 &
 \cellcolor[HTML]{FD6864}80 &
 \cellcolor[HTML]{FD6864}80 &
 \cellcolor[HTML]{FD6864}82 &
 \multicolumn{2}{c|}{\cellcolor[HTML]{FF8C00}77} &
 \cellcolor[HTML]{FFD700}56 &
 \cellcolor[HTML]{FFD700}59 &
 \cellcolor[HTML]{FFA500}62 &
 \cellcolor[HTML]{FFA500}63 &
 \multicolumn{2}{c|}{\cellcolor[HTML]{FFD700}58} &
 \cellcolor[HTML]{E0FFFF}31 &
 \cellcolor[HTML]{FFFFE0}44 &
 \cellcolor[HTML]{FFFFE0}49 &
 \cellcolor[HTML]{FFFFE0}46 &
 \multicolumn{2}{c|}{\cellcolor[HTML]{FFFFE0}42} &
 \cellcolor[HTML]{FFA500}69 &
 \cellcolor[HTML]{FFFFE0}48 &
 \cellcolor[HTML]{FFD700}55 &
 \cellcolor[HTML]{FFD700}58 &
\multicolumn{2}{c|}{\cellcolor[HTML]{FFD700}50} &
 \cellcolor[HTML]{FFFFE0}42 &
 \cellcolor[HTML]{FFFFE0}48 &
 \cellcolor[HTML]{FFD700}55 &
 \cellcolor[HTML]{FFD700}58 &
 \multicolumn{2}{c|}{\cellcolor[HTML]{FFD700}50} &
 \cellcolor[HTML]{FFD700}58 &
 \cellcolor[HTML]{FFD700}52 &
 \cellcolor[HTML]{FFFFE0}45 &
 \cellcolor[HTML]{FFFFE0}42 &
 \multicolumn{2}{c}{\cellcolor[HTML]{FFD700}50} \\ \midrule
\multicolumn{1}{c|}{\textbf{Gemini 2.5 Flash}} &
 \cellcolor[HTML]{FF8C00}74 &
 \cellcolor[HTML]{FF8C00}70 &
 \cellcolor[HTML]{FF8C00}72 &
 \cellcolor[HTML]{FFA500}68 &
 \multicolumn{2}{c|}{\cellcolor[HTML]{FF8C00}72} &
 \cellcolor[HTML]{FD6864}86 &
 \cellcolor[HTML]{FD6864}85 &
 \cellcolor[HTML]{FD6864}84 &
 \cellcolor[HTML]{FD6864}83 &
 \multicolumn{2}{c|}{\cellcolor[HTML]{FD6864}85} &
 \cellcolor[HTML]{E0FFFF}34 &
 \cellcolor[HTML]{E0FFFF}34 &
 \cellcolor[HTML]{E0FFFF}34 &
 \cellcolor[HTML]{E0FFFF}33 &
 \multicolumn{2}{c|}{\cellcolor[HTML]{E0FFFF}33} &
 \cellcolor[HTML]{E0FFFF}38 &
 \cellcolor[HTML]{FFFFE0}44 &
 \cellcolor[HTML]{FFFFE0}49 &
 \cellcolor[HTML]{FFFFE0}46 &
 \multicolumn{2}{c|}{\cellcolor[HTML]{FFFFE0}42} &
 \cellcolor[HTML]{FFA500}62 &
 \cellcolor[HTML]{FFD700}56 &
 \cellcolor[HTML]{FFD700}51 &
 \cellcolor[HTML]{FFD700}54 &
 \multicolumn{2}{c|}{\cellcolor[HTML]{FFD700}58} &
 \cellcolor[HTML]{FFFFE0}41 &
 \cellcolor[HTML]{FFFFE0}47 &
 \cellcolor[HTML]{FFD700}55 &
 \cellcolor[HTML]{FFD700}57 &
 \multicolumn{2}{c|}{\cellcolor[HTML]{FFFFE0}46} &
 \cellcolor[HTML]{FFD700}59 &
 \cellcolor[HTML]{FFD700}53 &
 \cellcolor[HTML]{FFFFE0}45 &
 \cellcolor[HTML]{FFFFE0}43 &
 \multicolumn{2}{c}{\cellcolor[HTML]{FFD700}54} \\ \midrule
\multicolumn{1}{c|}{\textbf{Qwen3-VL-30B}} &
 \cellcolor[HTML]{FF8C00}72 &
 \cellcolor[HTML]{FF8C00}78 &
 \cellcolor[HTML]{FF8C00}77 &
 \cellcolor[HTML]{FF8C00}79 &
 \multicolumn{2}{c|}{\cellcolor[HTML]{FF8C00}74} &
 \cellcolor[HTML]{FF8C00}76 &
 \cellcolor[HTML]{FD6864}80 &
 \cellcolor[HTML]{FF8C00}79 &
 \cellcolor[HTML]{FD6864}81 &
 \multicolumn{2}{c|}{\cellcolor[HTML]{FF8C00}78} &
 \cellcolor[HTML]{FFFFE0}47 &
 \cellcolor[HTML]{FFFFE0}48 &
 \cellcolor[HTML]{FFD700}50 &
 \cellcolor[HTML]{FFD700}50 &
 \multicolumn{2}{c|}{\cellcolor[HTML]{FFD700}50} &
 \cellcolor[HTML]{E0FFFF}34 &
 \cellcolor[HTML]{FFFFE0}43 &
 \cellcolor[HTML]{FFD700}51 &
 \cellcolor[HTML]{FFFFE0}48 &
 \multicolumn{2}{c|}{\cellcolor[HTML]{FFFFE0}45} &
 \cellcolor[HTML]{FFA500}66 &
 \cellcolor[HTML]{FFD700}57 &
 \cellcolor[HTML]{FFFFE0}49 &
 \cellcolor[HTML]{FFD700}52 &
 \multicolumn{2}{c|}{\cellcolor[HTML]{FFD700}55} &
 \cellcolor[HTML]{E0FFFF}39 &
 \cellcolor[HTML]{FFFFE0}46 &
 \cellcolor[HTML]{FFD700}56 &
 \cellcolor[HTML]{FFD700}58 &
 \multicolumn{2}{c|}{\cellcolor[HTML]{FFD700}50} &
 \cellcolor[HTML]{FFA500}61 &
 \cellcolor[HTML]{FFD700}54 &
 \cellcolor[HTML]{FFFFE0}44 &
 \cellcolor[HTML]{FFFFE0}42 &
 \multicolumn{2}{c}{\cellcolor[HTML]{FFD700}50} \\ \midrule
\multicolumn{1}{c|}{\textbf{GPT-5}} &
 \cellcolor[HTML]{FF8C00}71 &
 \cellcolor[HTML]{FF8C00}73 &
 \cellcolor[HTML]{FF8C00}75 &
 \cellcolor[HTML]{FF8C00}77 &
 \multicolumn{2}{c|}{\cellcolor[HTML]{FF8C00}72} &
 \cellcolor[HTML]{FF8C00}77 &
 \cellcolor[HTML]{FF8C00}79 &
 \cellcolor[HTML]{FD6864}80 &
 \cellcolor[HTML]{FD6864}82 &
 \multicolumn{2}{c|}{\cellcolor[HTML]{FF8C00}78} &
 \cellcolor[HTML]{FFD700}53 &
 \cellcolor[HTML]{FFD700}55 &
 \cellcolor[HTML]{FFD700}57 &
 \cellcolor[HTML]{FFD700}59 &
 \multicolumn{2}{c|}{\cellcolor[HTML]{FFD700}55} &
 \cellcolor[HTML]{E0FFFF}34 &
 \cellcolor[HTML]{FFFFE0}42 &
 \cellcolor[HTML]{FFFFE0}47 &
 \cellcolor[HTML]{FFFFE0}46 &
 \multicolumn{2}{c|}{\cellcolor[HTML]{FFFFE0}43} &
 \cellcolor[HTML]{FFA500}66 &
 \cellcolor[HTML]{FFD700}58 &
 \cellcolor[HTML]{FFD700}53 &
 \cellcolor[HTML]{FFD700}54 &
 \multicolumn{2}{c|}{\cellcolor[HTML]{FFD700}57} &
 \cellcolor[HTML]{E0FFFF}38 &
 \cellcolor[HTML]{FFFFE0}44 &
 \cellcolor[HTML]{FFD700}56 &
 \cellcolor[HTML]{FFD700}58 &
 \multicolumn{2}{c|}{\cellcolor[HTML]{FFFFE0}48} &
 \cellcolor[HTML]{FFA500}62 &
 \cellcolor[HTML]{FFD700}58 &
 \cellcolor[HTML]{FFFFE0}44 &
 \cellcolor[HTML]{FFFFE0}42 &
 \multicolumn{2}{c}{\cellcolor[HTML]{FFD700}52} \\ \midrule
\multicolumn{1}{c|}{\textbf{GPT-4o}} &
 \cellcolor[HTML]{E0FFFF}39 &
 \cellcolor[HTML]{FFFFE0}41 &
 \cellcolor[HTML]{E0FFFF}38 &
 \cellcolor[HTML]{E0FFFF}34 &
 \multicolumn{2}{c|}{\cellcolor[HTML]{E0FFFF}35} &
 \cellcolor[HTML]{FFD700}52 &
 \cellcolor[HTML]{FFD700}58 &
 \cellcolor[HTML]{FFD700}53 &
 \cellcolor[HTML]{FFA500}60 &
 \multicolumn{2}{c|}{\cellcolor[HTML]{FFD700}51} &
 \cellcolor[HTML]{ADD8E6}7 &
 \cellcolor[HTML]{ADD8E6}10 &
  \cellcolor[HTML]{ADD8E6}8 &
  \cellcolor[HTML]{ADD8E6}11 &
 \multicolumn{2}{c|}{\cellcolor[HTML]{ADD8E6}9} &
 \cellcolor[HTML]{ADD8E6}19 &
 \cellcolor[HTML]{ADD8E6}13 &
 \cellcolor[HTML]{FFFFE0}45 &
 \cellcolor[HTML]{E0FFFF}38 &
 \multicolumn{2}{c|}{\cellcolor[HTML]{FFD700}50} &
 \cellcolor[HTML]{FFD700}51 &
 \cellcolor[HTML]{FD6864}85 &
 \cellcolor[HTML]{FFD700}53 &
 \cellcolor[HTML]{FFA500}61 &
 \multicolumn{2}{c|}{\cellcolor[HTML]{FFFFE0}40} &
 \cellcolor[HTML]{ADD8E6}0 &
 \cellcolor[HTML]{ADD8E6}1 &
 \cellcolor[HTML]{ADD8E6}17 &
 \cellcolor[HTML]{FFD700}56 &
 \multicolumn{2}{c|}{\cellcolor[HTML]{FF8C00}77} &
 \cellcolor[HTML]{DC143C}100 &
 \cellcolor[HTML]{DC143C}98 &
 \cellcolor[HTML]{FD6864}82 &
 \cellcolor[HTML]{FFFFE0}43 &
\multicolumn{2}{c}{\cellcolor[HTML]{ADD8E6}23} \\ \bottomrule
\end{tabular}%
```
}
\caption{\textit{The VideoQGen Benchmark.} Evaluating multimodal models across reasoning types, including basic understanding (\textbf{BU}), attribution (\textbf{Atr}), event reasoning (\textbf{ER}), reverse reasoning (\textbf{RR}), and counterfactual inference (\textbf{CI}). \textbf{Gemini $2.5$ Pro} leads overall across relevance, answerability, and diversity dimensions, showing strong consistency across all reasoning categories.}
\label{tab:VideoQGen_results}
\end{table*}

\paragraph{\textit{VideoQGen Benchmark.}} To complement the \textit{VideoQA} task, \textit{UDVideoQA} introduces the \textit{first-of-its-kind} \textit{VideoQGen} benchmark for the traffic domain. This benchmark assesses a model’s ability to generate meaningful, diverse, and contextually grounded questions directly from untrimmed urban videos. A total of $2,000$ questions were generated from a diverse subset of videos covering variations in traffic density, lighting, and weather conditions. Using the standardized question template, these samples were evaluated across $8$ SOTA \textit{VideoQGen} models (see Table~\ref{tab:VideoQGen_results}), following human-assessed metrics adapted from established video reasoning studies~\cite{Xiao2021nextqa}. Human evaluators rated each generated question on a scoring policy scaled from $1$ to $5$ across three key criteria:\begin{itemize}\item \textbf{\textit{Relevance:}} Whether the generated question aligns with visible content in the video.\item \textbf{\textit{Answerability:}} Whether the question can be answered using only the visual context.

\item \textbf{\textit{Diversity:}} The variety and novelty of generated questions for a given clip.
\end{itemize}To ensure the reliability of these human judgments, inter-annotator agreement was measured using the Pearson correlation coefficient. All \textit{QA} pairs and associated metadata (video file paths and annotations) were stored in a structured \textit{CSV} format and integrated into the annotation tool for scalable validation. As shown in Table~\ref{tab:VideoQGen_results}, \textit{Gemini 2.5 Pro}~\cite{Google2025Gemini} achieved the highest overall scores, followed closely by the \textit{Qwen 3}~\cite{Qwen3_2025} series. These models demonstrate strong capability for generating contextually grounded and semantically rich questions. In contrast, \textit{GPT}-based models~\cite{OpenAI2024GPT4o} performed less effectively in this domain-specific setting, highlighting the challenges of general-purpose models in structured, real-world traffic environments (details in Supplementary).

\subsection{Dataset Statistics}
The \textit{UDVideoQA} dataset comprises $16$ hours of traffic footage recorded at $30$~\textit{fps}, resulting in approximately $1.7$ million frames. Of this, $8$ hours of footage were densely annotated, yielding $28,800$ \textit{QA} pairs, corresponding to an average of one question per second of video. While $8$ hours may seem limited in scope, the dataset's focus on high-density traffic intersections ensures that the footage is replete with complex multi-agent interactions. This high frequency of events provides a rich and concentrated source of data, making this highly efficient for \textit{VideoQA} analysis (details in Supplementary). The dataset maintains a balanced coverage across the five reasoning categories, with approximately $5,700$–$5,800$ \textit{QA} pairs per category. Approximately \textit{$70$\%} of questions are pedestrian or vehicle-centric, while the remaining \textit{$30$\%} focus on road infrastructure, signage, and environmental context. Linguistically, question forms containing \textit{`Is'}, \textit{`What'}, and \textit{`How'} dominate at roughly \textit{$95$\%}, \textit{$85$\%}, and \textit{$23$\%} within their respective reasoning groups, reflecting structured and consistent language usage. 

\begin{figure*}[!t]
  \centering
\includegraphics[width=0.80\linewidth]{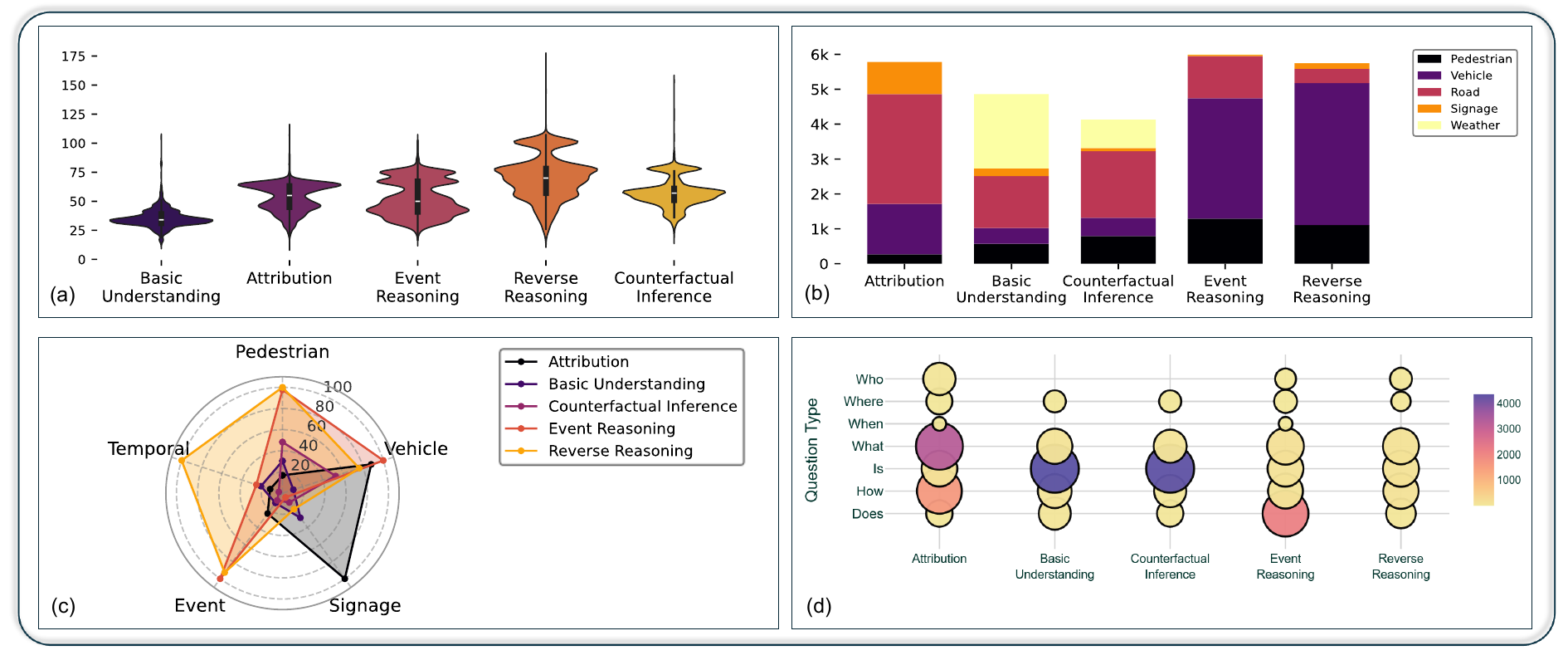}
\caption{\textit{UDVideoQA} dataset statistics. (a) Word frequency distribution by question type,
(b) Distribution of question categories across semantic domains, including pedestrians, vehicles, and environmental signage.
(c) Plot illustrating the spread of question sets across six contextual dimensions: pedestrian, temporal, vehicle, event, and signage, and 
(d) Frequency of questions containing key interrogative terms (\textit{``what'', ``where'', and ``is''}) across reasoning categories.}
\label{fig:data_stats}
\end{figure*}

To specifically assess temporal understanding, the dataset includes two key categories. The first, \textit{event reasoning}, accounts for cause-and-effect reasoning over time. Its prevalence of \textit{`Is'} based questions (e.g., \textit{``Focusing on the final three seconds.., is there any observable instance of a vehicle moving closer..?"}) requires models to perform a two-step verification: confirming the target object appeared within the specified interval and validating that the described action occurred. The second category, \textit{reverse reasoning}, tests the model’s ability to compare multiple events and infer their temporal order (e.g., \textit{“..which one entered the foreground... first..?”}). These questions require models to reason over the evolution of actions, distinguishing genuine video understanding from static, frame-based recognition (details in Supplementary). 
To further analyze dataset composition, we examined the underlying real-world priors across reasoning categories. The statistics, illustrated in Figure~\ref{fig:data_stats}, reveal notable patterns. Figure~\ref{fig:data_stats}(b) details the distribution of question categories, showing that \textit{event reasoning} is dominated by vehicle-related interactions, highlighting their role as primary causal agents. It also shows that \textit{basic understanding} questions often reference weather and illumination, reflecting perception of the broader environmental context. This contextual spread is further illustrated in Figure~\ref{fig:data_stats}(c), which maps the focus of each reasoning set across dimensions like \textit{`pedestrian'}, \textit{`vehicle'}, and \textit{`event'}. 

Furthermore, Figure~\ref{fig:data_stats}(a) and Figure~\ref{fig:data_stats}(d) provide insight into the linguistic structure, showing the word frequency distribution and the prevalence of key interrogative terms, respectively. This distribution mirrors real-world traffic dynamics and encourages models to develop domain-aware reasoning that integrates perception, context, and causality (details in Supplementary).

\subsection{QA Validation and Evaluation}

\textbf{\textit{QA Validation.}} 
To ensure the accuracy and clarity of annotations, a rigorous validation protocol was implemented. A team of ten trained annotators underwent domain-specific instruction on task requirements and the use of the annotation interface. For each video segment, annotators reviewed its corresponding \textit{QA} pair and made a binary decision-accept or reject-based on three criteria: \textbf{\textit{(1)}} question clarity, \textbf{\textit{(2)}} contextual relevance, and \textbf{\textit{(3)}} answerability using only the given video frames.
This structured workflow substantially improved efficiency, reducing manual validation time by an order of magnitude compared to earlier iterations. The validation tool is publicly released to facilitate transparency, reproducibility, and further research, as well as compatibility with other existing datasets ~\cite{RoadSocial2025,Xu2021SUTDTrafficQA}.

\textbf{\textit{QA Evaluation.}} Model performance on the \textit{UDVideoQA} benchmark is measured using \textit{VideoQA} accuracy metric, defined as the percentage of predictions that exactly match the corresponding ground-truth answers. For questions with multiple valid responses, a prediction is considered correct if it matches any of the annotated ground truths. This evaluation protocol ensures fair comparison across models with varying answer generation strategies.

\textbf{\textit{Model Compatibility.}}
The public release of this dataset is offered in \textit{JSONL} format, ensuring compatibility with leading multimodal frameworks of \textit{LLaVA}~\cite{Liu2024LLaVANeXT} and \textit{Qwen}~\cite{Qwen2_2024, Qwen3_2025}. This curated version establishes a reliable baseline for future work in large-scale \textit{VideoQA} and visual reasoning-based dataset development.

\section{Experiments and Analysis}
\label{sec:experiments}

To evaluate the effectiveness of \textit{UDVideoQA} 
as a benchmark for multimodal reasoning, a series of experiments was conducted to benchmark $10$ SOTA \textit{VideoLMs}. The analysis examines both zero-shot and fine-tuned performance, emphasizing the impact of domain adaptation on reasoning accuracy and visual grounding.

\subsection{Experimental Setup}
\paragraph{\textit{Dataset and Infrastructure.}}
The \textit{UDVideoQA} dataset was partitioned into training, validation, and test sets, with $5,000$ \textit{QA} pairs reserved for validation. The training and validation videos were taken from one group of intersections. The test set videos were taken from a completely different intersection, which was not used for training. All experiments were conducted on the university's high-performance computing cluster equipped with \textit{NVIDIA A$100$ GPUs}~\cite{Tsai2020GPU,Choquette2021GPU}. Model checkpoints and logs were automatically stored for reproducibility.

\paragraph{\textit{Model Configuration.}}
Both open-source and proprietary \textit{VideoLMs} were evaluated, including \textit{Gemini}~[\citenum{Google2024Gemini},\citenum{Google2025Gemini}], \textit{GPT-4o}~\cite{OpenAI2024GPT4o}, \textit{GPT-5}, \textit{Qwen-2.5-VL}~\cite{Alibaba2023QwenVL}, \textit{InternVL3}~\cite{Chen2024InternVL2}, \textit{VideoLLaMA 3}~\cite{zhang2025videollama3frontiermultimodal}, and \textit{NVILA}~\cite{Lin2024VILA} and \textit{LLava-NexT-Video}~\cite{Liu2024LLaVANeXT}.
For fine-tuning, the models were trained using \textit{low-rank adaptation (LoRA)}~\cite{Hu2021LoRA} on parameter scales ranging from $7$B to $32$B. All experiments employed a four-GPU setup with consistent optimization and batch configurations.

\begin{table*}[h!]
\centering
\resizebox{\linewidth}{!}{%
\begin{tabular}{cccccccccccccccccccc}
\hline
&
 &
 \multicolumn{6}{c}{\textbf{Morning}} &
 \multicolumn{6}{c}{\textbf{Afternoon}} &
 \multicolumn{6}{c}{\textbf{Evening}} \\ \cline{3-20} 
\multirow{-2}{*}{\textbf{Model Type}} &
\multirow{-2}{*}{\textbf{Model Name}} &
 \textbf{BU} &
 \textbf{Atr} &
 \textbf{ER} &
 \textbf{RR} &
 \textbf{CI} &
 \textbf{Overall} &
 \textbf{BU} &
 \textbf{Atr} &
 \textbf{ER} &
 \textbf{RR} &
 \textbf{CI} &
 \textbf{Overall} &
 \textbf{BU} &
 \textbf{Atr} &
 \textbf{ER} &
 \textbf{RR} &
 \textbf{CI} &
 \textbf{Overall} \\ \hline

\multirow{4}{*}{\rotatebox[origin=c]{0}{\hspace{1.2em}\textbf{Proprietary}}} 
 & \textbf{Gemini 2.5 Pro} & \cellcolor[HTML]{FD6864}90.00 & \cellcolor[HTML]{ADD8E6}22.22 & \cellcolor[HTML]{FD6864}88.89 & \cellcolor[HTML]{FD6864}86.10 & \cellcolor[HTML]{DC143C}91.70 & \cellcolor[HTML]{FF8C00}75.78 & \cellcolor[HTML]{FF8C00}78.27 & \cellcolor[HTML]{FFA500}65.44 & \cellcolor[HTML]{FF8C00}75.42 & \cellcolor[HTML]{FF8C00}73.81 & \cellcolor[HTML]{FF8C00}72.56 & \cellcolor[HTML]{FF8C00}73.10 & \cellcolor[HTML]{FD6864}87.39 & \cellcolor[HTML]{FFD700}50.00 & \cellcolor[HTML]{FFA500}62.09 & \cellcolor[HTML]{FF8C00}76.37 & \cellcolor[HTML]{FF8C00}79.80 & \cellcolor[HTML]{FF8C00}71.13 \\
 & \textbf{Gemini 2.5 Flash} & \cellcolor[HTML]{FD6864}80.00 & \cellcolor[HTML]{ADD8E6}22.22 & \cellcolor[HTML]{FF8C00}77.78 & \cellcolor[HTML]{FF8C00}77.78 & \cellcolor[HTML]{FD6864}89.88 & \cellcolor[HTML]{FFA500}69.53 & \cellcolor[HTML]{FD6864}82.87 & \cellcolor[HTML]{FF8C00}75.61 & \cellcolor[HTML]{FF8C00}74.91 & \cellcolor[HTML]{FF8C00}75.65 & \cellcolor[HTML]{FFA500}65.85 & \cellcolor[HTML]{FF8C00}74.98 & \cellcolor[HTML]{FFD700}50.00 & \cellcolor[HTML]{FFD700}50.00 & \cellcolor[HTML]{FFA500}68.68 & \cellcolor[HTML]{FF8C00}72.53 & \cellcolor[HTML]{FFA500}61.60 & \cellcolor[HTML]{FFA500}60.56 \\
 & \textbf{GPT-5} & \cellcolor[HTML]{FD6864}80.00 & \cellcolor[HTML]{ADD8E6}25.00 & \cellcolor[HTML]{FFD700}50.00 & \cellcolor[HTML]{FFA500}63.89 & \cellcolor[HTML]{FD6864}82.22 & \cellcolor[HTML]{FFA500}60.22 & \cellcolor[HTML]{FF8C00}74.83 & \cellcolor[HTML]{FFA500}69.96 & \cellcolor[HTML]{FF8C00}70.68 & \cellcolor[HTML]{FF8C00}70.13 & \cellcolor[HTML]{FFA500}68.90 & \cellcolor[HTML]{FF8C00}70.90 & \cellcolor[HTML]{FFD700}58.79 & \cellcolor[HTML]{FFD700}50.00 & \cellcolor[HTML]{FFFFE0}47.80 & \cellcolor[HTML]{FFD700}54.94 & \cellcolor[HTML]{FFD700}58.50 & \cellcolor[HTML]{FFD700}54.01 \\
 & \textbf{GPT-4o} & \cellcolor[HTML]{FF8C00}75.00 & \cellcolor[HTML]{ADD8E6}25.00 & \cellcolor[HTML]{FFA500}69.44 & \cellcolor[HTML]{FFFFE0}47.22 & \cellcolor[HTML]{FD6864}85.88 & \cellcolor[HTML]{FFA500}60.51 & \cellcolor[HTML]{FFD700}58.74 & \cellcolor[HTML]{ADD8E6}19.68 & \cellcolor[HTML]{E0FFFF}35.32 & \cellcolor[HTML]{ADD8E6}22.56 & \cellcolor[HTML]{FFFFE0}45.61 & \cellcolor[HTML]{E0FFFF}36.38 & \cellcolor[HTML]{FFD700}50.00 & \cellcolor[HTML]{ADD8E6}8.24 & \cellcolor[HTML]{FFD700}56.59 & \cellcolor[HTML]{FFD700}58.79 & \cellcolor[HTML]{FF8C00}76.38 & \cellcolor[HTML]{FFD700}50.00 \\ \hline

\multirow{6}{*}{\rotatebox[origin=c]{0}{\hspace{1.2em}\textbf{Open Source}}} 
 & \textbf{Qwen 2.5 32B} & \cellcolor[HTML]{FF8C00}75.66 & \cellcolor[HTML]{E0FFFF}36.11 & \cellcolor[HTML]{FFA500}66.67 & \cellcolor[HTML]{ADD8E6}25.00 & \cellcolor[HTML]{FF8C00}77.78 & \cellcolor[HTML]{FFD700}56.24 & \cellcolor[HTML]{FFA500}64.37 & \cellcolor[HTML]{FFFFE0}48.59 & \cellcolor[HTML]{FFD700}53.66 & \cellcolor[HTML]{FFFFE0}44.17 & \cellcolor[HTML]{FFD700}51.83 & \cellcolor[HTML]{FFD700}52.52 & \cellcolor[HTML]{FF8C00}72.07 & \cellcolor[HTML]{FFFFE0}40.78 & \cellcolor[HTML]{FFFFE0}45.96 & \cellcolor[HTML]{FF8C00}78.03 & \cellcolor[HTML]{FFD700}53.16 & \cellcolor[HTML]{FFD700}58.00 \\
 & \textbf{Qwen 2.5 7B (Fine Tune)} & \cellcolor[HTML]{FF8C00}74.80 & \cellcolor[HTML]{ADD8E6}16.67 & \cellcolor[HTML]{FF8C00}75.56 & \cellcolor[HTML]{FFD700}50.00 & \cellcolor[HTML]{FF8C00}77.78 & \cellcolor[HTML]{FFD700}58.96 & \cellcolor[HTML]{FFA500}63.22 & \cellcolor[HTML]{FFD700}53.11 & \cellcolor[HTML]{FFD700}54.27 & \cellcolor[HTML]{E0FFFF}39.88 & \cellcolor[HTML]{FFD700}53.66 & \cellcolor[HTML]{FFD700}52.83 & \cellcolor[HTML]{FFA500}66.07 & \cellcolor[HTML]{FF8C00}71.30 & \cellcolor[HTML]{FFD700}56.59 & \cellcolor[HTML]{FFD700}58.79 & \cellcolor[HTML]{FFD700}54.17 & \cellcolor[HTML]{FFA500}61.38 \\
 & \textbf{VideoLLama3} & \cellcolor[HTML]{FFA500}65.60 & \cellcolor[HTML]{ADD8E6}22.22 & \cellcolor[HTML]{FFD700}58.33 & \cellcolor[HTML]{FFD700}58.33 & \cellcolor[HTML]{FFA500}69.34 & \cellcolor[HTML]{FFD700}54.76 & \cellcolor[HTML]{FFD700}59.77 & \cellcolor[HTML]{E0FFFF}36.72 & \cellcolor[HTML]{FFFFE0}49.39 & \cellcolor[HTML]{ADD8E6}28.83 & \cellcolor[HTML]{FFFFE0}48.78 & \cellcolor[HTML]{FFFFE0}44.70 & \cellcolor[HTML]{FFA500}66.21 & \cellcolor[HTML]{FFD700}50.00 & \cellcolor[HTML]{FFD700}53.84 & \cellcolor[HTML]{FFD700}58.79 & \cellcolor[HTML]{FFFFE0}49.87 & \cellcolor[HTML]{FFD700}55.74 \\
 & \textbf{NVILA 8B} & \cellcolor[HTML]{FFD700}55.32 & \cellcolor[HTML]{ADD8E6}22.22 & \cellcolor[HTML]{FF8C00}72.22 & \cellcolor[HTML]{FFFFE0}47.22 & \cellcolor[HTML]{FFD700}53.33 & \cellcolor[HTML]{FFD700}50.06 & \cellcolor[HTML]{FFA500}64.37 & \cellcolor[HTML]{E0FFFF}33.90 & \cellcolor[HTML]{FFD700}54.88 & \cellcolor[HTML]{E0FFFF}30.06 & \cellcolor[HTML]{FFD700}52.44 & \cellcolor[HTML]{FFFFE0}47.13 & \cellcolor[HTML]{FFD700}52.31 & \cellcolor[HTML]{FFFFE0}49.82 & \cellcolor[HTML]{FFA500}49.19 & \cellcolor[HTML]{FFA500}45.47 & \cellcolor[HTML]{ADD8E6}50 & \cellcolor[HTML]{FFFFE0}49.44 \\
 & \textbf{InternVL3 38B} & \cellcolor[HTML]{FF8C00}70.00 & \cellcolor[HTML]{E0FFFF}36.11 & \cellcolor[HTML]{FFFFE0}47.22 & \cellcolor[HTML]{FFA500}66.67 & \cellcolor[HTML]{FFA500}67.22 & \cellcolor[HTML]{FFD700}57.44 & \cellcolor[HTML]{FFA500}61.49 & \cellcolor[HTML]{E0FFFF}36.72 & \cellcolor[HTML]{FFFFE0}46.95 & \cellcolor[HTML]{ADD8E6}26.99 & \cellcolor[HTML]{FFD700}59.15 & \cellcolor[HTML]{FFFFE0}46.26 & \cellcolor[HTML]{E0FFFF}38.33 & \cellcolor[HTML]{FFFFE0}47.22 & \cellcolor[HTML]{FFD700}53.33 & \cellcolor[HTML]{FFA500}65.00 & \cellcolor[HTML]{FFFFE0}40.02 & \cellcolor[HTML]{FFFFE0}48.78 \\
 & \textbf{Llava-NeXT Video 7B} & \cellcolor[HTML]{FFFFE0}44.80 & \cellcolor[HTML]{ADD8E6}2.78 & \cellcolor[HTML]{FFA500}63.89 & \cellcolor[HTML]{E0FFFF}33.33 & \cellcolor[HTML]{ADD8E6}22.22 & \cellcolor[HTML]{E0FFFF}33.40 & \cellcolor[HTML]{FFFFE0}45.40 & \cellcolor[HTML]{ADD8E6}16.95 & \cellcolor[HTML]{ADD8E6}23.78 & \cellcolor[HTML]{ADD8E6}16.56 & \cellcolor[HTML]{ADD8E6}25.61 & \cellcolor[HTML]{ADD8E6}25.66 & \cellcolor[HTML]{E0FFFF}30.10 & \cellcolor[HTML]{ADD8E6}1.10 & \cellcolor[HTML]{FFD700}54.40 & \cellcolor[HTML]{ADD8E6}23.63 & \cellcolor[HTML]{ADD8E6}5.49 & \cellcolor[HTML]{ADD8E6}22.94 \\

\hline
\end{tabular}%
}
\caption{The \textit{VideoQA} Benchmark. It presents comparative performance across proprietary and open-source video reasoning models evaluated under varying illumination conditions: high intensity (afternoon), medium intensity (morning), and low intensity (evening). Overall results show that Gemini 2.5 Pro achieved the highest accuracy in the Morning (75.78\%) and Evening (71.13\%) conditions, while Gemini 2.5 Flash scored the highest in the Afternoon(74.98\%). \textit{Llava-NeXT-Video} exhibited the weakest overall performance.}
\label{VideoQA}
\end{table*}

\subsection{Evaluation Methodology}
The evaluation framework integrates quantitative metrics and qualitative assessment through a two-stage process:
\begin{enumerate}
    \item \textbf{\textit{Zero-shot evaluation:}} All models are evaluated on the test set without additional training to assess generalization ability.
    \item \textbf{\textit{Fine-tuning:}} Selected open-source models are fine-tuned on the \textit{UDVideoQA} training set to measure adaptability and domain transfer performance.
\end{enumerate}

\paragraph{\textit{Semantic-Semantic Scoring via LLM Judge.}}
Traditional \textit{VideoQA} evaluation relies on lexical similarity metrics (\textit{BLEU}, \textit{ROUGE})~\cite{Ganesan2018ROGUE,Papineni2002BLEU,Yang2018RB} that penalize semantically correct but lexically different answers. To address this limitation, we introduce a semantic-semantic matching approach using an external \textit{LLM} Judge ~\cite{Marcu2024LingoQA}.
Specifically, \textit{Gemini 2.5 Pro}~\cite{Marcu2024LingoQA} was employed as a text-only evaluator. To mitigate bias, we employed a human annotator to provide unbiased assessments of answer correctness after the model's assessment. Each response was assigned a binary score, $S_{LLM} \in {0, 1}$, indicating whether the predicted answer semantically aligns with the ground truth.

\subparagraph{\textit{Weighted Complexity-based Scoring.}}
To capture the varying cognitive difficulty of reasoning categories, we introduce a weighted scoring formulation that prioritizes success in higher-order reasoning tasks.
Weights were assigned based on increasing inferential complexity: $1.0$ (BU), $1.2$ (Atr), $1.3$ (ER and RR), and $1.5$ (CI). The final model score $W$ is computed as:
\begin{equation}
W = \frac{1}{N} \sum_{i=1}^{N} S_{\text{LLM}_i} \cdot W_{c_i},
\end{equation}
where $S_{\text{LLM}_i}$ is the correctness score of the $i$-th response, and $W_{c_i}$ is the weight assigned to its reasoning category $c_i$. This formulation rewards success in cognitively demanding categories, especially counterfactual inference, which targets hallucination reduction and deeper causal understanding.
(details in Supplementary)

\paragraph{\textit{Human Validation for Bias Reduction.}}
To ensure reliability and minimize potential bias in \textit{LLM}-based evaluation, a \textit{human-in-the-loop validation} process was implemented. A “\textit{standard subset}” of $1$k predictions was randomly sampled across models and reasoning categories and independently annotated by domain experts. The Pearson correlation agreement (PCA) metric was computed between human and \textit{LLM} evaluations, and the prompts were iteratively refined until a high level of consistency was achieved.

\begin{figure}[t]
  \centering
\includegraphics[width=1\linewidth]{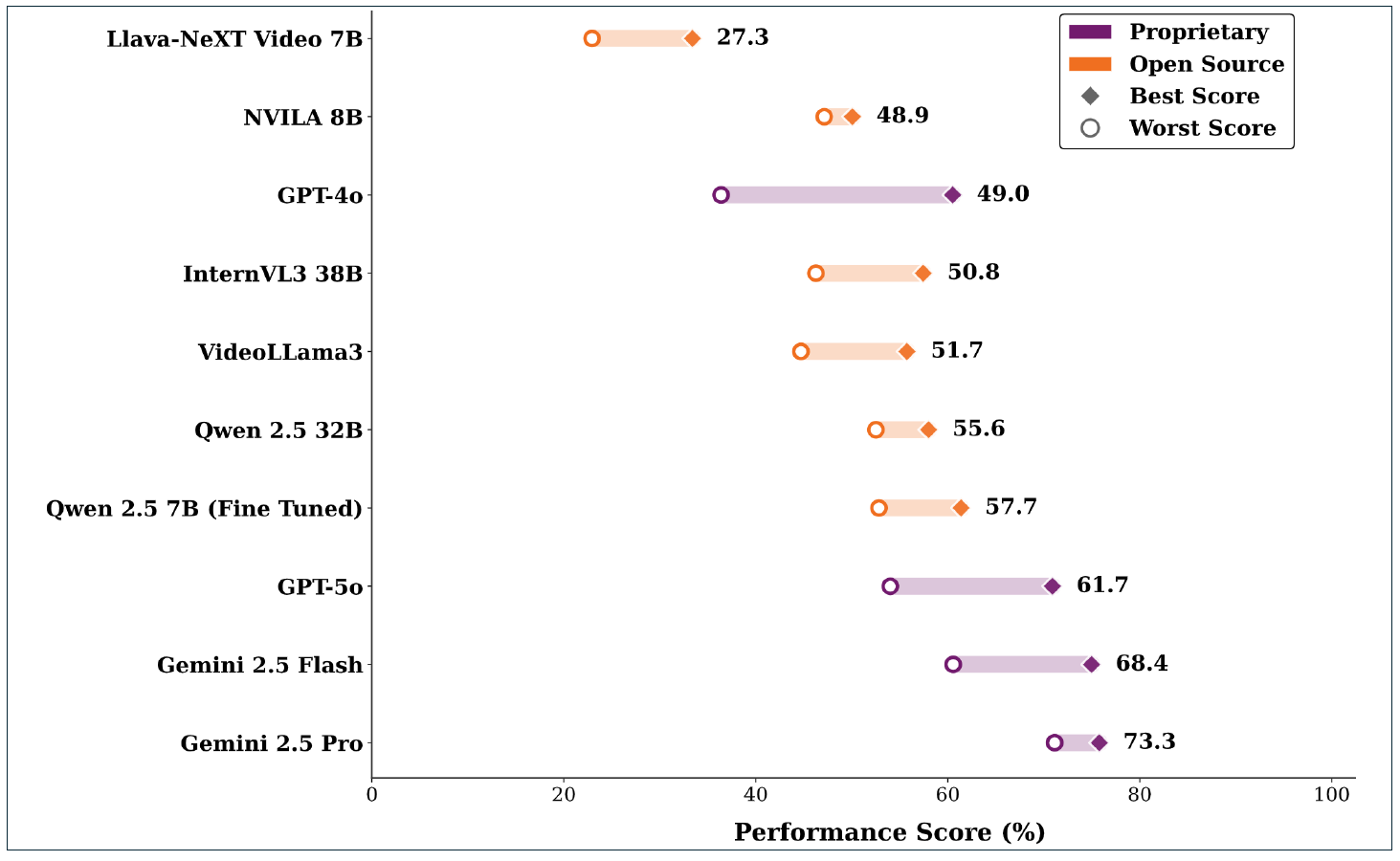}
\caption{Cross-model performance comparison on the \textit{UDVideoQA} benchmark. Each bar shows the range from worst to best performance of \textit{VideoLMs} used in this study.
}
\label{vqa_parallel}
\end{figure}

\subsection{Quantitative Evaluation}
The \textit{VideoQA} benchmarking results, presented in Table~\ref{VideoQA} and visualized in Figure~\ref{vqa_parallel}, reveal substantial performance variations and highlight key model-specific strengths and weaknesses.
Proprietary models, led by \textit{Gemini 2.5 Pro}, set the baseline with the highest overall scores across all conditions ($75.78$\% in \textit{``morning"}). This model excels in high-level reasoning, scoring $91.70$\% in \textit{CI} and $86.10$\% in \textit{RR}. However, the table exposes a critical weakness: this reasoning capability does not imply perceptual robustness. \textit{Gemini 2.5 Pro}'s \textit{Atr} score in the same \textit{``morning"} condition is only $22.22$\%, a trend of poor visual grounding seen across most proprietary models.

In contrast, open-source models show wider variance. Some, like \textit{LLaVA-NeXT-Video 7B}, struggle significantly, with performance bottoming out in the \textit{``evening"} condition, where its \textit{attribution} accuracy drops to just $1.10$\%. However, the results also highlight a key opportunity: compact, fine-tuned open-source models can be highly competitive. The \textit{Qwen-2.5-VL 7B (fine-tuned)} model achieved the highest open-source score in the \textit{``evening"} condition ($61.38$\%). This performance not only leads its peers but also significantly outperforms larger proprietary systems like \textit{GPT-5} ($54.01$\%) and \textit{GPT-4o} ($50.0$\%) in the same low-light scenario as in Table~\ref{VideoQA}. The figure shows the overall comparison of all models~\ref{vqa_parallel}.

\begin{table}[]
\centering
\resizebox{0.95\linewidth}{!}{%
\begin{tabular}{c|cc}
\toprule
\multirow{2}{*}{\textbf{Dataset}} & \multicolumn{2}{c}{\textbf{Overall Score across QA Pairs}} \\ \cline{2-3} 
            & \multicolumn{1}{c|}{\textbf{Baseline}} & \textbf{Fine Tuned} \\ \hline
RoadSocial~\cite{RoadSocial2025}  & \multicolumn{1}{c|}{52.75}             & 60                  \\ \hline
SUTDTraffic~\cite{Xu2021SUTDTrafficQA} & \multicolumn{1}{c|}{59.05}             & 62.7                \\ \bottomrule
\end{tabular}%
}
\caption{Cross-dataset generalization results of \textit{Qwen2.5-VL-7B} evaluated on RoadSocial and SUTDTrafficQA. The table reports the overall \textit{QA} accuracy (in \%) for the baseline model and after fine-tuning on \textit{UDVideoQA}.}
\label{tab:Cross-dataset generalization}
\end{table}

\subsection{Ablation Study}
\paragraph{\textit{Multi-Agent behavior Analysis.}}
This study analyzed traffic footage using an object detection and interaction framework to justify the dynamics of multi-agent behaviors in a traffic setting. This study employed the \textit{YOLOv11} model to perform per-frame object detection, identifying and localizing a set of traffic-related classes, including \textit{person}, \textit{bicycle}, \textit{car}, \textit{motorcycle}, \textit{bus}, and \textit{truck}, and all traffic-related entities. For each frame, the system generated a count of detected objects and then computed the Euclidean distances between the centroids of relevant agent pairs. To characterize potential interaction events, we extracted the minimum distance for three interaction types: vehicle-vehicle, pedestrian-pedestrian, and vehicle-pedestrian. (details in Supplementary)

\paragraph{\textit{Generalizability.}}
To test cross-domain generalization by training models on the custom \textit{UDVideoQA} dataset and testing them on the RoadSocial~\cite{RoadSocial2025} and SUTDTrafficQA~\cite{Xu2021SUTDTrafficQA} benchmarks. When it was evaluated across $500$ \textit{QA} pairs each in their respective \textit{QA} groups, the fine-tuned model demonstrated notable gains, achieving a \textbf{$7.3 \%$} improvement on RoadSocial and a \textbf{$4.3 \%$ }improvement on SUTDTrafficQA as shown in Table~\ref{tab:Cross-dataset generalization}. This enhanced performance is attributed to the model's improved ability to perceive objects in complex surroundings, developed from its training on the high-density traffic scenarios unique to the \textit{UDVideoQA} dataset (details in Supplementary).

\subsection{Qualitative Insights}

Qualitative analysis reveals several recurring patterns in model behavior regarding visual grounding and reliance on commonsense priors.

\begin{itemize}
    \item \textbf{\textit{Semantic vs. Contextual Grounding.}} Models often generate semantically correct answers (e.g., \textit{``collision''}) that lack proper \textit{contextual grounding}. For instance, a model might hallucinate a pedestrian when only vehicles were present, highlighting the limitations of text-based evaluation.

    \item \textbf{\textit{Attribution and Commonsense Discrepancy.}} A recurring discrepancy was observed: models often excelled at higher-order \textit{CI} (which relies on commonsense priors) but failed at basic \textit{Atr} tasks demanding precise pixel-level recognition. We attribute this failure to the high spatio-temporal density of agents in our data, which complicates precise visual grounding.

    \item \textbf{\textit{Causal and Temporal Challenges.}} Higher-order \textit{causal}, \textit{event}, and \textit{reverse reasoning} remain difficult, as current models struggle to model the interdependent, temporal cause-and-effect chains that these tasks require. The data in Table \ref{VideoQA} substantiates this claim, showing a recurring and significant performance gap where models' scores for \textit{event reasoning} and \textit{reverse reasoning}  are dramatically lower than for \textit{basic understanding}, confirming that complex temporal reasoning remains a key difficulty.
    
\end{itemize}
\section{Conclusion}
\label{sec:conclusion}
This paper presents \textbf{\textit{UDVideoQA}}, a large-scale benchmark for multimodal reasoning in urban environments, integrating $16$ hours of real-world video, a privacy-preserving framework, and $28,800$ QA-pairs to evaluate perception and causality in multi-agent contexts. Experiments revealed a fundamental limitation in current \textit{VLMs}, highlighting a significant gap between abstract reasoning and concrete perception, where models excel at complex counterfactual inference but fail at basic visual perception. On a promising note, domain-specific fine-tuning enables compact, open-source models like \textit{Qwen-2.5-VL 7B} to achieve performance comparable to proprietary systems, improving low-level attribution and demonstrating cross-domain generalization. These findings are reinforced by the \textit{UDVideoQA} dataset focuses on controlled, high-fidelity traffic scenes that emphasize consistency and causal clarity over uncontrolled scale. By releasing the dataset, annotation tools, and benchmarks, \textit{UDVideoQA} establishes a robust and ethically grounded foundation for advancing real-world multimodal reasoning and developing the next generation of perception–reasoning aligned \textit{VLMs}.

\section{Acknowledgment}
This research is supported by the NSF through the Partnerships for Innovation Program under Grant (\#2329780). The authors gratefully thank ARGOS Vision for providing the data and acknowledge Research Computing at Arizona State University for providing the research computing and storage resources that contributed to the results reported in this paper \cite{jennewein2023sol}.

{
 \small
 \bibliographystyle{ieeenat_fullname}
\bibliography{main}

@String(CVPR= {IEEE Conf. Comput. Vis. Pattern Recog.})

@String(ICCV= {Int. Conf. Comput. Vis.})

@String(ECCV= {Eur. Conf. Comput. Vis.})

@String(ICIP = {IEEE Int. Conf. Image Process.})

@String(AAAI = {AAAI})

@String(CVPRW= {IEEE Conf. Comput. Vis. Pattern Recog. Worksh.})

@String(CVPR  = {CVPR})

@String(ICCV  = {ICCV})

@String(ECCV  = {ECCV})

@String(ICIP  = {ICIP})

@String(CVPRW= {CVPRW})

@inproceedings{Xu2021SUTDTrafficQA,
  title={SUTD-TrafficQA: A Question Answering Benchmark and an Efficient Network for Video Reasoning Over Traffic Events},
  author={Xu, Li and Huang, He and Liu, Jun},
  booktitle=CVPR,
  pages={9878--9888},
  year={2021}
}

@inproceedings{Malla2023DRAMA,
  title={DRAMA: Joint Risk Localization and Captioning in Driving},
  author={Malla, Srikanth and Choi, Chiho and Dwivedi, Isht and Choi, Joon Hee and Li, Jiachen},
  booktitle={WACV},
  year={2023}
}

@inproceedings{Marcu2024LingoQA,
  title={LingoQA: Visual Question Answering for Autonomous Driving},
  author={Marcu, Ana-Maria and Chen, Long and H{\"u}nermann, Jan and Karnsund, Alice and Hanotte, Benoit and Chidananda, Prajwal and Nair, Saurabh and Badrinarayanan, Vijay and Kendall, Alex and Shotton, Jamie and Arani, Elahe and Sinavski, Oleg},
  booktitle=ECCV,
  year={2024}
}

@inproceedings{Qian2024NuScenesQA,
  title={NuScenes-QA: A Multi-modal Visual Question Answering Benchmark for Autonomous Driving Scenario},
  author={Qian, Tianwen and Chen, Jingjing and Zhuo, Linhai and Jiao, Yang and Jiang, Yu-Gang},
  booktitle=AAAI,
  year={2024}
}

@article{Zhou2025TUMTrafficVideoQA,
  title={TUMTraffic-VideoQA: A Benchmark for Unified Spatio-Temporal Video Understanding in Traffic Scenes},
  author={Zhou, Xingcheng and Larintzakis, Konstantinos and Guo, Hao and Zimmer, Walter and Liu, Mingyu and Cao, Hu and Zhang, Jiajie and Lakshminarasimhan, Venkatnarayanan and Strand, Leah and Knoll, Alois C},
  journal={arXiv preprint arXiv:2502.02449},
  year={2025}
}

@misc{RoadSocial2025,
  title={RoadSocial: A Diverse VideoQA Dataset and Benchmark for Road Event Understanding from Social Video Narratives},
  author={Parikh, Chirag and Rawat, Deepti and R. T., Rakshitha and Ghosh, Tathagata and Sarvadevabhatla, Ravi Kiran},
  year={2025},
  eprint={2503.21459},
  archivePrefix={arXiv},
  primaryClass={cs.CV}
}

@inproceedings{Carreira2017Kinetics,
  title     = {Quo Vadis, Action Recognition? {A} New Model and the Kinetics Dataset},
  author    = {Carreira, Joao and Zisserman, Andrew},
  booktitle = {Proceedings of the IEEE Conference on Computer Vision and Pattern Recognition (CVPR)},
  year      = {2017},
  pages     = {6299--6308}
}

@article{Yan2024LLaMAVID,
  title   = {{LLaMA-VID}: An Image is Worth 2 Tokens in Large Language Models},
  author  = {Li, Yanwei and Wang, Chengyao and Jia, Jiaya},
  journal = {arXiv preprint arXiv:2311.17043},
  year    = {2024}
}

@inproceedings{Wu2019LongTerm,
  title     = {Long-Term Feature Banks for Detailed Video Understanding},
  author    = {Wu, Chao-Yuan and Feichtenhofer, Christoph and Fan, Haoqi and He, Kaiming and Krahenbuhl, Philipp and Girshick, Ross},
  booktitle = {Proceedings of the IEEE/CVF Conference on Computer Vision and Pattern Recognition (CVPR)},
  year      = {2019},
  pages     = {284--293}
}

@inproceedings{Lei2021ClipBERT,
  title     = {Less is More: ClipBERT for Video-and-Language Learning via Sparse Sampling},
  author    = {Lei, Jie and Li, Linjie and Zhou, Luowei and Gan, Zhe and L. Berg, Tamara and Bansal, Mohit and Liu, Jingjing},
  booktitle = {Proceedings of the IEEE/CVF Conference on Computer Vision and Pattern Recognition (CVPR)},
  year      = {2021},
  pages     = {7331-7341}
}

@article{Lin2024VILA,
  title   = {VILA: On Pre-training for Visual Language Models},
  author  = {Lin, Ji and Yin, Hongxu and Ping, Wei and Lu, Yao and Molchanov, Pavlo and Tao, Andrew and Mao, Huizi and Kautz, Jan and Shoeybi, Mohammad and Han, Song},
  journal = {arXiv preprint arXiv:2312.07533},
  year    = {2024}
}

@article{Tsai2020GPU,
  title   = {Evaluating the Performance of NVIDIA's A100 Ampere GPU for Sparse Linear Algebra Computations},
  author  = {Tsai, Yuhsiang Mike and Cojean, Terry and Anzt, Hartwig},
  journal = {arXiv preprint arXiv:2008.08478},
  year    = {2020}
}

@article{Choquette2021GPU,
  title   = {NVIDIA A100 Tensor Core GPU: Performance and Innovation},
  author  = {Choquette, Jack and Gandhi, Wishwesh and Giroux, Olivier and Stam, Nick and Krashinsky, Ronny},
  journal = {IEEE Micro},
  year    = {2021}
}

@article{Google2024Gemini,
  title   = {Gemini 1.5: Unlocking multimodal understanding across millions of tokens of context},
  author  = {Gemini Team, Google},
  journal = {arXiv preprint arXiv:2402.10170},
  year    = {2024}
}

@article{OpenAI2024GPT4o,
  title   = {GPT-4 Technical Report},
  author  = {OpenAI},
  journal = {arXiv preprint arXiv:2303.08774},
  year    = {2024}
}

@article{Qwen2_2024,
    title   = {Qwen2 Technical Report},
    author  = {Qwen Team},
    journal = {arXiv preprint arXiv:2407.10671},
    year    = {2024}
}

@article{Qwen3_2025,
    title   = {Qwen3 Technical Report},
    author  = {Qwen Team},
    journal = {arXiv preprint arXiv:2505.09388},
    year    = {2025}
}

@article{Liu2024LLaVANeXT,
  author    = {Liu, Haotian and Li, Chunyuan and Li, Yuanhan and Li, Bo and Lee, Yong Jae},
  title     = {{LLaVA-NeXT}: Improved reasoning, OCR, and world knowledge},
  journal   = {arXiv preprint arXiv:2402.16694},
  year      = {2024}
}

@article{Hu2021LoRA,
  author    = {Hu, Edward J. and Shen, Yelong and Wallis, Phillip and Allen-Zhu, Zeyuan and Li, Yuanzhi and Wang, Shean and Wang, Lu and Chen, Weizhu},
  title     = {LoRA: Low-Rank Adaptation of Large Language Models},
  journal   = {arXiv preprint arXiv:2106.09685},
  year      = {2021}
}

@article{Ganesan2018ROGUE,
  author    = {Ganesan, Kavita},
  title     = {ROUGE 2.0: Updated and Improved Measures for Evaluation of Summarization Tasks},
  journal   = {arXiv preprint arXiv:1803.01937},
  year      = {2018}
}

@article{Papineni2002BLEU,
  author    = {Papineni, Kishore and Roukos, Salim and Ward, Todd and Zhu, Wei-Jing},
  title     = {BLEU: a method for automatic evaluation of machine translation},
  journal   = {ACL '02: Proceedings of the 40th Annual Meeting on Association for Computational Linguistics},
  year      = {2002}
}

@article{Yang2018RB,
  author    = {Yang, An and Liu, Kai and Liu, Jing and Lyu, Yajuan and Li, Sujian},
  title     = {Adaptations of ROUGE and BLEU to Better Evaluate Machine Reading Comprehension Task},
  journal   = {arXiv:1806.03578},
  year      = {2018}
}

@misc{zhu2025internvl3exploringadvancedtraining,
      title={InternVL3: Exploring Advanced Training and Test-Time Recipes for Open-Source Multimodal Models}, 
      author={Jinguo Zhu and Weiyun Wang and Zhe Chen and Zhaoyang Liu and Shenglong Ye and Lixin Gu and Hao Tian and Yuchen Duan and Weijie Su and Jie Shao and Zhangwei Gao and Erfei Cui and Xuehui Wang and Yue Cao and Yangzhou Liu and Xingguang Wei and Hongjie Zhang and Haomin Wang and Weiye Xu and Hao Li and Jiahao Wang and Nianchen Deng and Songze Li and Yinan He and Tan Jiang and Jiapeng Luo and Yi Wang and Conghui He and Botian Shi and Xingcheng Zhang and Wenqi Shao and Junjun He and Yingtong Xiong and Wenwen Qu and Peng Sun and Penglong Jiao and Han Lv and Lijun Wu and Kaipeng Zhang and Huipeng Deng and Jiaye Ge and Kai Chen and Limin Wang and Min Dou and Lewei Lu and Xizhou Zhu and Tong Lu and Dahua Lin and Yu Qiao and Jifeng Dai and Wenhai Wang},
      year={2025},
      eprint={2504.10479},
      archivePrefix={arXiv},
      primaryClass={cs.CV},
      url={https://arxiv.org/abs/2504.10479}, 
}

@misc{zhang2025videollama3frontiermultimodal,
      title={VideoLLaMA 3: Frontier Multimodal Foundation Models for Image and Video Understanding}, 
      author={Boqiang Zhang and Kehan Li and Zesen Cheng and Zhiqiang Hu and Yuqian Yuan and Guanzheng Chen and Sicong Leng and Yuming Jiang and Hang Zhang and Xin Li and Peng Jin and Wenqi Zhang and Fan Wang and Lidong Bing and Deli Zhao},
      year={2025},
      eprint={2501.13106},
      archivePrefix={arXiv},
      primaryClass={cs.CV},
      url={https://arxiv.org/abs/2501.13106}, 
}

@article{vishal2024eyes,
  title={Eyes on the Road: State-of-the-Art Video Question Answering Models Assessment for Traffic Monitoring Tasks},
  author={Vishal, Joseph Raj and Basina, Divesh and Choudhary, Aarya and Chakravarthi, Bharatesh},
  journal={arXiv preprint arXiv:2412.01132},
  year={2024}
}

@inproceedings{Zhao2021hacs,
  title     = {HACS: Human Action Clips and Segments Dataset for Recognition and Temporal Localization},
  author    = {Zhao, Hang and Torralba, Antonio and Torresani, Lorenzo and Yan, Zhicheng},
  booktitle = {Proceedings of the IEEE/CVF International Conference on Computer Vision (ICCV)},
  year      = {2019},
  pages     = {8668-8678}
}

@inproceedings{Xiao2021nextqa,
  title     = {NExT-QA: Next Phase of Question-Answering to Explaining Temporal Actions},
  author    = {Xiao, Junbin and Shang, Xindi and Yao, Angela and Chua, Tat-Seng},
  booktitle = {Proceedings of the IEEE/CVF Conference on Computer Vision and Pattern Recognition (CVPR)},
  year      = {2021},
  pages     = {4937--4947}
}

@article{Wang2024DriveLM,
  author    = {Xing, Shuo and Qian, Chengyuan and Wang, Yuping and Hua, Hongyuan and Tian, Kexin and Zhou, Yang and Tu, Zhengzhong},
  title     = {{OpenEMMA}: Open-Source Multimodal Model for End-to-End Autonomous Driving},
  journal   = {arXiv preprint arXiv:2412.15208},
  year      = {2024}
}

@inproceedings{Jamaludin2023Rank2Tell,
  author    = {Sachdeva, Enna and Agarwal, Nakul and Chundi, Suhas and Roelofs, Sean and Li, Jiachen and Kochenderfer, Mykel and Choi, Chiho and Dariush, Behzad},
  title     = {Rank2Tell: A Multimodal Driving Dataset for Joint Importance Ranking and Reasoning},
  booktitle = {Proceedings of the IEEE/CVF Winter Conference on Applications of Computer Vision (WACV)},
  year      = {2024},
  pages     = {7513-7522}
}

@inproceedings{Singh2021ROAD,
  author    = {Singh, Gurkirt and Akrigg, Stephen and Di Maio, Manuele and Fontana, Valentina and Javanmard Alitappeh, Reza and Khan, Salman},
  title     = {{ROAD}: The {Road} Event Awareness Dataset for Autonomous Driving},
  booktitle = {Proceedings of the IEEE/CVF International Conference on Computer Vision (ICCV)},
  year      = {2021},
  pages     = {13327--13336}
}

@inproceedings{Kim2020BDDX,
  author    = {Kim, Jinkyu and Moon, Suhong and Rohrbach, Anna and Darrell, Trevor and Canny, John},
  title     = {Advisable Learning for Self-Driving Vehicles by Internalizing Observation-to-Action Rules},
  booktitle = {Proceedings of the IEEE/CVF Conference on Computer Vision and Pattern Recognition (CVPR)},
  year      = {2020},
  pages     = {9661-9670}
}

@inproceedings{Rasouli2019BDDOIA,
  author    = {Rasouli, Amir and Kotseruba, Iuliia and Kunic, T. and Tsotsos, John K.},
  title     = {A multimodal dataset for pedestrian action anticipation},
  booktitle = {2019 IEEE/CVF International Conference on Computer Vision Workshop (ICCVW)},
  year      = {2019},
  pages     = {3602--3611}
}

@inproceedings{Marziliano2002BlurMetric,
  author    = {Marziliano, Pina and Dufaux, Frederic and Winkler, Stefan and Ebrahimi, Touradj},
  title     = {A no-reference perceptual blur metric},
  booktitle = {Proceedings of the International Conference on Image Processing (ICIP)},
  year      = {2002},
  volume    = {3},
  pages     = {57--60},
  doi       = {10.1109/ICIP.2002.1038902}
}

@inproceedings{Crete2007BlurEffect,
  author    = {Crete-Roffet, F. and Dolmiere, T. and Ladret, P. and Nicolas, M.},
  title     = {The blur effect: perception and estimation with a new no-reference perceptual blur metric},
  booktitle = {Human Vision and Electronic Imaging XII},
  year      = {2007},
  volume    = {6492},
  pages     = {64920I},
  doi       = {10.1117/12.702792}
}

@misc{fang2024abductiveegoviewaccidentvideo,
      title={Abductive Ego-View Accident Video Understanding for Safe Driving Perception}, 
      author={Jianwu Fang and Lei-lei Li and Junfei Zhou and Junbin Xiao and Hongkai Yu and Chen Lv and Jianru Xue and Tat-Seng Chua},
      year={2024},
      eprint={2403.00436},
      archivePrefix={arXiv},
      primaryClass={cs.CV},
      url={https://arxiv.org/abs/2403.00436}, 
}

@misc{Jocher2023YOLOv8,
  author       = {Glenn Jocher and Ayush Chaurasia and Jing, Li},
  title        = {YOLO by Ultralytics},
  howpublished = {\url{https://github.com/ultralytics/ultralytics}},
  year         = {2023}
}

@misc{OpenAI2025GPT5,
  author       = {OpenAI},
  title        = {Introducing GPT-5},
  howpublished = {\url{https://openai.com/index/introducing-gpt-5/}},
  year         = {2025},
  month        = aug,
  note         = {[Online; accessed 11-Nov-2025]}
}

@article{Kirillov2024SAM2,
  title  = {Segment Anything Model 2},
  author = {Kirillov, Alexander and et al.},
  journal= {arXiv preprint arXiv:2407.03991},
  year   = {2024}
}

@misc{zhang2022bytetrackmultiobjecttrackingassociating,
      title={ByteTrack: Multi-Object Tracking by Associating Every Detection Box}, 
      author={Yifu Zhang and Peize Sun and Yi Jiang and Dongdong Yu and Fucheng Weng and Zehuan Yuan and Ping Luo and Wenyu Liu and Xinggang Wang},
      year={2022},
      eprint={2110.06864},
      archivePrefix={arXiv},
      primaryClass={cs.CV},
      url={https://arxiv.org/abs/2110.06864}, 
}

@inproceedings{Rebecq2018ESIM,
  author    = {Rebecq, Henri and Gehrig, Daniel and Scaramuzza, Davide},
  title     = {{ESIM}: An Open Event Camera Simulator},
  booktitle = {Conference on Robot Learning (CoRL)},
  year      = {2018}
}

@inproceedings{Hu2021V2E,
  author    = {Hu, Yuhuang and Tsur, Elia and Vianello, Andrea and San, Daniel and Delbruck, Tobi},
  title     = {v2e: From video frames to realistic DVS events},
  booktitle = {2021 IEEE/CVF Conference on Computer Vision and Pattern Recognition Workshops (CVPRW)},
  year      = {2021},
  pages     = {1801--1809},
  doi       = {10.1109/CVPRW53098.2021.00196}
}

@article{Google2025Gemini,
  title   = {Gemini 2.5: Pushing the Frontier with Advanced Reasoning, Multimodality, Long Context, and Next Generation Agentic Capabilities},
  author  = {Gemini Team, Google},
  journal = {arXiv preprint arXiv:2507.06261},
  year    = {2025}
}

@article{Alibaba2023QwenVL,
  title   = {Qwen-VL: A Versatile Vision-Language Model for Understanding, Localization, Text Reading, and Image Generation},
  author  = {Bai, Jinze and Bai, Shuai and Yang, Shusheng and Wang, Shijie and Tan, Sinan and Wang, Peng and Lin, Junyang and Zhou, Chang and Zhou, Jingren},
  journal = {arXiv preprint arXiv:2308.12966},
  year    = {2023}
}

@article{Chen2024InternVL2,
  title   = {Expanding Performance Boundaries of Open-Source Multimodal Models with Model, Data, and Test-Time Scaling},
  author  = {InternVL Team},
  journal = {arXiv preprint arXiv:2412.05271},
  year    = {2024}
}

@article{khanam2024yolov11,
  title={Yolov11: An overview of the key architectural enhancements},
  author={Khanam, Rahima and Hussain, Muhammad},
  journal={arXiv preprint arXiv:2410.17725},
  year={2024}
}

@misc{NeuromorphicSystems2021IEBCS,
  author       = {neuromorphicsystems},
  title        = {{IEBCS: ICNS Event Based Camera Simulator}},
  year         = {2025},
  howpublished = {\url{https://github.com/neuromorphicsystems/IEBCS}},
  note         = {Accessed: 2025-11-04},
}

@article{xiong2024sam2,
  title={Sam2-unet: Segment anything 2 makes strong encoder for natural and medical image segmentation},
  author={Xiong, Xinyu and Wu, Zihuang and Tan, Shuangyi and Li, Wenxue and Tang, Feilong and Chen, Ying and Li, Siying and Ma, Jie and Li, Guanbin},
  journal={arXiv preprint arXiv:2408.08870},
  year={2024}
}

@inproceedings{chakravarthi2024recent,
  title={Recent event camera innovations: A survey},
  author={Chakravarthi, Bharatesh and Verma, Aayush Atul and Daniilidis, Kostas and Fermuller, Cornelia and Yang, Yezhou},
  booktitle={European Conference on Computer Vision},
  pages={342--376},
  year={2024},
  organization={Springer}
}

@inproceedings{chakravarthi2023event,
  title={Event-based sensing for improved traffic detection and tracking in intelligent transport systems toward sustainable mobility},
  author={Chakravarthi, Bharatesh and Manoj Kumar, M and Pavan Kumar, BN},
  booktitle={International Conference on Interdisciplinary Approaches in Civil Engineering for Sustainable Development},
  pages={83--95},
  year={2023},
  organization={Springer}
}

@inproceedings{jennewein2023sol,
  title={The Sol Supercomputer at Arizona State University},
  author={Jennewein, Douglas M. and Lee, Johnathan and Kurtz, Chris and Dizon, William and Shaeffer, Ian and Chapman, Alan and Chiquete, Alejandro and Burks, Josh and Carlson, Amber and Mason, Natalie and others},
  booktitle={Practice and Experience in Advanced Research Computing 2023: Computing for the Common Good (PEARC '23)},
  pages={296--301},
  year={2023}
}
}

\clearpage
\appendix

\twocolumn[
\begin{center}
    {\LARGE \bfseries 
    \textit{UDVideoQA}: A Traffic Video Question Answering Dataset for Multi-Object Spatio-Temporal Reasoning in Urban Dynamics\\
    Supplementary Material
    \par}
\end{center}

\vspace{1.0em}
]


\section*{0. Preliminary}
\label{sec:preliminary}

\noindent
\textbf{\textit{Ethical Disclaimer and Dataset Scope.}}  
All human subjects, vehicles, and entities containing potential personally identifiable information have been anonymized in full compliance with institutional ethical standards. The \textit{UDVideoQA} dataset employs an event-driven, motion-based blurring technique to ensure privacy preservation while maintaining scene fidelity.  

\noindent
\textbf{\textit{Dataset Scope and Representativeness.}}  
\begin{itemize}
    \item The \textit{UDVideoQA} dataset comprises \textit{$16$ hours of urban traffic footage}, providing a focused yet robust baseline for studying multimodal reasoning in dynamic environments.  
    \item Data were collected from  \textit{metropolitan regions} under diverse traffic, weather, and lighting conditions, ensuring consistent quality and controlled variability.  
    \item While geographically specific, this design promotes reproducibility and establishes a solid foundation for future community-driven multi-city extensions.  
\end{itemize}

\noindent
\textbf{\textit{Community Expansion and Open Collaboration.}}  
The \textit{UDVideoQA} benchmark is an \textit{open and evolving resource}, designed to grow through community participation. Contributions are encouraged via:
\begin{itemize}
    \item \textbf{\textit{Dataset Extension:}} Adding new recordings from varied regions and conditions. \textit{(\url{https://huggingface.co/UDVideoQA})}
    \item \textbf{\textit{Tool Enhancement:}} Improving the open-source annotation and validation toolkit. \textit{(\url{https://github.com/UDVideoQA/UDVideoQA-Annotation_Tools})}
    \item \textbf{\textit{Benchmark Expansion:}} Evaluating emerging \textit{VLMs} on the public leaderboard to advance multimodal reasoning. \textit{(\url{https://github.com/UDVideoQA/UDVideoQA-finetune})}

\end{itemize}

\noindent
Together, these efforts position \textit{UDVideoQA} as a collaborative, ethically grounded benchmark that fosters transparent and reproducible research in real-world multimodal reasoning.

\begin{figure}[t]
  \centering
  \includegraphics[width=0.95\linewidth]{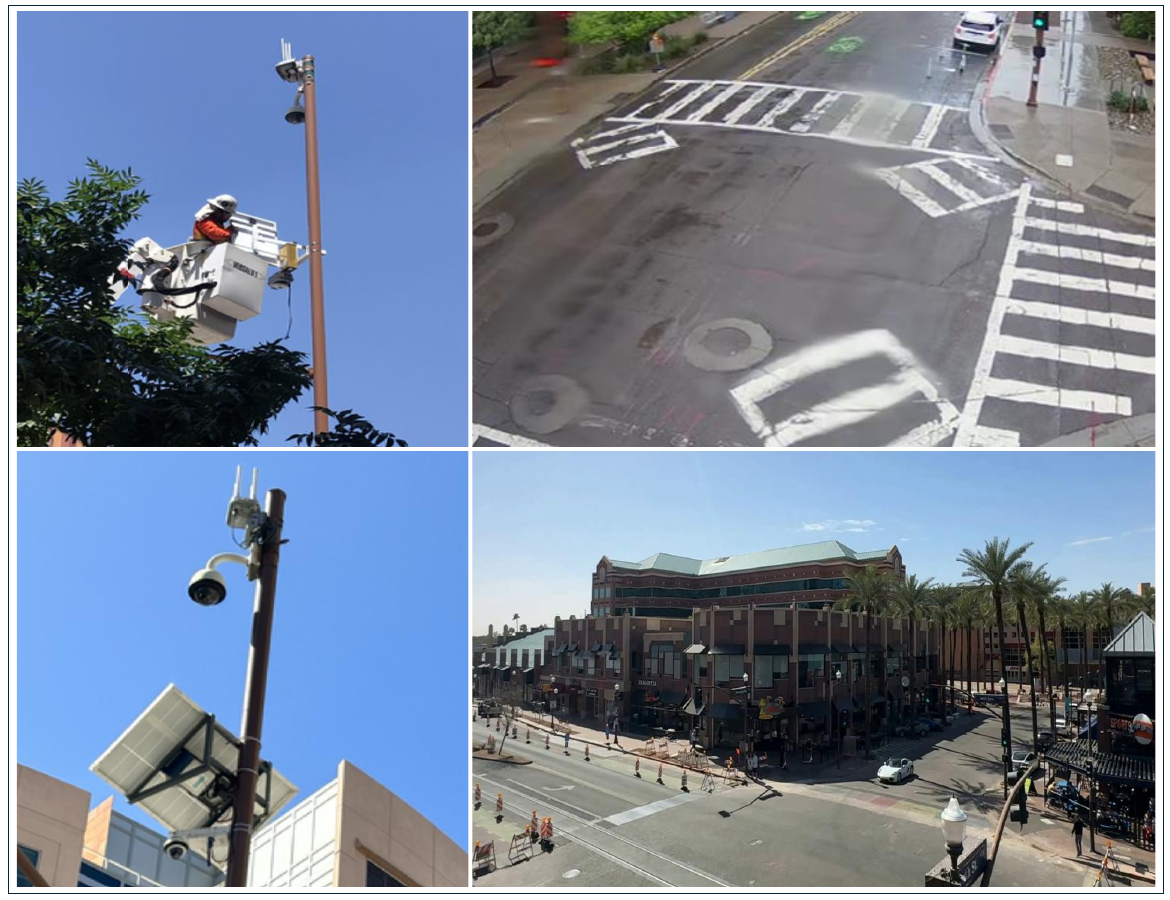}
  \caption{Deployment sites of traffic cameras used for data collection. The selected intersections capture diverse lighting and weather conditions, providing representative coverage of real-world urban traffic environments.}
  \label{camera_Details}
\end{figure}

\section{Data Collection and Setup}
\label{sec:data_collection}

\paragraph{\textit{Camera Deployment and Configuration.}}  
The data collection infrastructure comprised camera units strategically deployed across major urban intersections as part of an ongoing traffic study conducted jointly between a university research group and the metropolitan city authority. Each unit was designed for long-term, real-world deployment and integrated a low-power computational module based on the NVIDIA Jetson platform. This edge-computing configuration enabled on-board AI processing for object detection and classification directly at the point of capture, substantially reducing the need for continuous high-bandwidth video transmission. The deployed cameras, shown in Figure~\ref{camera_Details}, formed a distributed sensing network optimized for high-fidelity visual data acquisition under real-world urban conditions.

The system architecture was tailored for single-modality, high-resolution visual recording to ensure consistent frame quality across deployment sites. Each camera captured video at a resolution of \textit{$1920$ \ $\times$ $1080$ pixels (Full HD)} and a frame rate of \textit{$30$ fps}, preserving fine-grained spatial details such as vehicle markings, pedestrian trajectories, and traffic signal states. This configuration minimizes motion blur and maintains temporal fidelity, both of which are essential for reasoning-oriented \textit{VideoQA} tasks that depend on precise spatio-temporal understanding.

\begin{figure}[t]
  \centering
  \includegraphics[width=0.95\linewidth]{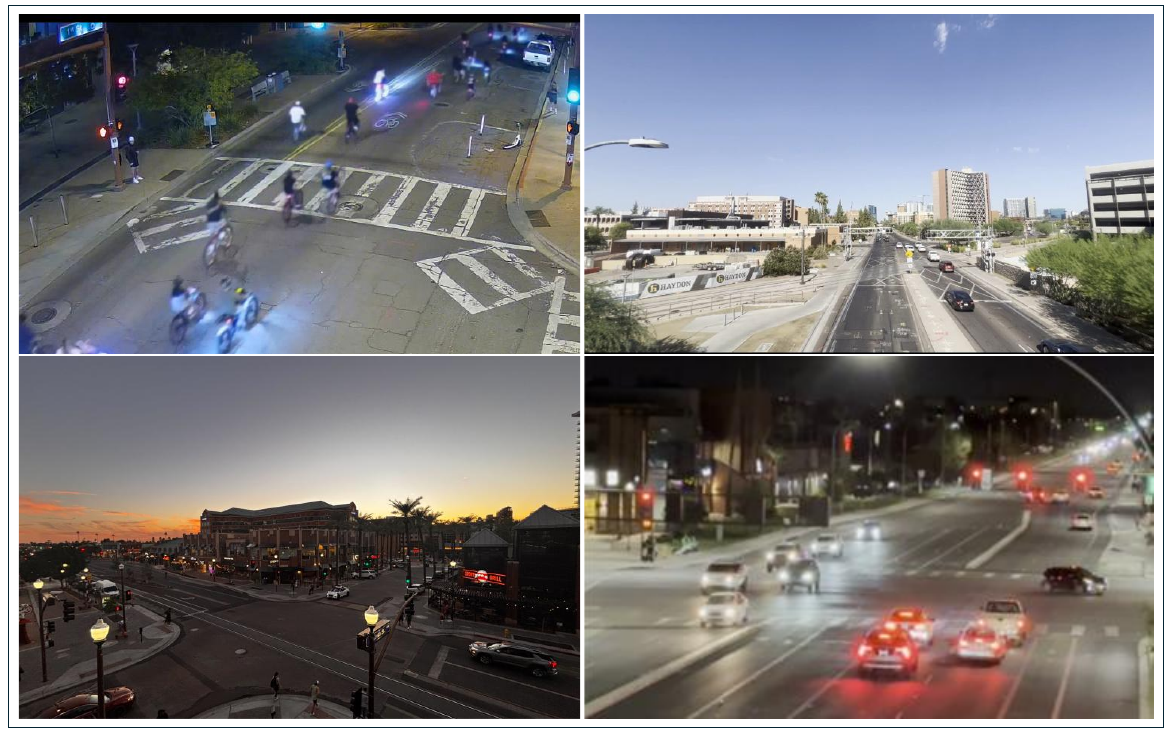}
  \caption{Strategic placement of traffic-monitoring cameras showing varied intersection types, traffic densities, and environmental conditions.}
  \label{fig:intersections}
\end{figure} 

\paragraph{\textit{Environmental Diversity and Recording Conditions.}}  
To capture the full variability of urban traffic dynamics, recordings were collected under diverse environmental conditions. The dataset spans multiple times of day including morning ($7$–$9$ AM), midday ($11$ AM–$1$ PM), evening ($5$–$7$ PM), and nighttime ($8$ PM–$12$ AM), and encompasses a wide range of weather scenarios such as clear, overcast, rainy, and dust-storm conditions. This diversity ensures that the dataset reflects realistic illumination changes and environmental variations encountered in daily traffic scenes, enabling the study of model robustness under challenging lighting, occlusion, and reflection conditions commonly observed in outdoor video data.

\paragraph{\textit{Urban Traffic Site Selection.}}  
The cameras were installed at multiple intersections that were deliberately selected to represent distinct intersection types and traffic configurations. Site selection prioritized locations with complex and dynamic agent interactions such as pedestrian crossings, turning vehicles, and micro-mobility activity, capturing the diverse behavioral patterns that emerge in dense urban environments. Each intersection exhibited high-frequency events involving multiple agents, providing rich contexts for evaluating causal reasoning and temporal understanding in \textit{VideoQA} models.
Figure~\ref{fig:intersections} shows traffic sites, highlighting the variety of traffic scenes covered in the dataset. In total, $16$ hours of traffic footage were collected, capturing a broad spectrum of real-world traffic participants and interactions that are critical for studying visual reasoning and multimodal understanding in complex urban scenes. The \textit{UDVideoQA} dataset is designed to form a robust foundation for benchmarking next-generation vision–language models under realistic conditions.

\section{Data Anonymity Pipeline}
\label{sec:Data_anonymity_pipeline}

Ensuring ethical data handling and privacy preservation was central to the design of the \textit{UDVideoQA} dataset. Since the collected traffic footage inherently contains pedestrians and vehicles that may reveal identifiable features, a dedicated anonymization pipeline was implemented to safeguard privacy without compromising visual fidelity. This section describes the ethical foundation, the transition from semantic to motion-based anonymization, the development of the event-driven dynamic blurring framework, and its empirical evaluation.

\subsection{Ethical Framework and IRB Compliance}
\label{sec:irb_compliance}

The university institutional review board (IRB) determined that the proposed work, conducting general traffic and pedestrian video collection under protocol STUDY00018234, does not constitute \textit{“research involving human subjects”} under DHHS and FDA regulations. Therefore, a formal review was not required. Nevertheless, because the data were captured in public spaces containing dynamic human and vehicular activity, the project adhered to stringent ethical guidelines to ensure complete anonymization of any potentially identifiable entities prior to dataset release. All processing and validation steps were performed under this compliance framework to balance ethical responsibility with the need for high-quality, research-grade data.

\subsection{Semantic vs. Motion-based Anonymization}
\label{sec:semantic_vs_motion}

Our initial anonymization experiments employed a semantic pipeline that combined an object detector (\textit{YOLOv11}~\cite{khanam2024yolov11}) with a segmentation model (\textit{SAM2}~\cite{Kirillov2024SAM2}) to generate per-object blurring masks. To enhance temporal consistency and manage occlusions, a multi-object tracking (MOT) framework, \textit{ByteTrack}~\cite{zhang2022bytetrackmultiobjecttrackingassociating}, was integrated to maintain identity associations across frames.
While this detector–segmenter pipeline achieved accurate anonymization in controlled conditions, it showed critical weaknesses in real-world traffic scenes. Detection accuracy degraded under occlusion, low illumination, or rapid motion, which the MOT framework could not correct due to unreliable initial detections. Furthermore, semantic segmentation occasionally misidentified static background objects (e.g., signage, advertisements), leading to unnecessary blurring and loss of contextual integrity.

\begin{figure}[t!]
  \centering
  \includegraphics[width=1\linewidth]{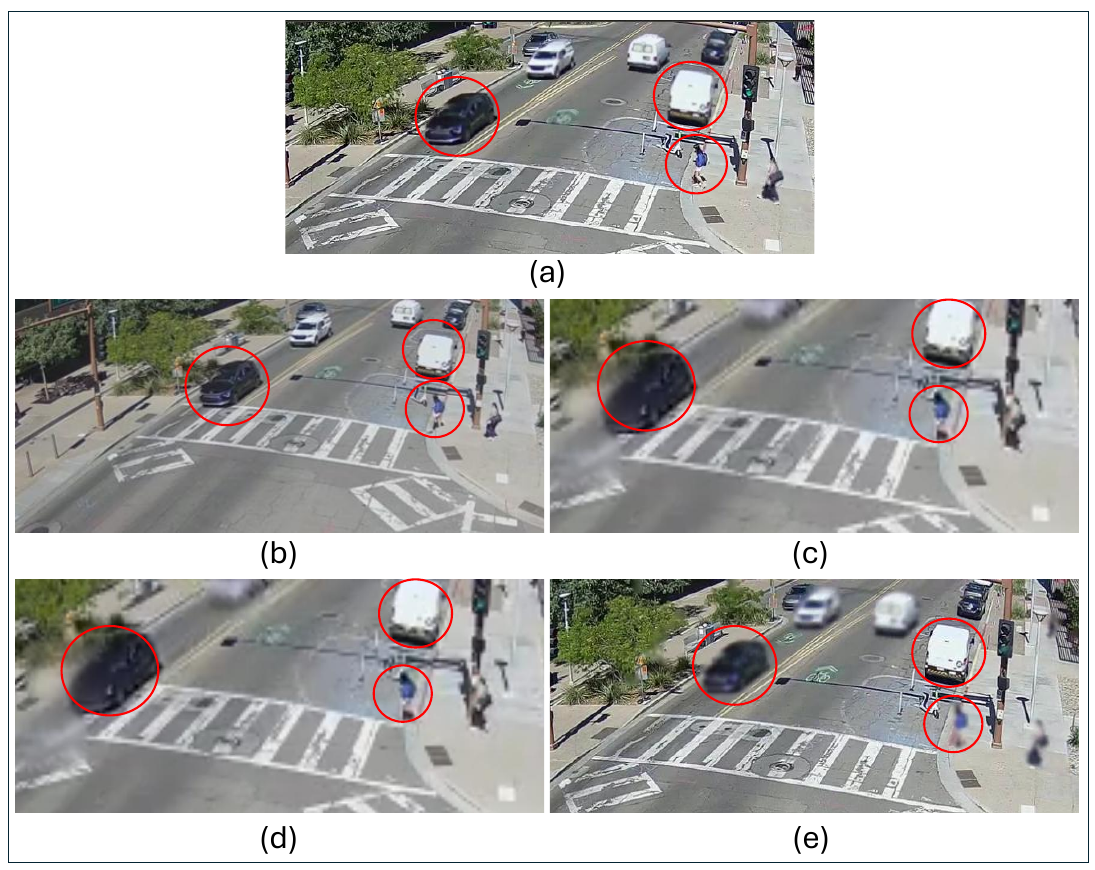}
  \caption{ Qualitative comparison of semantic and motion-based anonymization methods. (a) Original frame, (b) YOLOv11 + SAM2 semantic baseline, (c) IEBCS, (d) ESIM, and (e) V2E. The semantic approach (b) fails to blur the pedestrian highlighted in purple, whereas the event-based methods (c–e) accurately anonymize moving agents while preserving static background details.}
  \label{Semantic_and_motion-based_anonymization}
\end{figure}

To overcome these limitations, we adopted a motion-based anonymization approach inspired by the sensing principles of event cameras~\cite{chakravarthi2024recent, chakravarthi2023event}. Leveraging event-driven vision simulators, this framework detects motion directly through pixel-intensity changes rather than relying on semantic object cues. Unlike semantic methods that rely on object category predictions, these simulators operate directly on pixel intensity changes, detecting motion at the signal level. This allows for context-aware blurring of only moving entities while preserving static environmental details such as road markings or infrastructure. Figure~\ref{Semantic_and_motion-based_anonymization} qualitatively compares the semantic baseline and event-driven anonymization methods, illustrating improved consistency and precision in dynamic scenes.

\subsection{Motion-based Dynamic Blurring}
\label{sec:motion_blur}

\noindent
\textbf{\textit{Experimental Setup.}}  
A comparative analysis was conducted to identify the most effective event-based blurring technique for the annotation framework. The evaluation compared \textit{IEBCS}~\cite{NeuromorphicSystems2021IEBCS} against two other event simulators, \textit{ESIM}~\cite{rebecq2018esim} and \textit{V2E}~\cite{hu2021v2e}, as well as a semantic baseline combining \textit{YOLOv11}~\cite{khanam2024yolov11} with \textit{SAM2}~\cite{xiong2024sam2}. Experiments were performed on video clips recorded under both high- and low-intensity lighting to assess robustness across varied illumination conditions.  
To ensure fair comparison, all event simulators were normalized to apply a standardized blur operation derived from their generated motion masks.

\noindent
\textbf{\textit{Evaluation Metrics.}}  
Four quantitative metrics were used to assess blur quality:
\begin{itemize}
    \item \textbf{\textit{Fraction Blurred (F):}} Proportion of successfully anonymized objects, $F = B/N$.
    \item \textbf{\textit{Mean Edge Reduction ($\bar{r}$):}} Average reduction in edge strength per video.
    \item \textbf{\textit{Mean SSIM Reduction:}} Average change in structural similarity index measure (SSIM) between original and blurred regions, where higher (1 - \text{SSIM}) indicates stronger structural alteration.
    \item \textbf{\textit{Processing Time (s):}} Wall-clock time required to process one video clip.
\end{itemize}

\noindent
\textbf{\textit{Edge-Strength Metric.}}  
Blur effectiveness was evaluated per detected object using an edge-strength reduction metric computed via the Canny filter.  
Let `$I_{\text{orig}}$' and `$I_{\text{blur}}$' represent the grayscale image patches before and after blurring, respectively.  
Define `$C(\cdot)$' as the binary edge detector operator (Canny), producing an edge map `$E(I) = C(I) \in \{0,1\}^{H \times W}$', where `$E(I)_{i,j} = 1$' if an edge is detected at pixel $(i,j)$, and `$0$' otherwise.  
The total edge strength is defined as:
\[
S(I) = \|E(I)\|_0 = \sum_{i,j} E(I)_{i,j}.
\]

\noindent
\textbf{\textit{Per-Object Reduction.}}  
For each detected object (region of interest), the fractional edge reduction is given by:
\[
r = 1 - \frac{S_{\text{blur}}}{S_{\text{orig}}}, \quad \text{clipped to } 0 \le r \le 1,
\]
where `$S_{\text{orig}} = S(I_{\text{orig}})$' and `$S_{\text{blur}} = S(I_{\text{blur}})$'.  
Objects with low edge counts ($S_{\text{orig}} < \epsilon$, default $\epsilon = 8$ pixels) are excluded from evaluation.  
A blur is considered successful if $r \ge \tau$, where the blur threshold `$\tau$' is set to 0.45.

\vspace{2pt}
\noindent
\textbf{\textit{Aggregate Metrics.}}  
For a frame `$f$' with evaluated objects `$\mathcal{O}_f$', let `$N_f = |\mathcal{O}_f|$' denote the number of objects, `$B_f = \sum_{o \in \mathcal{O}_f} \mathbf{1}\{r_o \ge \tau\}$' the number of blurred objects, and `$\bar{r}_f = \frac{1}{N_f} \sum_{o \in \mathcal{O}_f} r_o$' the mean reduction.  
Aggregated over the video, the following quantities are reported:
\begin{itemize}
    \item Total evaluated objects: $N = \sum_f N_f$,
    \item Total blurred objects: $B = \sum_f B_f$,
    \item Per-video fraction blurred: $F = B/N$,
    \item Mean per-video reduction: $\bar{r} = (1/N)\sum_f\sum_{o \in \mathcal{O}_f} r_o$.
\end{itemize}

\vspace{2pt}
\noindent
\textbf{\textit{Edge Filter Parameters.}}  
The Canny edge detector was applied with thresholds $(T_{\text{low}}, T_{\text{high}}) = (100, 200)$.  
Default constants include an edge-count minimum `$\epsilon = 8$' and a blur-acceptance threshold `$\tau = 0.45$'.

\begin{table*}[t!]
\centering
\resizebox{0.85\textwidth}{!}{%
\begin{tabular}{c|cccc|cccc}
\toprule 
& \multicolumn{4}{c|}{\textbf{High Intensity Light}} & \multicolumn{4}{c}{\textbf{Low Intensity Light}} \\ \cline{2-9}
\multirow{-2}{*}{\textbf{Event Simulator}} &
\begin{tabular}[c]{@{}c@{}}Fraction \\ Blurred\end{tabular} &
\begin{tabular}[c]{@{}c@{}}Mean Edge \\ Detection\end{tabular} &
\begin{tabular}[c]{@{}c@{}}Mean SSIM \\ Reduction\end{tabular} &
\begin{tabular}[c]{@{}c@{}}CPU Processing \\ time(s)\end{tabular} &
\begin{tabular}[c]{@{}c@{}}Fraction \\ Blurred\end{tabular} &
\begin{tabular}[c]{@{}c@{}}Mean Edge \\ Detection\end{tabular} &
\begin{tabular}[c]{@{}c@{}}Mean SSIM \\ Reduction\end{tabular} &
\begin{tabular}[c]{@{}c@{}} CPU Processing \\ Time(s)\end{tabular} \\ \midrule
Yolo + SAM2 & 0.28  & 0.36 & 0.22 & 1500    & 0.51 & 0.49 & 0.17 & 1500    \\ \hline
\rowcolor[HTML]{FFFC9E}
\textbf{IEBCS} & \textbf{0.24} & \textbf{0.30} & \textbf{0.15} & \textbf{105.3} & \textbf{0.37} & \textbf{0.39} & \textbf{0.13} & \textbf{104.6} \\ \hline
ESIM        & 0.025 & 2.2  & 0.70 & 79.8  & 0.09 & 0.24 & 0.68 & 80    \\ \hline
V2E         & 0.94  & 0.87 & 0.64 & 132   & 0.96 & 0.91 & 0.56 & 132   \\ \bottomrule 
\end{tabular}
}
\caption{Comparative evaluation of event-based simulators against the YOLOv11 + SAM2 semantic baseline for dynamic, privacy-preserving blurring. The selected method, IEBCS, achieves the best trade-off between targeted anonymization and preservation of scene detail, particularly under low-light conditions.}
\label{tab:event_sim_blur}

\end{table*}

\subsection{Motion-based Anonymization Evaluation}
\label{sec:event_sim_eval}

The quantitative results of the semantic vs. motion-based anonymization comparative study are summarized in Table~\ref{tab:event_sim_blur}. The analysis highlights distinct trade-offs between processing speed, sensitivity, and selective anonymization across simulators.

\textit{ESIM} achieved some of the highest per-object reductions (mean edge reduction $= 2.2$, SSIM reduction $= 0.70$ under bright lighting), but it successfully anonymized only $2.5\%$ of detected objects (Fraction Blurred $= 0.025$), indicating poor sensitivity. Conversely, \textit{V2E} exhibited the opposite behavior, achieving very high motion response (Fraction Blurred $> 0.94$) but excessive frame-wide blurring due to its over-sensitivity to global motion. Although the semantic \textit{YOLOv11+SAM2} baseline was computationally efficient, it suffered from semantic misdetections failing to blur fast-moving agents while incorrectly redacting static elements such as billboards.

In contrast, \textit{IEBCS} demonstrated the most balanced performance, combining robust anonymization of dynamic entities with high temporal consistency and minimal loss of scene context. While its processing time is moderately higher than the fastest simulators, this computational cost is justified by superior reliability under challenging lighting conditions and by its ability to preserve fine-grained visual cues essential for downstream \textit{VideoQA} tasks. Accordingly, \textit{IEBCS} was selected as the core anonymization component in the \textit{UDVideoQA} annotation pipeline.

\section{The Question-Answer Taxonomy}
\label{sec:qa_taxonomy}

The \textit{UDVideoQA} benchmark employs a hierarchical taxonomy designed to systematically evaluate the reasoning capabilities of \textit{VLMs} across multiple cognitive levels. This taxonomy organizes all question–answer (QA) pairs into five primary reasoning categories \textit{attribution (Atr)}, \textit{basic understanding (BU)}, \textit{event reasoning (ER)}, \textit{reverse reasoning (RR)}, and \textit{counterfactual inference (CI)} corresponding to progressively higher tiers of perceptual and inferential complexity. The design follows a human–interpretable structure that captures both perceptual grounding and causal reasoning in dynamic urban scenes.

\paragraph{\textit{Hierarchical Reasoning Structure.}}
Each \textit{QA} pair in \textit{UDVideoQA} is annotated with three hierarchical attributes:
\begin{itemize}
    \item \textbf{\textit{Reasoning Category:}} One of the five core categories -
    \textit{Atr, BU, ER, RR, and CI}
    \item \textbf{\textit{Difficulty Level:}} Each textit{QA} instance is labeled as \textit{easy}, \textit{medium}, or \textit{complex}, based on the temporal span, contextual ambiguity, and the number of entities involved.
    \item \textbf{\textit{Agent Focus:}} Questions are further categorized as \textit{pedestrian-centric} or \textit{vehicular-centric}, reflecting the primary interacting agents in the scene.
\end{itemize}

\begin{figure}
\centering
\begin{minipage}{0.95\linewidth}
\begin{tcolorbox}[
    colback=Melon!10,
    colframe=Apricot,
    boxrule=1pt,
    title={\centering \itshape Representative QA Examples from the \textit{UDVideoQA} Taxonomy},
    fonttitle=\bfseries\normalsize,
    sharp corners,
    enhanced,
    left=4pt, right=4pt, top=4pt, bottom=4pt
]
\scriptsize
\begin{lstlisting}
{
  "category": "Attribution",
  "difficulty": "Medium",
  "sub-category": "Pedestrian",
  "question": "What is the pedestrian in the red jacket doing while waiting for the light?",
  "answer": "They are looking down at their phone."
},
{
  "category": "Event Reasoning",
  "difficulty": "Complex",
  "sub-category": "Pedestrian",
  "question": "A silver car is turning right on a red light. Why does it suddenly stop before crossing the crosswalk?",
  "answer": "Because a pedestrian, who had the 'Walk' signal, stepped off the curb, and the car had to yield the right-of-way."
},
{
  "category": "Counterfactual Inference",
  "difficulty": "Complex",
  "sub-category": "Vehicular",
  "question": "If the white van had not turned left so slowly, what would the approaching black car have most likely done?",
  "answer": "The black car would likely have continued straight through the intersection without needing to brake, as its path would have been clear."
}
\end{lstlisting}
\end{tcolorbox}

\end{minipage}
\caption{Representative examples from the \textit{UDVideoQA question–answer} taxonomy.}
\label{example_Sets}
\end{figure}
\vspace{5pt}

This combination of reasoning type ($5$), difficulty tier ($3$), and agent focus ($2$) results in a total of \textit{$30$ distinct taxonomic tags}, providing fine-grained control over \textit{QA} generation and evaluation. The taxonomy enables consistent benchmarking of both perceptual understanding (e.g., object attributes) and higher-order causal reasoning.
The examples (see Figure ~\ref{example_Sets}) demonstrate how the taxonomy manifests in annotated data. Each \textit{QA} pair is stored in a structured \textit{JSON} schema with explicit reasoning labels and sub-categories, ensuring consistency across annotation and validation pipelines.
All \textit{QA} pairs were generated and validated using the \textit{UDVideoQA} annotation tool (see Section~\ref{sec:annotation_tool}). During validation, annotators confirmed question clarity, contextual relevance, and answerability based solely on video frames. This structured taxonomy ensures uniform coverage across reasoning categories and facilitates balanced performance assessment of multimodal models across perception, temporality, and causality.

\begin{figure*}[]
\centering
\begin{minipage}{0.95\linewidth}
\begin{tcolorbox}[
    colback=Melon!10,
    colframe=Apricot,
    boxrule=1pt,
    title={\centering \textit{System Prompt for \textit{VideoQGen}}},
    fonttitle=\bfseries\large,
    arc=0mm,
    left=4pt,
    right=4pt,
    top=4pt,
    bottom=4pt
]
\textbf{\textit{Role:}} You are a video questionnaire assistant for creating high-quality \textit{VideoQA} data for the \textit{UDVideoQA} benchmark. Your sole task is to watch a provided video clip and generate exactly $10$ high-quality, diverse question--answer (\textit{QA}) pairs based \textit{only} on its content.
\vspace{2pt}

\vspace{5pt}
\noindent\textbf{\textit{Core Principles (Anti-Hallucination \& Forgetting):}}
\begin{enumerate}
    \item \textbf{\textit{Video-Only Grounding:}} All questions and answers must be derived solely from the visual and audio content within the provided clip. No external knowledge is allowed.
    \item \textbf{\textit{No Ambiguity or Inference:}} Do not infer or invent unobserved details. Each \textit{QA} pair must be unambiguous and verifiable directly from the video.
    \item \textbf{\textit{Resist Hallucination:}} Counterfactual questions intentionally include false premises to test hallucination resistance. All other categories must remain strictly factual.
\end{enumerate}

\noindent\textbf{\textit{QA Generation Categories (2 QAs each):}}
\begin{enumerate}
    \item \textbf{\textit{Basic Understanding:}} This category focuses on the static, high-level context of the scene. For a given video, questions should be answerable by observing the overall setting, environment, and the presence of salient, non-dynamic objects.
    \begin{itemize}
        \item \textbf{\textit{Tags:}} \texttt{environment}, \texttt{lighting}, \texttt{weather}, \texttt{time of day}, \texttt{location type}, \texttt{road conditions}, \texttt{salient object presence}
        \item \textbf{\textit{Difficulty:}} \texttt{Easy}
        \item \textit{Example:} ``Is it raining in the video?'' - ``No''
        \item \textit{Vehicular Example:} ``What is the weather condition shown in the clip?'' - ``It is a clear, sunny day.''
    \end{itemize}
    
    \item \textbf{\textit{Attribution:}} This category targets the specific, static attributes of objects (pedestrians or vehicles) in the scene. For a given video, questions involve counting, identifying colors, makes, models, or text found on signage.
    \begin{itemize}
        \item \textbf{\textit{Tags:}} \texttt{color}, \texttt{make/model of car}, \texttt{vehicular attributes}, \texttt{pedestrian attributes}, \texttt{signage text}, \texttt{counting (static agents)}, \texttt{object identification}
        \item \textbf{\textit{Difficulty:}} \texttt{Easy-Medium}
        \item \textit{Example:} ``How many people are wearing hats?'' - ``Two''
        \item \textit{Vehicular Example:} ``What color is the lead car stopped at the traffic light?'' - ``It is a red sedan.''
    \end{itemize}

    \item \textbf{\textit{Event Reasoning:}} This category assesses the understanding of dynamic actions, agent interactions, and the temporal (forward) order of events. For a given video, questions focus on what pedestrians or vehicles *do* and in what sequence.
    \begin{itemize}
        \item \textbf{\textit{Tags:}} \texttt{actions}, \texttt{temporal order}, \texttt{agent interaction}, \texttt{pedestrian movement}, \texttt{vehicle movement}, \texttt{causality}, \texttt{event start/end}
        \item \textbf{\textit{Difficulty:}} \texttt{Medium}
        \item \textit{Example:} ``Which person walks onto the screen first?'' - ``The woman in the red coat''
        \item \textit{Vehicular Example:} ``Does the white sedan signal before turning left?'' - ``Yes, it uses its left turn signal.''
    \end{itemize}

    \item \textbf{\textit{Reverse Reasoning:}} This is a complex form of temporal reasoning. For a given video, questions require the model to recall the state of the scene or an agent *before* a specific key event occurred, testing memory of reverse temporal order.
    \begin{itemize}
        \item \textbf{\textit{Tags:}} \texttt{reverse temporal}, \texttt{agent state (prior)}, \texttt{scene state (prior)}, \texttt{pre-event context}, \texttt{agent memory}, \texttt{complex interaction chain}
        \item \textbf{\textit{Difficulty:}} \texttt{Complex}
        \item \textit{Example:} ``Just before the dog barked, what was the man holding?'' - ``A newspaper''
        \item \textit{Vehicular Example:} ``What was the pedestrian doing just before the traffic light turned green?'' - ``The pedestrian was waiting on the sidewalk.''
    \end{itemize}
    
    \item \textbf{\textit{CounterFactual Inference:}} This category tests the model's ability to resist hallucination by rejecting false premises. For a given video, the question introduces a hypothetical or factually incorrect scenario. The correct answer must first identify the premise as false and then state the correct fact.
    \begin{itemize}
        \item \textbf{\textit{Tags:}} \texttt{hypothetical}, \texttt{negation of fact}, \texttt{false premise}, \texttt{hallucination test}, \texttt{robustness}, \texttt{logical reasoning}, \texttt{factual verification}
        \item \textbf{\textit{Difficulty:}} \texttt{Complex}
        \item \textit{Example:} ``After the motorcycle ran the red light, did the police car follow it?'' - ``This is a hypothetical question. The motorcycle did not run the red light; it stopped and waited for the green light.''
    \end{itemize}
\end{enumerate}
\end{tcolorbox}
\end{minipage}
\end{figure*}

\begin{figure*}[h]
\centering
\begin{minipage}{0.95\textwidth}
\begin{tcolorbox}[
    colback=Melon!10,
    colframe=Apricot,
    boxrule=1pt,
    title={\centering \textit{Continued System Prompt...}},
    fonttitle=\bfseries\large,
   arc=0mm,
    left=4pt,
    right=4pt,
    top=4pt,
    bottom=4pt
]

\noindent\textbf{\textit{Task and Output Format:}}
\begin{enumerate}
    \item \textbf{\textit{Analyze the Clip:}} Perform a comprehensive, frame-by-frame analysis of the entire video clip.
    \item \textbf{\textit{Strict QA Quota:}} Generate \textbf{exactly $10$} \textit{QA} pairs. These must be distributed as \textit{exactly $2$ pairs for each of the $5$ categories} defined above.
    \item \textbf{\textit{Answer Fidelity:}} Answers must be concise, direct, and factually grounded in the video. Avoid conversational padding (e.g., use \textit{``Yes"} or \textit{``No"} instead of \textit{``Yes, the car does..."} or \textit{``No, it is not..."}). For counterfactual questions, the answer must follow the example format: first identify the premise as false/hypothetical, then state the correct fact.
    \item \textbf{\textit{No Speculation:}} If an attribute is ambiguous or not visible (e.g., \textit{a license plate number, a distant car's make, the number of people in a far-away bus}), you must not invent an answer. Instead, formulate a question where the ambiguity is the answer (e.g., Q: \textit{``Is the license plate on the blue car readable?"} A: \textit{``No, it is too blurry to be read."})
    \item \textbf{Output Format:} Construct a single Markdown table with the following exact columns: \texttt{Index}, \texttt{Video File Path}, \texttt{Question}, \texttt{Category}, and \texttt{Answer}.
    \begin{itemize}
        \item \texttt{Index} must be a number from $1$ to $10$,
        \texttt{Video File Path} must be the literal file path of the video clip being analyzed, and 
        \texttt{Category} must exactly match one of the $5$ category titles.
    \end{itemize}
\end{enumerate}
\end{tcolorbox}
\caption{Structured system prompt used for generating grounded and diverse question–answer pairs in the \textit{UDVideoQA} benchmark.}
\label{fig:videoqa-system-prompt}
\end{minipage}
\end{figure*}

\begin{figure*}[h]
\centering
\begin{minipage}{0.95\linewidth}
\begin{tcolorbox}[
    colback=Melon!10,
    colframe=Apricot,
    boxrule=1pt,
    title={\centering \textit{Two-Step ``Analyze-then-Generate'' Prompting Strategy}},
    fonttitle=\bfseries\large,
    arc=0mm,
    sharp corners,
    enhanced,
    left=4pt,
    right=4pt,
    top=4pt,
    bottom=4pt
]
\tcbset{
    colback=Melon!10,
    colframe=Apricot, 
    boxrule=0.5pt,
    sharp corners,
    enhanced,
    fonttitle=\bfseries, 
    left=4pt,
    right=4pt,
    top=4pt,
    bottom=4pt
}

\begin{tcolorbox}[
    colback=Melon!10,
    colframe=Apricot,
    boxrule=0.5pt,
    title={\textbf{\textit{Step 1: Factual Analysis Prompt (Chain-of-Thought Extraction)}}},
    sharp corners, 
    enhanced,
    left=4pt,
    right=4pt,
    top=4pt,
    bottom=4pt
]
\textbf{\textit{User Prompt:}} \begin{quote} \small You are a visual analysis model tasked with describing the provided video clip.

\textbf{\textit{Instructions:}}
\begin{enumerate}
    \item Observe carefully and describe only what is directly visible. 
    \item Summarize the scene context: environment, lighting, and weather conditions.
    \item Identify static elements (vehicles, pedestrians, signs, buildings) with key attributes such as color, size, or type.
    \item List up to five key dynamic events in chronological order with timestamps.
\end{enumerate}

\textbf{\textit{Constraints:}} Use only direct observations. Do not infer off-screen events.  
End with: [End of Step 1].
\end{quote}
\end{tcolorbox}
\begin{tcolorbox}[
colback=Melon!10,
colframe=Apricot, boxrule=0.5pt,title={\textbf{\textit{Step 2: Grounded QA Generation Prompt (Controlled Synthesis)}}},
    sharp corners, 
    enhanced,
    left=4pt,
    right=4pt,
    top=4pt,
    bottom=4pt
]

\textbf{\textit{User Prompt:}}
\begin{quote}
\small
Using the Step 1 factual analysis, generate exactly \textit{$10$ question--answer (QA) pairs} grounded exclusively in those observations.

\textbf{\textit{Rules:}}
\begin{enumerate}
    \item Produce 2 QA pairs from each of 5 categories:
    \item Each QA must be directly verifiable from Step 1. 
    \item Counterfactuals must explicitly correct false premises using factual details. 
    \item Answers must be concise, literal, and non-redundant.
\end{enumerate}

\textbf{\textit{Output Format:}} Present results in a Markdown table.  
End with: [End of Step 2].
\end{quote}
\end{tcolorbox}

\end{tcolorbox}
\caption{Two-stage prompting pipeline used to generate structured and visually grounded \textit{QA} pairs for the \textit{UDVideoQA} dataset.}
\label{fig:cot-prompt-strategy-robust}
\end{minipage}
\end{figure*}

\section{\textit{VideoQGen Benchmark}}
\label{sec:vqg_benchmark}

Beyond evaluating model answers, an equally important capability for multimodal systems is the ability to formulate meaningful, grounded questions about a visual scene. The \textit{VideoQGen} benchmark measures this generative ability by asking models to produce diverse, answerable questions that span our five reasoning categories (see Section~\ref{sec:qa_taxonomy}). In doing so, \textit{VideoQGen} probes whether models understand which aspects of a scene are informative, temporally relevant, and useful for downstream \textit{VideoQA} evaluation.

\begin{figure*}[t!]
  \centering
\includegraphics[width=0.75\linewidth]{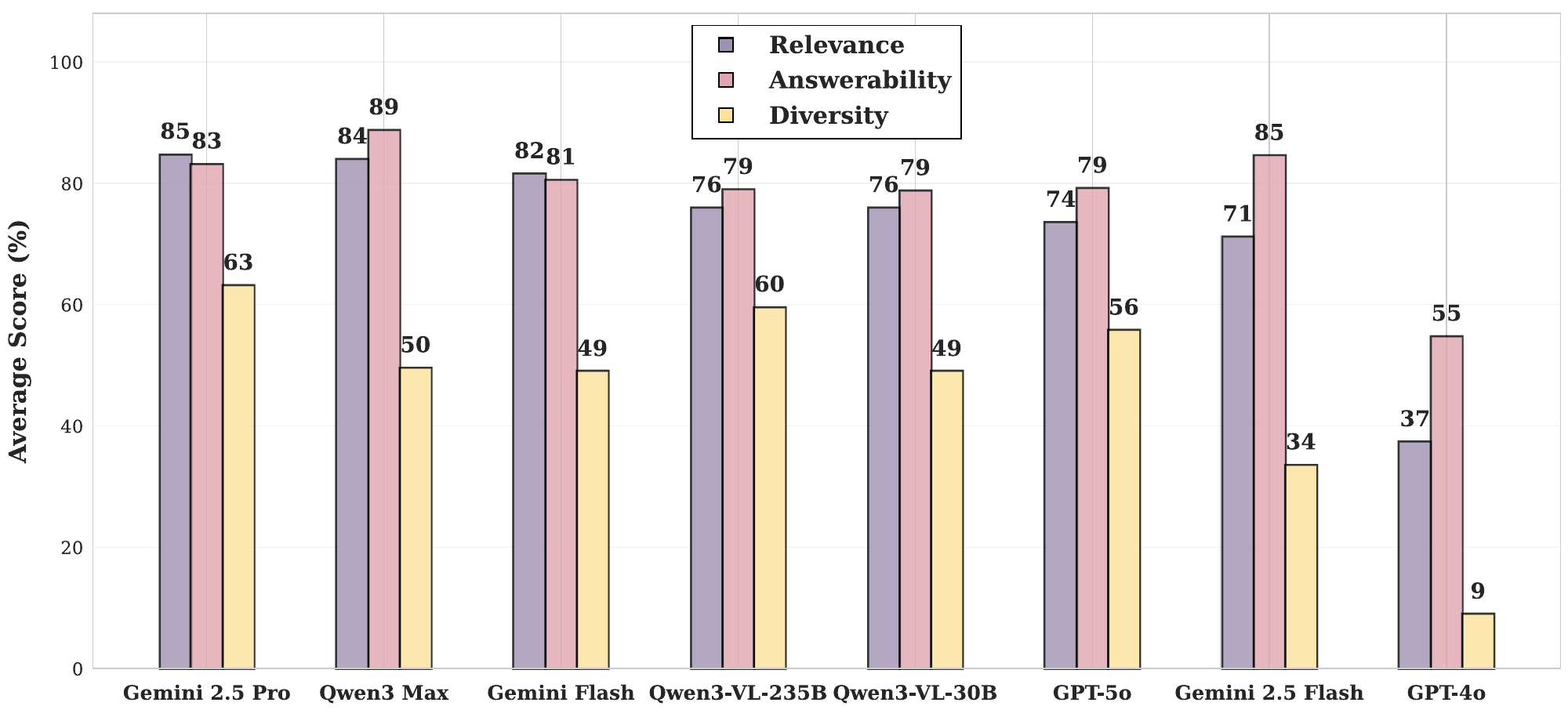}
\caption{Overall performance of \textit{VideoQGen} models across core metrics - \textit{relevance}, \textit{answerability}, and \textit{diversity}. \textit{Gemini 2.5 Pro} and \textit{Qwen3 Max} achieve the highest scores, while \textit{GPT-4o} performs consistently lower and generates less diverse questions.}
\label{vgen_barchart}
\end{figure*}

\subsection{Methodology}

\paragraph{\textit{Task Formulation and Prompting.}}
The study standardizes the generation task with a structured system prompt (see Figure~\ref {fig:videoqa-system-prompt}) that instructs the model to act as a \textit{VideoQA} data-creation assistant for the \textit{UDVideoQA} benchmark. For each $10$-second clip, the model must generate exactly $10$ question–answer pairs: two for each of the five reasoning categories \textit{(BU, Atr, ER, RR, and CI)}. Prompts require questions to be \textbf{\textit{(i)}} answerable from the clip alone, \textbf{\textit{(ii)}} succinct and unambiguous, and \textbf{\textit{(iii)}} returned as a markdown table to enforce a consistent output format. We applied a two-stage user prompt template (see Figure~\ref {fig:cot-prompt-strategy-robust}) to a curated subset of $100$ video clips and evaluated outputs from eight leading multimodal models.

\paragraph{\textit{Human Evaluation Protocol.}} 
Three human annotators scored each generated question across several capabilities (see Table~\ref{tab:vqg_qualitative_examples}):  
\begin{itemize}
  
  \item
  \textit{\textbf{Relevance:}}
  \textit{``Is the question pertinent to the visual content of the clip?"}

  \item \textit{\textbf{Answerability:}} \textit{``Can the question be answered using only the clip (no external knowledge)?" }
  
  \item 
  \textit{\textbf{Diversity:}} \textit{``Does the set of $10$ questions for a clip demonstrate linguistic and conceptual variety (penalizes repetition)?"}
  
  \item 
 \textit{ \textbf{Semantic focus:} \textit{``Each question was labeled as \textit{pedestrian} or \textit{vehicular centric} to measure agent attention?"}}
 
  \item 
  \textit{\textbf{Query complexity:}} 
  \textit{``Questions were classified as \textit{specific (grounded)} or \textit{generic}?"}
  
\end{itemize}

In alignment with the dataset workflow illustrated in the main paper, this evaluation was embedded within a \textit{human-in-the-loop} quality control process. Annotators not only provided numerical ratings but also flagged questions exhibiting ambiguity, hallucination, or poor grounding for refinement. These flagged cases were reviewed during the \textit{categorical verification} and \textit{ground-truth validation} stages shown in the pipeline. The iterative review ensured that only semantically valid and contextually grounded \textit{QA} pairs were retained in the final benchmark.  
To ensure annotation reliability, we computed Cronbach's $\alpha$, yielding coefficients of 0.73 (Relevance), 0.74 (Answerability), and 0.79 (Diversity). A generated question was retained only if all annotator pairs demonstrated positive correlation ($r \geq 0$) on this dimension. This consensus-based filtering discarded inconsistent items, resulting in an 89\% retention rate for the final benchmark.

\begin{table*}[t!]
\centering
\resizebox{0.95\textwidth}{!}{%
\begin{tabular}{c|ccc}
\toprule
\textbf{Model} &
  \multicolumn{3}{c}{\textbf{Questions}} \\ \cline{2-4} 
 &
  \multicolumn{1}{c|}{\textbf{Low Intensity Light}} &
  \multicolumn{1}{c|}{\textbf{Medium  Intensity Light}} &
  \textbf{High Intensity Light} \\ \midrule
\textbf{Gemini 2.5 Pro} &
  \multicolumn{1}{c|}{\cellcolor[HTML]{A8D08D}\begin{tabular}[c]{@{}c@{}}Are there any discernible architectural features on the \\ buildings flanking the street?\end{tabular}} &
  \multicolumn{1}{c|}{\cellcolor[HTML]{A8D08D}\begin{tabular}[c]{@{}c@{}}What was the pedestrian doing as the dark car was \\ approaching the crosswalk to make its turn?\end{tabular}} &
  \cellcolor[HTML]{A8D08D}\begin{tabular}[c]{@{}c@{}}Before the pedestrian in the dark shirt begins crossing \\ the street at 0:07, what is the state of the intersection?\end{tabular} \\ \hline
\textbf{Qwen3 Max} &
  \multicolumn{1}{c|}{\cellcolor[HTML]{C6E0B4}\begin{tabular}[c]{@{}c@{}}Are there any temporary construction barriers or signs \\ present on the sidewalk near the right edge of the frame?\end{tabular}} &
  \multicolumn{1}{c|}{\cellcolor[HTML]{A8D08D}\begin{tabular}[c]{@{}c@{}}If the traffic light for the horizontal street had remained red,\\  would the black sedan have entered the intersection at $0.6$ seconds?\end{tabular}} &
  \cellcolor[HTML]{A8D08D}\begin{tabular}[c]{@{}c@{}}If the white van had begun to turn right at $0:02$, which \\ pedestrian group would it have encountered?\end{tabular} \\ \hline
\textbf{Gemini Flash} &
  \multicolumn{1}{c|}{\cellcolor[HTML]{C6E0B4}\begin{tabular}[c]{@{}c@{}}Counting all visible vehicles, both moving and \\ stationary, how many are white?\end{tabular}} &
  \multicolumn{1}{c|}{\cellcolor[HTML]{E2F0D9}\begin{tabular}[c]{@{}c@{}}Is it true that a cyclist uses the green-painted bike lane at any point \\ in the video?\end{tabular}} &
  \cellcolor[HTML]{C6E0B4}\begin{tabular}[c]{@{}c@{}}A person in a white shirt and dark pants enters from the \\ bottom left. What path do they take?\end{tabular} \\ \hline
\textbf{Qwen3-VL-23B} &
  \multicolumn{1}{c|}{\cellcolor[HTML]{E2F0D9}\begin{tabular}[c]{@{}c@{}}Was there any person standing on the sidewalk near the\\  orange traffic signs on the bottom right?\end{tabular}} &
  \multicolumn{1}{c|}{\cellcolor[HTML]{C6E0B4}\begin{tabular}[c]{@{}c@{}}If the silver sedan had not entered the intersection at $0.3$s, would \\ the white autonomous vehicle have been blocked from turning left?\end{tabular}} &
  \cellcolor[HTML]{C6E0B4}\begin{tabular}[c]{@{}c@{}}If the traffic light for the crosswalk had turned red at $5$ seconds, \\ would the pedestrians have been in the middle of the street?\end{tabular} \\ \hline
\textbf{Gemini 2.5 Flash} &
  \multicolumn{1}{c|}{\cellcolor[HTML]{E2F0D9}\begin{tabular}[c]{@{}c@{}}Is it true that any pedestrian attempted to cross the street\\  without using a designated crosswalk?\end{tabular}} &
  \multicolumn{1}{c|}{\cellcolor[HTML]{C6E0B4}\begin{tabular}[c]{@{}c@{}}If the pedestrian at the bottom right had decided to cross at 0:03\\ , would they have entered the silver car's path?\end{tabular}} &
  \cellcolor[HTML]{E2F0D9}\begin{tabular}[c]{@{}c@{}}How many silver/grey passenger cars are moving away from the \\ camera in the upper lanes?\end{tabular} \\ \hline
\textbf{Qwen3-VL-30B} &
  \multicolumn{1}{c|}{\cellcolor[HTML]{FFD580}Is the traffic light for the main road green?} &
  \multicolumn{1}{c|}{\cellcolor[HTML]{FFD580}\begin{tabular}[c]{@{}c@{}}What is the color of the traffic light for vehicles going\\  straight through the intersection?\end{tabular}} &
  \cellcolor[HTML]{E2F0D9}\begin{tabular}[c]{@{}c@{}}Would the silver car have been able to pass the white van if the \\ van had not stopped?\end{tabular} \\ \hline
\textbf{GPT-5} &
  \multicolumn{1}{c|}{\cellcolor[HTML]{FFD580}How many vehicles are visible between 0 and 2 s?} &
  \multicolumn{1}{c|}{\cellcolor[HTML]{FFD580}\begin{tabular}[c]{@{}c@{}}If the signal had stayed red for the entire clip, would any\\  vehicle have entered the intersection?\end{tabular}} &
  \cellcolor[HTML]{FFCCC9}Is the weather condition clear and dry? \\ \hline
\textbf{GPT-4o} &
  \multicolumn{1}{c|}{\cellcolor[HTML]{FFCCC9}Is the scene recorded during daylight or nighttime?} &
  \multicolumn{1}{c|}{\cellcolor[HTML]{FFCCC9}\begin{tabular}[c]{@{}c@{}}Would the vehicle have proceeded if the pedestrian hadn't \\ entered the crosswalk?\end{tabular}} &
  \cellcolor[HTML]{FFCCC9}How many pedestrians are visible during the clip? \\ \bottomrule
\end{tabular}%
}
\caption{Representative examples of generated questions across lighting conditions (low, medium, and high intensity). \textit{Gemini 2.5 Pro} achieves the highest reasoning quality, while \textit{GPT-4o} produces repetitive, low-diversity questions.}
\label{tab:vqg_qualitative_examples}
\end{table*}

\subsection{Results and Analysis}

\paragraph{\textit{Overall Performance.}}
\textit{Gemini 2.5 Pro} and \textit{Qwen3 Max} emerge as the top-performing models, achieving the highest scores in \textit{relevance} and \textit{answerability}, frequently exceeding $80$\% across reasoning categories as depicted in Figure~\ref{vgen_barchart}. These results indicate a strong capacity to generate questions that are both faithful to the video content and consistent with the prompt’s structural constraints. The \textit{Qwen} family demonstrates remarkable stability across scales, with the lighter $30$B variant achieving near-parity with its larger counterpart.

\paragraph{\textit{The Diversity Challenge}}
Across all models, \textit{diversity} lags behind other metrics. Even the leading \textit{Gemini 2.5 Pro} attains only low ($60$\%) diversity scores, a significant gap compared to its ($80$\%+) relevance and answerability. Other models exhibit sharper declines, with \textit{GPT-4o} averaging as low as $7$–$11$\% (see Figure~\ref{vgen_barchart}). This pattern suggests that while current models follow structured prompts effectively, they often resort to formulaic phrasing or repetitive semantics, limiting linguistic and conceptual variation.

\begin{figure}[t]
\centering
\includegraphics[width=1\linewidth]{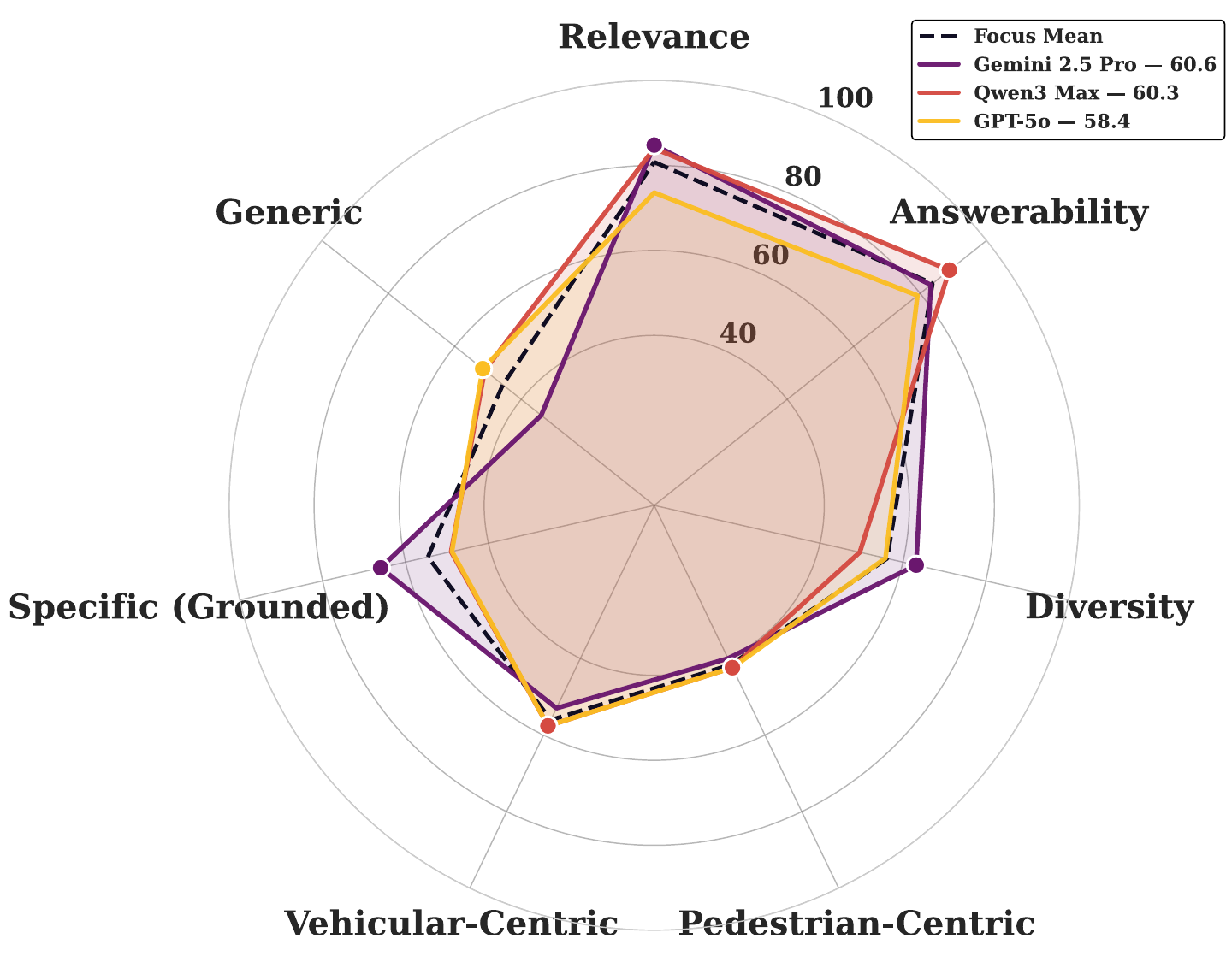}
\caption{Distinct performance profiles of \textit{VideoQGen} models across evaluation metrics. \textit{Gemini 2.5 Pro} excels in \textit{specific (grounded)} and \textit{relevance}, while \textit{Qwen3 Max} leads in \textit{answerability}. The \textit{focus mean} serves as a baseline for comparison.}

\label{fig:vqa_radial_graph} 
\end{figure}

\paragraph{\textit{Specificity and Grounding.}}
Encouragingly, the best-performing models generate a high proportion of \textit{specific (grounded)} questions, demonstrating a clear understanding of fine-grained spatio-temporal cues. As shown in Figure~\ref{fig:vqa_radial_graph}, \textit{Gemini 2.5 Pro} exhibits strong performance in \textit{relevance} and \textit{grounding}, while \textit{Qwen3 Max} dominates in \textit{answerability}. These findings confirm that structured prompting enables modern \textit{VLMs} to produce contextually rich, non-trivial questions that extend beyond surface-level object descriptions.

\paragraph{\textit{Failure modes.}}
The \textit{GPT} series, particularly \textit{GPT-4o}, performs inconsistently on this task. Despite robust general multimodal capabilities, it underperforms in structured generation, yielding low diversity and relevance alongside a higher share of \textit{generic} questions. This behavior highlights a mismatch between general reasoning proficiency and the disciplined reasoning required to produce grounded, logically coherent video-based questions.

\paragraph{\textit{Qualitative Insights.}}
Table~\ref{tab:vqg_qualitative_examples} presents qualitative samples across models and lighting conditions, illustrating clear differences in reasoning depth.  
High-performing models such as \textit{Gemini 2.5 Pro} and \textit{Qwen3 Max} (green cells) consistently formulate multi-step reasoning questions, e.g.,  
\textit{“Before the pedestrian begins crossing, what was the state of the intersection?” (reverse reasoning)} or  \textit{
“If the traffic light had remained red, would the black sedan have entered?” (counterfactual reasoning).}  
They also generate temporally anchored queries such as  
\textit{“What was the pedestrian doing as the dark car approached the crosswalk?”}, capturing fine-grained agent interactions even under low-light conditions.
  
\begin{figure*}[t]
\centering

\begin{subfigure}{\textwidth}
    \centering
    \includegraphics[width=\textwidth]{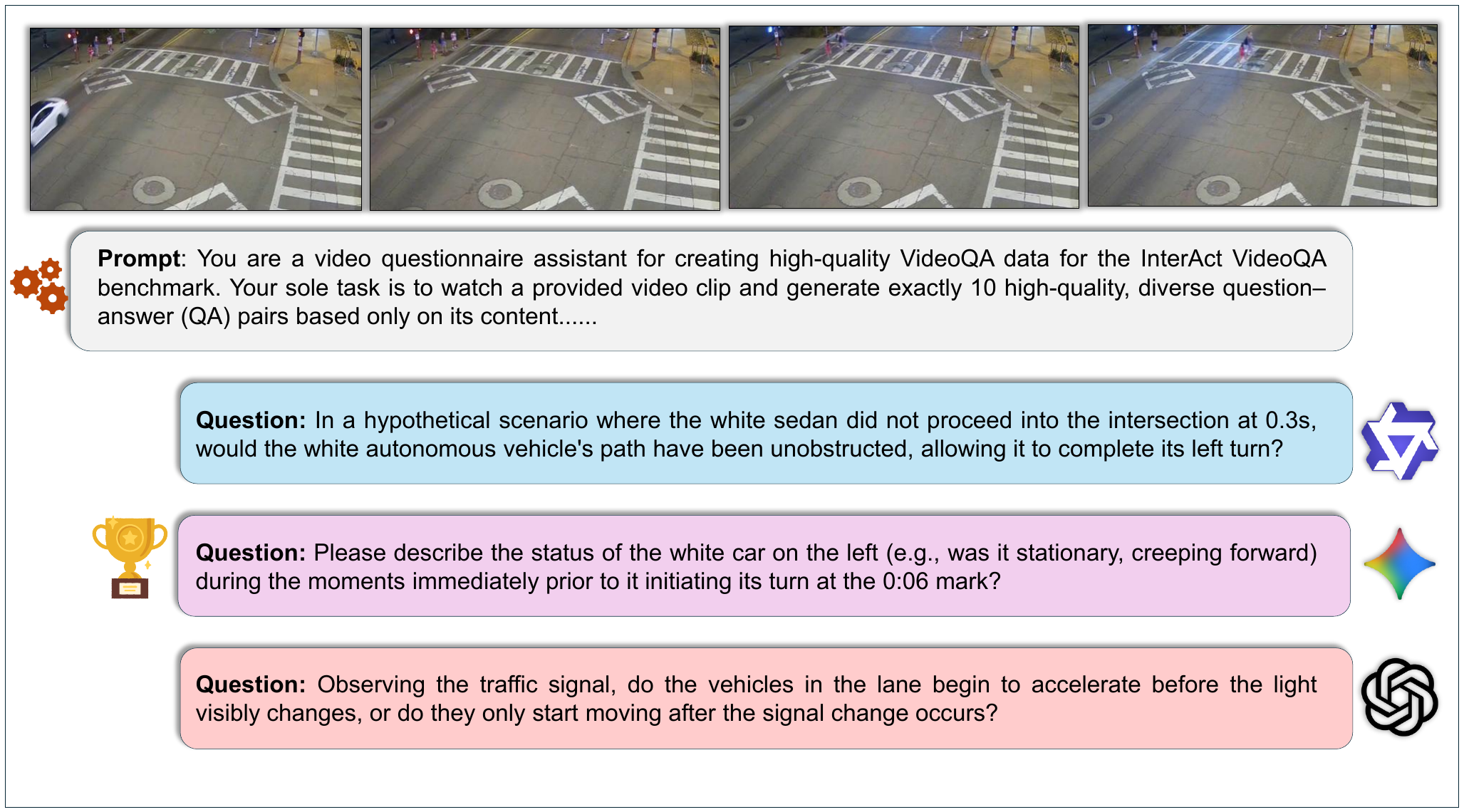}
    \caption{Night-time clips}
    \label{fig:video_qgen_night}
\end{subfigure}

\vspace{1em}

\begin{subfigure}{\textwidth}
    \centering
    \includegraphics[width=\textwidth]{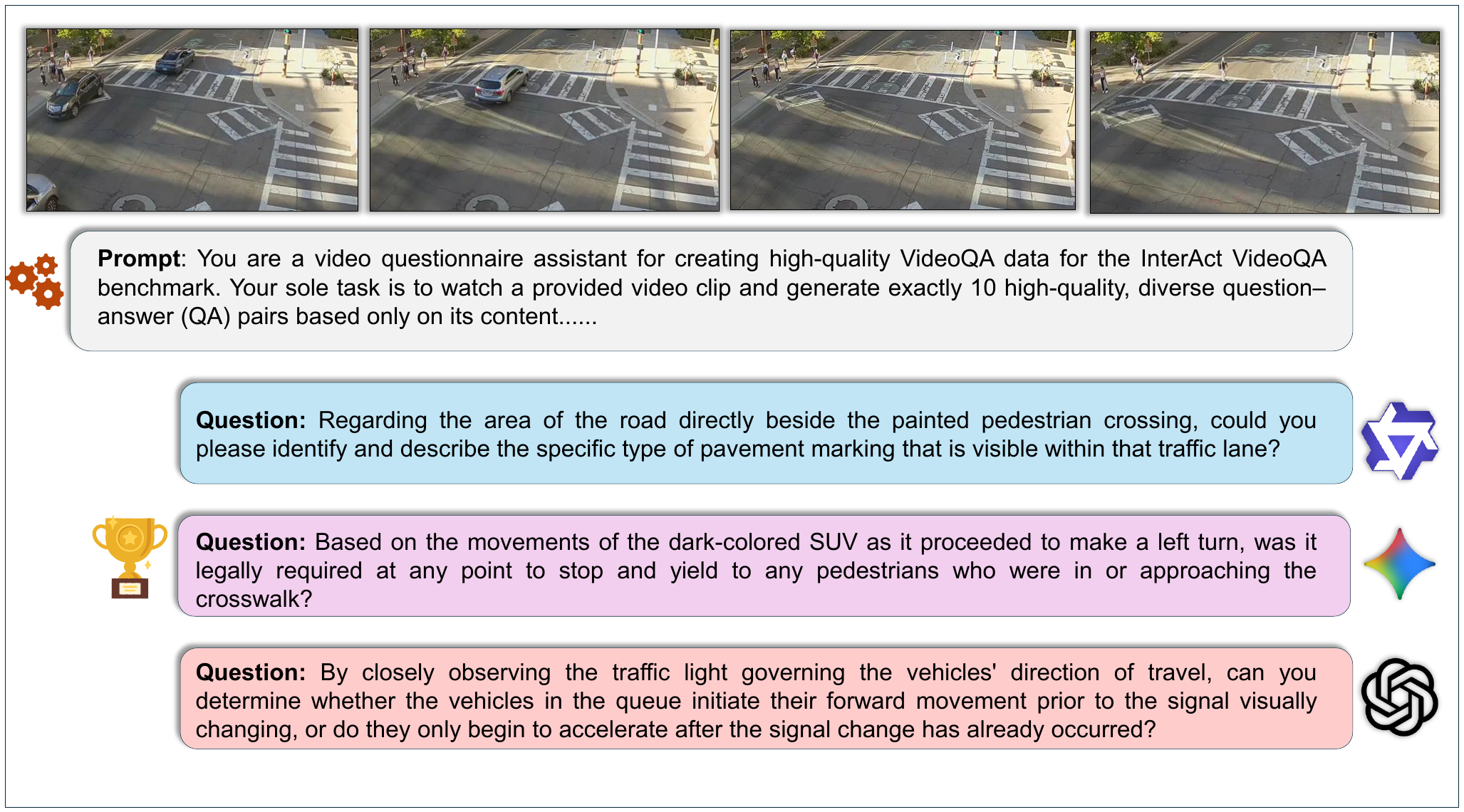}
    \caption{Early-hours clips}
    \label{fig:vqgen_day_night}
\end{subfigure}

\caption{Example \textit{VideoQGen} outputs across lighting conditions.}
\label{fig:vqgen_combined}
\end{figure*}

In contrast, weaker models like \textit{GPT-4o} and \textit{GPT-5} (red cells) default to simplistic or generic queries, such as weather descriptions or basic counting (\textit{“Is the weather clear and dry?”}), and (\textit{“How many pedestrians are visible?”}). These outputs reflect a failure to engage with complex, multi-agent dynamics, precisely the reasoning dimension the benchmark is designed to evaluate.
Together, these findings confirm that \textit{VideoQGen} effectively differentiates between models that exhibit deep spatio-temporal reasoning and those limited to surface-level perception. Structured prompting enables models to produce semantically rich, logically grounded questions, but genuine linguistic diversity remains a bottleneck. The results underscore that while modern \textit{VLMs} are capable of generating high-quality questions when guided precisely. The figures presented below show a few examples of \textit{VideoQGen} (see Figure \ref{fig:video_qgen_night} and Figure \ref{fig:vqgen_day_night} for night and day, respectively).

\section{The \textit{UDVideoQA} Annotation Tool}
\label{sec:annotation_tool}

\begin{figure}[t]
  \centering
\includegraphics[width=0.98\linewidth]{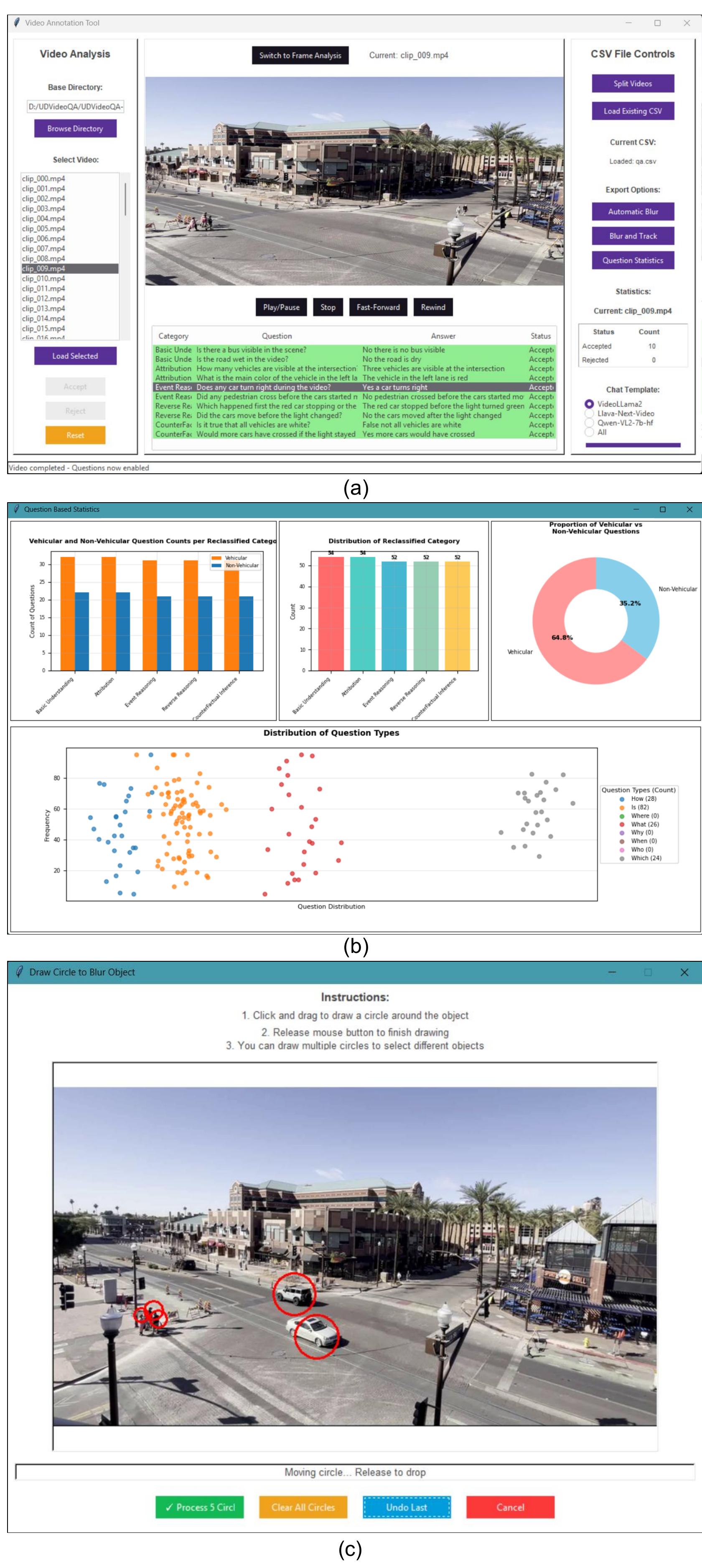}
\caption{User interface of the \textit{UDVideoQA} Annotation Tool showing (top) the primary manual annotation workspace, (middle) real-time statistics for \textit{QA} pairs, and (bottom) fine-grained blur tagging using YOLOv11 and SAM2.}
\label{Annotation_Tool}
\end{figure}

The \textit{UDVideoQA} is a comprehensive, semi-automated application designed to streamline the end-to-end creation of the \textit{VideoQA} dataset. It consolidates the entire workflow from raw video ingestion and anonymization to \textit{QA} validation and model-ready export within a single, unified framework (as in Figure~\ref{Annotation_Tool}). Developed in Python using the \texttt{Tkinter} framework and the \texttt{ttkbootstrap} library, the tool features a modern, responsive interface optimized for long annotation sessions and cross-platform deployment. The architecture ensures efficient dataset curation while maintaining strict ethical and quality-control standards, as it follows a tri-panel design philosophy for intuitive workflow management:  
\textbf{\textit{(1)}} the \textit{data navigation panel} manages input sources,  
\textbf{\textit{(2)}} the \textit{primary annotation workspace} provides interactive playback and \textit{QA} validation, and  
\textbf{\textit{(3)}} the \textit{processing and export panel} handles anonymization, batch processing, and data serialization.  
This logical structure guides annotators through clearly defined stages of data ingestion, validation, and export.

\subsection{Data Ingestion and Pre-processing}
Handling raw, long-form surveillance footage poses significant challenges for annotation scalability. To address this, the tool integrates an automated video segmentation utility that converts source videos into standardized $10$s clips. The user specifies input and output directories, and the tool processes all major video formats (e.g., \texttt{.mp4}, \texttt{.avi}, \texttt{.mov}, \texttt{.mkv}), preserving resolution and codec fidelity. Each clip is automatically named sequentially (e.g., \texttt{clip\_001.mp4}) to maintain dataset organization.

Once segmented, clips can be loaded directly into the \textit{primary annotation workspace}, which supports two analysis modes:  
\textit{video playback mode} for temporal reasoning and causal interpretation, and  
\textit{frame analysis mode} for fine-grained spatial inspection.  
Playback controls (play/pause, stop, frame-step, and \textit{fps-aware} $10$s skips) ensure efficient navigation. To enforce contextual awareness, the tool automatically locks the \textit{QA} validation list until the entire clip is reviewed, prompting annotators to observe the full temporal context before annotation.

The initial workflow stage addresses the challenge of handling raw, long-form surveillance footage. The tool integrates an automated segmentation utility that allows an annotator to select a source directory and an output directory. The system then processes all supported video formats (e.g., \texttt{.mp4}, \texttt{.avi}, \texttt{.mov}, \texttt{.mkv}) and segments them into standardized 10-second clips, preserving the original resolution and codec while sequentially naming the output (e.g., \texttt{clip\_001.mp4}).
Once segmented, clips are loaded into the primary annotation workspace. This module supports two distinct analysis modes: a \textit{}{video playback mode} for assessing temporal context and causal reasoning, and a \textit{frame analysis mode} for fine-grained, frame-by-frame spatial inspection. Standard media controls (play/pause, stop, and FPS-aware $10$-second skips) are provided. A crucial quality control gate is implemented: upon loading a new video, the \textit{QA} validation list remains locked, and a status message indicates ``Auto-playing video", guaranteeing that the annotator reviews the entire visual context at least once before beginning annotation.

\subsection{Human-in-the-Loop Anonymization Workflow}
A key innovation of \textit{UDVideoQA} tool is its two-stage \textit{hybrid anonymization pipeline}, which integrates automated motion-based blurring with AI-assisted manual refinement to ensure both efficiency and ethics are maintained.

\paragraph{\textit{Stage 1: Automated Motion-Based Blurring.}}
In the first stage, the \textit{IEBCS}-based motion-blur module anonymizes all dynamic agents in a single operation. Operating on pixel-intensity changes, this event-driven method effectively protects moving subjects while avoiding the semantic misclassifications typical of detector-based approaches. It serves as a high-recall, low-latency first pass for anonymization.

\paragraph{\textit{Stage 2: Manual AI-Assisted Track-and-Blur.}}
For fine-grained correction, the \textit{Manual Track-and-Blur} feature combines \textit{YOLOv11} for detection and the \textit{SAM2} for pixel-accurate masking. Annotators can draw a simple region of interest around a target object, prompting the system to generate an initial \textit{SAM} mask and predict its motion trajectory using velocity vectors. A \textit{``predict–update”} cycle then maintains smooth tracking: the predicted position is updated each frame via \textit{YOLO} detections, with masks regenerated only when object shape or trajectory deviates significantly. This approach maintains high temporal precision while minimizing computational overhead, even in cases of occlusion or low light.

\begin{figure}
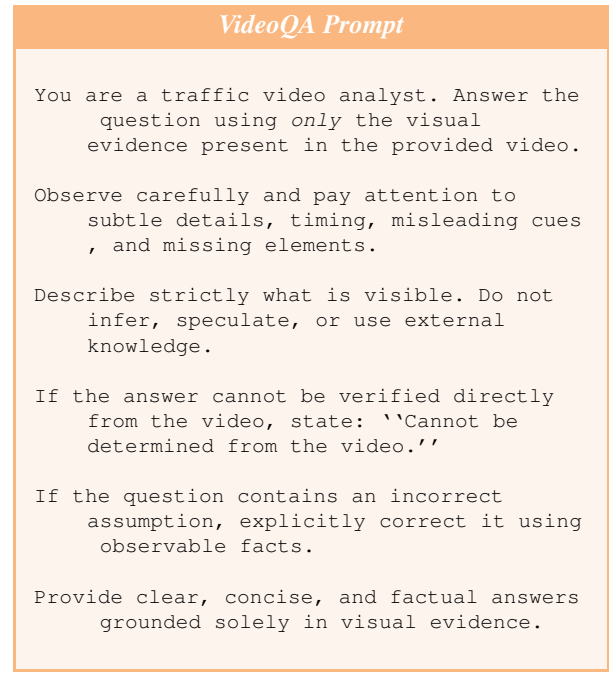

\centering
\begin{minipage}{0.95\linewidth}
\begin{tcolorbox}[
    colback=Melon!10,
    colframe=Apricot,
    boxrule=1pt,
    title={\centering \itshape VideoQA Prompt},
    fonttitle=\bfseries\normalsize,
    sharp corners,
    enhanced,
    left=4pt, right=4pt, top=4pt, bottom=4pt
]
\scriptsize
\begin{lstlisting}
You are a traffic video analyst. Answer the question using  (*@\textit{only}@*) the visual evidence present in the provided video.

Observe carefully and pay attention to subtle details, timing, misleading cues, and missing elements.

Describe strictly what is visible. Do not infer, speculate, or use external knowledge.

If the answer cannot be verified directly from the video, state: ``Cannot be determined from the video.''

If the question contains an incorrect assumption, explicitly correct it using observable facts.

Provide clear, concise, and factual answers grounded solely in visual evidence.
\end{lstlisting}

\end{tcolorbox}
\end{minipage}
\caption{Prompt template used for \textit{VideoQA} model evaluation, instructing models to generate factual, visually grounded answers based solely on the provided video content.}
\label{fig:videoqa_prompt}
\end{figure}

\subsection{QA Validation and Refinement Module}

Beyond anonymization, \textit{UDVideoQA} tool facilitates rigorous manual validation and refinement of the \textit{QA} pairs generated by large language models. \textit{QA} pairs are imported from \textit{CSV} format and displayed in an interactive \texttt{ttk.Treeview} table with columns for category, question, answer, and validation \textit{status} (\textit{none}, \textit{accepted}, \textit{rejected}).  
The validation interface supports keyboard shortcuts (\texttt{Enter} to accept, \texttt{Backspace} to reject) for rapid annotation throughput.
Rejected questions are not discarded outright. Instead, a double-click opens an editable dialog where annotators can modify the question, answer, or category. Upon correction, the record automatically updates to \textit{``accepted''}, turning the validation stage into an active refinement loop. Annotators may save incremental progress for batches of videos before export, ensuring recovery and continuity.

\subsection{Integrated Analytics Dashboard}
To aid quality monitoring, \textit{UDVideoQA} tool features an integrated \textit{analytics dashboard} that provides live statistics on annotation progress, \textit{QA} category distribution, and validation ratios. The dashboard supports categorical verification, hallucination detection, and timestamp consistency checks, aligning with the \textit{human-in-the-loop} refinement shown in the main pipeline (see Figure~\ref{Annotation_Tool}). This feedback loop ensures annotation consistency and enables transparent reporting of validation coverage during dataset curation.

\subsection{Model-Ready Benchmark Export}
The final stage of the workflow serializes validated \textit{QA} data into standardized, model-compatible formats. Only entries marked as \textit{accepted} are exported, while rejected or pending items remain archived for traceability. The export module automatically generates \textit{JSONL} chat templates optimized for major \textit{VideoLM} frameworks, including \textit{VideoLLaMA}, \textit{LLaVA-NeXT-Video}, and \textit{Qwen-VL}.  
Built-in profiles ensure proper formatting of conversation turns and metadata fields (e.g., \textit{timestamps, reasoning type}). The export process includes robust quality checks for missing values and generates timestamped \textit{CSV} backups to preserve annotation states before serialization. This standardized export pipeline allows seamless benchmarking and reproducibility across different multimodal architectures.

\begin{figure*}[t]
\centering
\begin{minipage}{0.95\linewidth}
\begin{tcolorbox}[
    colback=Melon!10,
    colframe=Apricot,
    boxrule=1pt,
    title={\centering \textit{  Scoring Policy for the LLM Judge as Prompt for Evaluation}},
    fonttitle=\bfseries\large,
    sharp corners,
    enhanced
]
\textbf{\textit{Role:}} You are an impartial, strict grader for \textit{VideoQA} focused on semantic-semantic understanding.

\vspace{3pt}
\noindent\textbf{\textit{Input Format:}}
\begin{itemize}
    \item CSV columns: \texttt{index}, \texttt{video\_file\_path}, \texttt{question}, \texttt{category}, \texttt{answer}, \texttt{model\_answer}.
    \item Read the CSV
    \item Normalize Unicode and smart quotes so text compares correctly.
\end{itemize}

\vspace{3pt}
\noindent\textbf{\textit{Derived Fields (per row):}}
\begin{itemize}
    \item \textbf{\textit{clip\_id:}} \texttt{video\_file\_path} without extension.
    \item \textbf{\textit{q\_idx:}} Question index within each \texttt{clip\_id}, in pasted order, starting at 1.
    \item \textbf{\textit{subtype:}} Determined from the lowercased question text:
    \begin{itemize}
        \item \textbf{\textit{pedestrian:}} if mentions \texttt{pedestrian, pedestrians, person, people, man, woman, child, cyclist, bicyclist, walker}.
        \item \textbf{\textit{vehicular:}} if mentions \texttt{car, cars, vehicle, vehicles, truck, bus, van, sedan, suv, motorcycle, lane, traffic, driving, overtake, approach, foreground}.
        \item otherwise, \textbf{\textit{background}}.
    \end{itemize}
    \item \textbf{\textit{category:}} Use given text (trim whitespace). Canonicalize close variants (e.g., \texttt{CounterFActual Inference}) to one of:
    \textit{Basic Understanding, Attribution, Event Reasoning, Reverse Reasoning, Counterfactual Inference.}
    \item \textbf{\textit{ground\_truth:}} Complete answer 
    \item \textbf{\textit{model\_answer:}} Trimmed text (empty if blank or NA).
\end{itemize}

\vspace{3pt}
\noindent\textbf{\textit{Grading Logic:}}
\begin{itemize}
    \item \textbf{\textit{correct}} $\in \{1, 0\}$ (no partial credit).
    \item \textbf{\textit{Binary/categorical:}} accept exact or unambiguous semantic equivalence (case-insensitive; e.g., ``yes.'' = ``yes'').
    \item \textbf{\textit{short textual:}} accept meaning-identical paraphrases; if missing key info or contradicts ground truth $\rightarrow$ 0.
    \item \textbf{\textit{If}} \texttt{model\_answer} is blank, ``i don't know,'' ``unsure,'' or contradictory $\rightarrow$ 0.
    \item Do not infer beyond what is written in the ground truth.
\end{itemize}

\vspace{3pt}
\noindent\textbf{\textit{Weights by Category:}}
\begin{center}
\begin{tabular}{ll} 
\textbf{Category} & \textbf{Weight} \\
\toprule
Basic Understanding & 1.0 \\
Attribution & 1.2 \\
Event Reasoning & 1.3 \\
Reverse Reasoning & 1.3 \\
Counterfactual Inference & 1.5 \\
\bottomrule
\end{tabular}
\end{center}

\vspace{3pt}
\noindent\textbf{\textit{Scoring:}}
\[
\text{points} = \text{correct} \times \text{weight}
\]
\end{tcolorbox}
\caption{Grading prompt used for \textit{VideoQA} benchmark evaluation. The LLM-based grader enforces strict, non-inferential correctness through canonicalized categories, subtype detection, and category-weighted scoring.
}
\label{fig:videoqa-grading-prompt}
\end{minipage}
\end{figure*}

\section{Experiments}
This section describes our experimental protocol, evaluation methodology, quantitative benchmark results, and qualitative error analysis on the \textit{UDVideoQA} dataset. Our goals are \textbf{\textit{(i)}} to measure zero-shot reasoning capabilities of SOTA multimodal models, \textbf{\textit{(ii)}} to evaluate gains achievable via parameter-efficient fine-tuning, and \textbf{\textit{(iii)}} to diagnose systematic failure modes that limit current \textit{VideoQA} performance.
  
\begin{figure*}[t]
\centering

\includegraphics[width=0.95\textwidth]{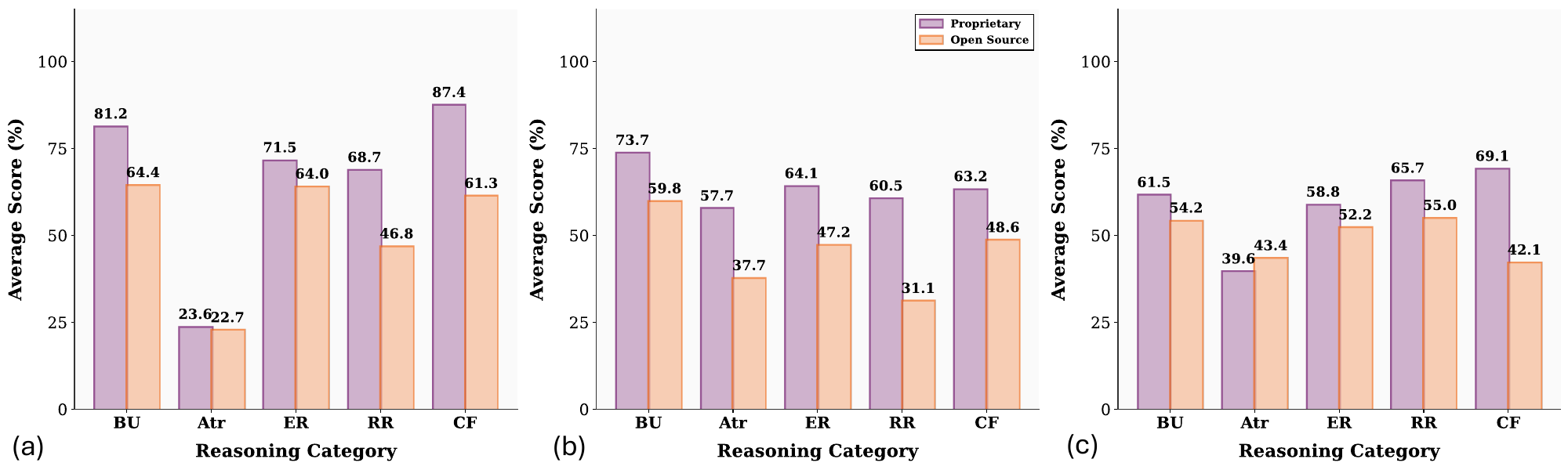}
\caption{Average \textit{VideoQA} performance of proprietary and open-source models across five reasoning categories under different lighting conditions: (a) day, (b) afternoon, and (c) nighttime scenarios.}
\label{fig:vqa_barchart_graph} 
\end{figure*}

\subsection{Experimental Setup}
\label{sec:exp_setup}

\paragraph{\textit{Zero-shot Evaluation.}} All candidate models are first tested directly on the \textit{UDVideoQA} test set to measure out-of-the-box generalization and reasoning capabilities without any domain-specific tuning. Each model was given the same \textit{VideoQA} prompt (see Figure~\ref{fig:videoqa_prompt}) to ensure consistency 

\paragraph{\textit{Fine-tuning (parameter-efficient).}} Selected open-source models were fine-tuned on the \textit{UDVideoQA} training split to quantify adaptation benefits. Fine-tuning configurations include:

\begin{itemize}
  \item \textbf{\textit{InternVL-38B:}} LoRA adapters with rank $r=16$ and $\alpha=32$ were applied. The core LLM, vision backbone, and MLP layers were frozen (\texttt{--freeze\_llm}, \texttt{--freeze\_backbone}, and \texttt{--freeze\_mlp}) so only adapter parameters were trained. Jobs were run via SLURM (\texttt{srun}) or \texttt{torchrun} on multi-GPU nodes.
  \item \textbf{\textit{Qwen-VL (32B):}} LoRA with rank $r=64$, main components frozen and only LoRA adapters trained for efficiency.
  \item \textbf{\textit{Qwen-VL (7B):}} The vision tower was unfrozen and trained (no LoRA) to better adapt the small model’s visual encoder. Training used 4×A100 GPUs.
\end{itemize}

All fine-tuning runs used standard data augmentation and early stopping tuned on a held-out validation subset. Hyperparameter sweeps and compute budgets were kept consistent across models to ensure fair comparisons.

\subsection{Evaluation Methodology}
\label{sec:eval_method}

To avoid brittle lexical matches and better assess semantic correctness and reasoning complexity, we adopt a multi-stage, hybrid evaluation:

\paragraph{\textit{LLM-based Semantic Scoring (LLM Judge).}}
We use an external LLM judge (Gemini 2.5 Pro) driven by the \textit{Strict Grading System Prompt} (see Figure~\ref{fig:videoqa-grading-prompt}) to perform semantic–semantic matching. The judge assigns a binary score ($\text{correct}\in\{0,1\}$) after normalizing text, canonicalizing categories, and assessing semantic equivalence rather than raw string overlap. The grader rejects blank, non-committal, or contradictory responses \cite{Marcu2024LingoQA}.

\paragraph{\textit{Weighted Complexity-based Scoring.}}
To reflect varying cognitive load across question types, we weight categories by difficulty: \textit{basic understanding} ($1.0$), \textit{attribution} ($1.2$), \textit{event reasoning} ($1.3$), \textit{reverse reasoning} ($1.3$), and \textit{counterfactual inference} ($1.5$). The final per-question points equal the binary correctness times the category weight, and model scores are aggregated over the evaluation set.

\paragraph{\textit{Human validation.}}
To calibrate and control for potential \textit{LLM} judge biases, we randomly sample a \textit{``gold-standard''} subset ($1$K predictions) that is independently annotated by domain experts. An agreement between the human labels and the \textit{LLM} judge is reported. Disagreement cases were inspected and used to refine the grading prompt and normalization heuristics.

\begin{figure*}[t]
\centering
\includegraphics[width=0.90\textwidth]{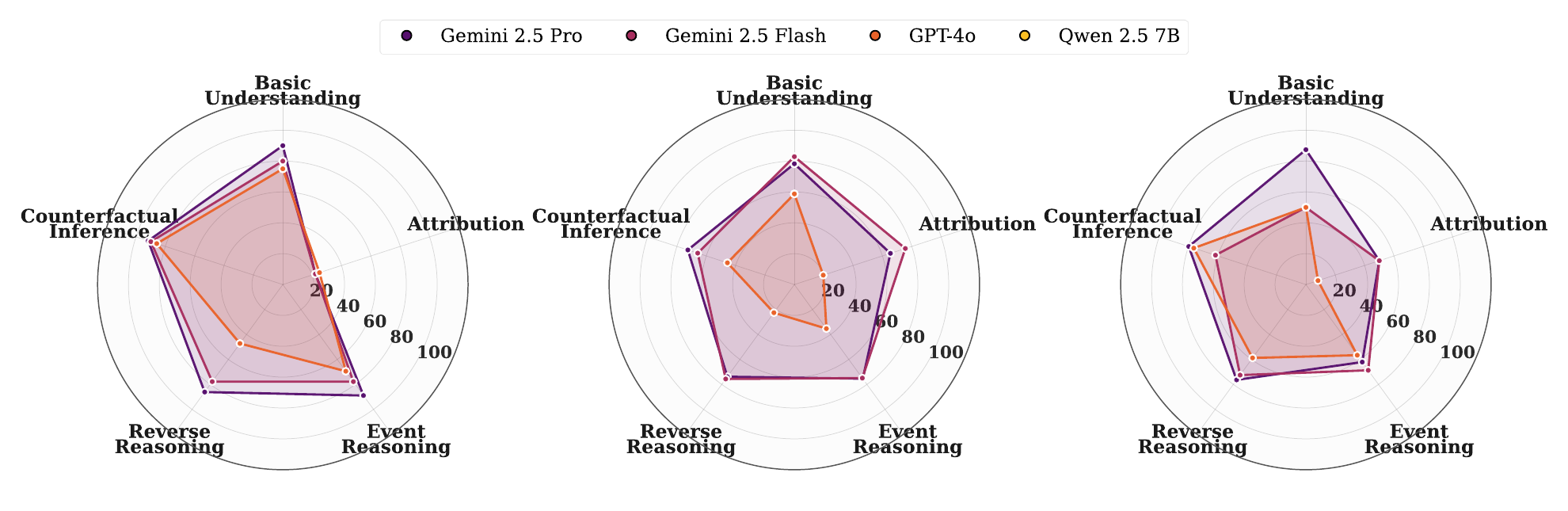}
\caption{Comparison of four leading multimodal models (\textit{Gemini 2.5 Pro}, \textit{Gemini 2.5 Flash}, \textit{GPT-4o}, and \textit{Qwen 2.5 7B}) across five \textit{VideoQA} reasoning dimensions, highlighting variations in perceptual and causal understanding.}
\label{fig:vqa_radar_chart} 
\end{figure*}

\begin{table*}[t]
\centering

\label{tab:others_vehicular_color}
\resizebox{1\linewidth}{!}{%
\begin{tabular}{ll|cccccc|cccccc}
\toprule
\multirow{2}{*}{\textbf{Model Type}} & \multirow{2}{*}{\textbf{Model Name}} &
\multicolumn{6}{c|}{\textbf{Pedestrian-centric}} &
\multicolumn{6}{c}{\textbf{Vehicular-centric}} \\ 
\cline{3-14}
 &  & \textbf{BU} & \textbf{Atr} & \textbf{ER} & \textbf{RR} & \textbf{CI} & \textbf{Overall} & \textbf{BU} & \textbf{Atr} & \textbf{ER} & \textbf{RR} & \textbf{CI} & \textbf{Overall} \\ 
\midrule
\multirow{4}{*}{\textbf{Proprietary}} 
 & \textbf{Gemini 2.5 Pro} &
  \cellcolor[HTML]{FD6864}93.43 & \cellcolor[HTML]{F8A102}83.03 & \cellcolor[HTML]{FFCE93}73.32 & \cellcolor[HTML]{FFCE93}76.39 & \cellcolor[HTML]{FD6864}92.81 & \cellcolor[HTML]{F8A102}87.90 &
  \cellcolor[HTML]{FFFE65}66.67 & \cellcolor[HTML]{FFFC9E}33.84 & \cellcolor[HTML]{FCFF2F}69.96 & \cellcolor[HTML]{FFCE93}77.10 & \cellcolor[HTML]{F8A102}84.72 & \cellcolor[HTML]{FCFF2F}68.59 \\
 & \textbf{Gemini 2.5 Flash} &
  \cellcolor[HTML]{FCFF2F}68.93 & \cellcolor[HTML]{F8A102}87.27 & \cellcolor[HTML]{FFCE93}72.12 & \cellcolor[HTML]{FFCE93}78.12 & \cellcolor[HTML]{FD6864}94.77 & \cellcolor[HTML]{F8A102}80.21 &
  \cellcolor[HTML]{FFCE93}71.11 & \cellcolor[HTML]{FFFC9E}38.62 & \cellcolor[HTML]{FFCE93}72.46 & \cellcolor[HTML]{FFCE93}73.68 & \cellcolor[HTML]{FFCE93}78.60 & \cellcolor[HTML]{FCFF2F}67.93 \\
 & \textbf{GPT-5} &
  \cellcolor[HTML]{FCFF2F}68.93 & \cellcolor[HTML]{F8A102}80.61 & \cellcolor[HTML]{FFCE93}72.12 & \cellcolor[HTML]{FFCE93}76.39 & \cellcolor[HTML]{FD6864}92.16 & \cellcolor[HTML]{FFCE93}78.01 &
  \cellcolor[HTML]{FFCE93}71.11 & \cellcolor[HTML]{FFFC9E}39.49 & \cellcolor[HTML]{FFFE65}55.13 & \cellcolor[HTML]{FFFE65}59.75 & \cellcolor[HTML]{F8A102}82.10 & \cellcolor[HTML]{FCFF2F}60.21 \\
 & \textbf{GPT-4o} &
  \cellcolor[HTML]{FCFF2F}59.42 & \cellcolor[HTML]{96FFFB}25.45 & \cellcolor[HTML]{FCFF2F}56.49 & \cellcolor[HTML]{96FFFB}28.91 & \cellcolor[HTML]{F8A102}83.01 & \cellcolor[HTML]{FCFF2F}55.29 &
  \cellcolor[HTML]{FFFC9E}42.22 & \cellcolor[HTML]{ECF4FF}7.32 & \cellcolor[HTML]{FFFC9E}47.25 & \cellcolor[HTML]{FFFC9E}44.58 & \cellcolor[HTML]{FFCE93}71.53 & \cellcolor[HTML]{FFFC9E}44.56 \\
\hline
\multirow{6}{*}{\textbf{Open Source}} 
 & \textbf{Qwen 2.5 32B} &
  \cellcolor[HTML]{F8A102}87.90 & \cellcolor[HTML]{F8A102}81.82 & \cellcolor[HTML]{FCFF2F}59.38 & \cellcolor[HTML]{FFFC9E}41.38 & \cellcolor[HTML]{F8A102}86.93 & \cellcolor[HTML]{F8A102}80.07 &
  \cellcolor[HTML]{FCFF2F}53.33 & \cellcolor[HTML]{FCFF2F}63.48 & \cellcolor[HTML]{FFFC9E}47.95 & \cellcolor[HTML]{FCFF2F}54.49 & \cellcolor[HTML]{FCFF2F}65.07 & \cellcolor[HTML]{FCFF2F}56.74 \\
 & \textbf{Qwen 2.5 7B} &
  \cellcolor[HTML]{FFCE93}78.10 & \cellcolor[HTML]{FFCE93}79.39 & \cellcolor[HTML]{FCFF2F}65.63 & \cellcolor[HTML]{FFFC9E}46.55 & \cellcolor[HTML]{FD6864}90.85 & \cellcolor[HTML]{FFCE93}77.85 &
  \cellcolor[HTML]{FCFF2F}53.33 & \cellcolor[HTML]{FCFF2F}65.65 & \cellcolor[HTML]{FCFF2F}53.63 & \cellcolor[HTML]{FFFC9E}50.46 & \cellcolor[HTML]{FFFFC7}48.91 & \cellcolor[HTML]{FCFF2F}53.87 \\
 & \textbf{VideoLLaMA3} &
  \cellcolor[HTML]{F8A102}86.74 & \cellcolor[HTML]{96FFFB}16.97 & \cellcolor[HTML]{FCFF2F}64.06 & \cellcolor[HTML]{FFFC9E}37.93 & \cellcolor[HTML]{F8A102}86.27 & \cellcolor[HTML]{FCFF2F}65.86 &
  \cellcolor[HTML]{FFFC9E}46.67 & \cellcolor[HTML]{FCFF2F}59.13 & \cellcolor[HTML]{FFFC9E}50.16 & \cellcolor[HTML]{FFFFC7}47.37 & \cellcolor[HTML]{FFCE93}71.62 & \cellcolor[HTML]{FCFF2F}55.85 \\
 & \textbf{NVILA 8B} &
  \cellcolor[HTML]{F8A102}83.57 & \cellcolor[HTML]{96FFFB}16.97 & \cellcolor[HTML]{FCFF2F}60.94 & \cellcolor[HTML]{96FFFB}29.31 & \cellcolor[HTML]{F8A102}86.19 & \cellcolor[HTML]{FCFF2F}63.68 &
  \cellcolor[HTML]{FCFF2F}57.78 & \cellcolor[HTML]{FCFF2F}56.81 & \cellcolor[HTML]{FCFF2F}64.01 & \cellcolor[HTML]{FCFF2F}52.06 & \cellcolor[HTML]{FFFFC7}46.72 & \cellcolor[HTML]{FCFF2F}55.18 \\
 & \textbf{Llava-NeXT-Video 7B} &
  \cellcolor[HTML]{FCFF2F}69.45 & \cellcolor[HTML]{ECF4FF}7.88 & \cellcolor[HTML]{FFFFC7}37.50 & \cellcolor[HTML]{96FFFB}20.69 & \cellcolor[HTML]{96FFFB}24.84 & \cellcolor[HTML]{FFFC9E}38.62 &
  \cellcolor[HTML]{FFFC9E}37.78 & \cellcolor[HTML]{ECF4FF}8.70 & \cellcolor[HTML]{FFFC9E}43.22 & \cellcolor[HTML]{96FFFB}21.67 & \cellcolor[HTML]{ECF4FF}9.61 & \cellcolor[HTML]{96FFFB}22.93 \\
 & \textbf{InternVL 38B} &
  \cellcolor[HTML]{F8A102}87.83 & \cellcolor[HTML]{96FFFB}28.66 & \cellcolor[HTML]{FFFC9E}62.14 & \cellcolor[HTML]{FFFC9E}39.66 & \cellcolor[HTML]{F8A102}89.47 & \cellcolor[HTML]{FCFF2F}69.47 &
  \cellcolor[HTML]{FFFC9E}37.78 & \cellcolor[HTML]{FCFF2F}50.58 & \cellcolor[HTML]{FFFC9E}47.62 & \cellcolor[HTML]{FFFC9E}50.47 & \cellcolor[HTML]{FFCE93}77.09 & \cellcolor[HTML]{FCFF2F}55.42 \\
\bottomrule
\end{tabular}%
}
\caption{Comparison of model performance across \textit{pedestrian-centric} and \textit{vehicular-centric} scenarios. Color intensity indicates relative accuracy within each reasoning subcategory.}
\label{Quanti_ped_Results}
\end{table*}

\subsection{Quantitative Results}
\label{sec:quant_results}

Table~\ref {Quanti_ped_Results} and Figure~\ref{fig:vqa_barchart_graph} summarize results across models, reasoning categories, and day/night splits. Figure~\ref {fig:vqa_radar_chart} shows performance across the top $4$ performing models. Some key observations are:

\paragraph{\textit{Proprietary vs. Open-source.}}  
Proprietary models achieve the strongest zero-shot performance. Notably, \textit{Gemini 2.5 Pro} attains the top scores across multiple time windows (morning: $75.78$\%, evening: $71.13$\%) and leads in pedestrian-centric ($87.90$\%) and vehicular-centric ($68.59$\%) settings. \textit{Gemini 2.5 Flash} performs best in the afternoon ($74.98$\%).

\paragraph{\textit{Fine-tuned Models Competitiveness.}}  
Parameter-efficient fine-tuning yields substantial gains. The fine-tuned \textit{Qwen 2.5 VL-7B} and \textit{Qwen 2.5 VL-32B} show meaningful improvements in evening conditions (e.g., \textit{Qwen-7B}: $61.38$\%), closing the gap to proprietary models on several categories. Some architectures (e.g., \textit{Llava-NeXT-Video 7B}) still struggle at fine-grained perception (very low attribution in some lighting conditions), indicating architecture- or training-data-induced limits.

\paragraph{\textit{Failure of Fine-grained Attribute Localization.}} 
Across models, we observe systematic attribution errors on fine-grained visual items (e.g., road markings). Example: when asked, \textit{“What large marking is painted in the central lane near the camera?”} ground truth: \textit{“A white ‘X’ road marking”}, model predictions ranged widely (\textit{``diamond'', ``bicycle symbol''}, \textit{``white arrow''}), evidencing a decoupling between scene-level reasoning and low-level visual recognition.

\subsection{Qualitative Insights and Error analysis}
\label{sec:qualitative_analysis}

Qualitative study of model outputs reveals consistent patterns that quantitative metrics alone do not capture(see Table ~\ref{tab:video_qa_day}, Table ~\ref{tab:video_qa_morning} and Table ~\ref{tab:video_qa_evening}). Two recurring phenomena stand out: \textbf{\textit{(1)}} an \textit{inference–attribution scores}, where models succeed at high-level inference but fail at low-level visual grounding, and \textbf{\textit{(2)}} \textit{systematic temporal and premise reasoning failures}. Below, we summarize these failure modes, illustrate them with examples from our test splits, and discuss their implications.

\paragraph{\textit{Inference–Attribution Scores}}
Models frequently demonstrate strong inferential ability, correctly proposing plausible outcomes or causal links while simultaneously failing to ground those inferences in the visual evidence. This produces seemingly good \textit{counterfactual} or \textit{event-reasoning} scores even as the underlying visual attribution is incorrect.

\noindent\textbf{\textit{Example (Counterfactual / Hallucination):}}
\begin{itemize}
  \item \textbf{\textit{Question:}} \textit{``If the van had not yielded, what might have occurred?''}
  \item \textbf{\textit{Ground Truth:}} \textit{``It might have collided with the oncoming car.''}
  \item \textbf{\textit{Prediction:}} \textit{``The impact could have resulted in severe injuries for both the pedestrian and the van's occupants.''}
\end{itemize}

\noindent\textbf{\textit{Analysis.}} The model correctly predicts a collision (matching the ground truth's high-level outcome) but hallucinates an absent pedestrian and invents injury details. In many cases, the model relies on a generic \textit{``traffic accident''} script learned from language priors rather than verifying whether a pedestrian is actually present in the clip. This behavior inflates performance on inference-style tasks while undermining factual fidelity.

\begin{table*}[t]
\resizebox{\linewidth}{!}{%
\begin{tabular}{
>{\columncolor[HTML]{FFFFFF}}c |
>{\columncolor[HTML]{FFFFFF}}c 
>{\columncolor[HTML]{FFFFFF}}c 
>{\columncolor[HTML]{FFFFFF}}c 
>{\columncolor[HTML]{FFFFFF}}c 
>{\columncolor[HTML]{FFFFFF}}c 
>{\columncolor[HTML]{FFFFFF}}c 
>{\columncolor[HTML]{FFFFFF}}c 
>{\columncolor[HTML]{FFFFFF}}c 
>{\columncolor[HTML]{FFFFFF}}c 
>{\columncolor[HTML]{FFFFFF}}c 
>{\columncolor[HTML]{FFFFFF}}c }
\toprule
\textbf{Model} &
  \multicolumn{1}{c|}{\cellcolor[HTML]{FFFFFF}\textbf{Gemini 2.5 Pro}} &
  \multicolumn{1}{c|}{\cellcolor[HTML]{FFFFFF}\textbf{\begin{tabular}[c]{@{}c@{}}Gemini 2.5 \\ Flash\end{tabular}}} &
  \multicolumn{1}{c|}{\cellcolor[HTML]{FFFFFF}\textbf{GPT-4o}} &
  \multicolumn{1}{c|}{\cellcolor[HTML]{FFFFFF}\textbf{GPT-5}} &
  \multicolumn{1}{c|}{\cellcolor[HTML]{FFFFFF}\textbf{\begin{tabular}[c]{@{}c@{}}Qwen 2.5 VL - \\ 7B (Fine Tuned)\end{tabular}}} &
  \multicolumn{1}{c|}{\cellcolor[HTML]{FFFFFF}\textbf{Qwen 2.5 VL - 7B}} &
  \multicolumn{1}{c|}{\cellcolor[HTML]{FFFFFF}\textbf{InternVL3-38B}} &
  \multicolumn{1}{c|}{\cellcolor[HTML]{FFFFFF}\textbf{Qwen 2.5 VL 32B}} &
  \multicolumn{1}{c|}{\cellcolor[HTML]{FFFFFF}\textbf{VideoLlama3}} &
  \multicolumn{1}{c|}{\cellcolor[HTML]{FFFFFF}\textbf{NVILA 8B}} &
  \textbf{Llava-NeXT-Video } \\ \midrule
\cellcolor[HTML]{FFD966}\textbf{Question} & \multicolumn{11}{c}{\cellcolor[HTML]{FFD966}Is there any overtaking observed?} \\ \hline
\textbf{Ground Truth} & \multicolumn{11}{c}{\cellcolor[HTML]{FFFFFF}No} \\ \hline
\cellcolor[HTML]{D9EAD3}\textbf{Best Answer} &
  \multicolumn{1}{c|}{\cellcolor[HTML]{D9EAD3}\begin{tabular}[c]{@{}c@{}}No overtaking is \\ observed as there \\ are no moving \\ vehicles\end{tabular}} &
  \multicolumn{1}{c|}{\cellcolor[HTML]{D9EAD3}No} &
  \multicolumn{1}{c|}{\cellcolor[HTML]{D9EAD3}No} &
  \multicolumn{1}{c|}{\cellcolor[HTML]{D9EAD3}No} &
  \multicolumn{1}{c|}{\cellcolor[HTML]{D9EAD3}\begin{tabular}[c]{@{}c@{}}No overtaking is \\ observed in the \\ image.\end{tabular}} &
  \multicolumn{1}{c|}{\cellcolor[HTML]{D9EAD3}\begin{tabular}[c]{@{}c@{}}No, there is no overtaking \\ observed {[}...{]}. The road \\ appears to be empty with \\ no vehicles in motion {[}...{]}\end{tabular}} &
  \multicolumn{1}{c|}{\cellcolor[HTML]{D9EAD3}\begin{tabular}[c]{@{}c@{}}No, there is no \\ overtaking \\ observed.\end{tabular}} &
  \multicolumn{1}{c|}{\cellcolor[HTML]{D9EAD3}\begin{tabular}[c]{@{}c@{}}No, there is no overtaking \\ observed {[}...{]}. The scene shows \\ an empty road with no vehicles \\ or pedestrians present. {[}...{]}\end{tabular}} &
  \multicolumn{1}{c|}{\cellcolor[HTML]{D9EAD3}\begin{tabular}[c]{@{}c@{}}The video does \\ not show any \\ overtaking.\end{tabular}} &
  \multicolumn{1}{c|}{\cellcolor[HTML]{D9EAD3}\begin{tabular}[c]{@{}c@{}}No overtaking \\ is observed\\ in the image.\end{tabular}} &
  \cellcolor[HTML]{D9EAD3}\begin{tabular}[c]{@{}c@{}}Based on the image \\ provided, it appears \\ that there is no over-\\ taking place. {[}...{]}\end{tabular} \\ \midrule \midrule 
\cellcolor[HTML]{FFD966}\textbf{Question} & \multicolumn{11}{c}{\cellcolor[HTML]{FFD966}How many cars exited the foreground?} \\ \hline
\textbf{Ground Truth} & \multicolumn{11}{c}{\cellcolor[HTML]{FFFFFF}Four} \\ \hline
\cellcolor[HTML]{F4CCCC}\textbf{Worst Answer} &
  \multicolumn{1}{c|}{\cellcolor[HTML]{F4CCCC}\begin{tabular}[c]{@{}c@{}}Zero cars exited \\ the foreground as \\ the car moved away \\ from the camera.\end{tabular}} &
  \multicolumn{1}{c|}{\cellcolor[HTML]{F4CCCC}\begin{tabular}[c]{@{}c@{}}One car\\ exited the\\ fore-\\ ground\end{tabular}} &
  \multicolumn{1}{c|}{\cellcolor[HTML]{F4CCCC}Two} &
  \multicolumn{1}{c|}{\cellcolor[HTML]{F4CCCC}0} &
  \multicolumn{1}{c|}{\cellcolor[HTML]{F4CCCC}\begin{tabular}[c]{@{}c@{}}One car exited \\ the fore-ground \\ in the image.\end{tabular}} &
  \multicolumn{1}{c|}{\cellcolor[HTML]{F4CCCC}\begin{tabular}[c]{@{}c@{}}{[}...{]} there is one car that \\ appears to have just exited \\ the foreground on the \\ right side {[}...{]}\end{tabular}} &
  \multicolumn{1}{c|}{\cellcolor[HTML]{F4CCCC}\begin{tabular}[c]{@{}c@{}}One car exited \\ the foreground\end{tabular}} &
  \multicolumn{1}{c|}{\cellcolor[HTML]{F4CCCC}\begin{tabular}[c]{@{}c@{}}{[}...{]} there is one car that exits \\ the foreground. It can be seen\\  moving from the right side {[}...{]} \\ eventually exiting the frame.\end{tabular}} &
  \multicolumn{1}{c|}{\cellcolor[HTML]{F4CCCC}\begin{tabular}[c]{@{}c@{}}{[}...{]} it appears that \\ one car has exited the \\ foreground. The road \\ is mostly empty {[}...{]}\end{tabular}} &
  \multicolumn{1}{c|}{\cellcolor[HTML]{F4CCCC}\begin{tabular}[c]{@{}c@{}}There is one \\ car that exited \\ the foreground.\end{tabular}} &
  \cellcolor[HTML]{F4CCCC}\begin{tabular}[c]{@{}c@{}}There are three cars \\ in the foreground of \\ the image.\end{tabular} \\ \bottomrule
\end{tabular}%
}
\caption{Representative examples of best and worst model-generated answers under the \textit{daytime} scenario.}
\label{tab:video_qa_day}
\end{table*}

\begin{table*}[ht]
\centering
\resizebox{\textwidth}{!}{%
\begin{tabular}{
>{\columncolor[HTML]{FFFFFF}}c |
>{\columncolor[HTML]{FFFFFF}}c 
>{\columncolor[HTML]{FFFFFF}}c 
>{\columncolor[HTML]{FFFFFF}}c 
>{\columncolor[HTML]{FFFFFF}}c 
>{\columncolor[HTML]{FFFFFF}}c 
>{\columncolor[HTML]{FFFFFF}}c 
>{\columncolor[HTML]{FFFFFF}}c 
>{\columncolor[HTML]{FFFFFF}}c 
>{\columncolor[HTML]{FFFFFF}}c 
>{\columncolor[HTML]{FFFFFF}}c 
>{\columncolor[HTML]{FFFFFF}}c }
\hline
\textbf{Model} &
  \multicolumn{1}{c|}{\cellcolor[HTML]{FFFFFF}\textbf{Gemini 2.5 Pro}} &
  \multicolumn{1}{c|}{\cellcolor[HTML]{FFFFFF}\textbf{\begin{tabular}[c]{@{}c@{}}Gemini 2.5 \\ Flash\end{tabular}}} &
  \multicolumn{1}{c|}{\cellcolor[HTML]{FFFFFF}\textbf{GPT-4o}} &
  \multicolumn{1}{c|}{\cellcolor[HTML]{FFFFFF}\textbf{GPT-5}} &
  \multicolumn{1}{c|}{\cellcolor[HTML]{FFFFFF}\textbf{\begin{tabular}[c]{@{}c@{}}Qwen 2.5 VL - \\ 7B (Fine Tuned)\end{tabular}}} &
  \multicolumn{1}{c|}{\cellcolor[HTML]{FFFFFF}\textbf{Qwen 2.5 VL - 7B}} &
  \multicolumn{1}{c|}{\cellcolor[HTML]{FFFFFF}\textbf{InternVL3-38B}} &
  \multicolumn{1}{c|}{\cellcolor[HTML]{FFFFFF}\textbf{Qwen 2.5 VL 32B}} &
  \multicolumn{1}{c|}{\cellcolor[HTML]{FFFFFF}\textbf{VideoLlama3}} &
  \multicolumn{1}{c|}{\cellcolor[HTML]{FFFFFF}\textbf{NVILA 8B}} &
  \textbf{Llava-NeXT-Video} \\ \midrule
\cellcolor[HTML]{FFD966}\textbf{Question} &
\multicolumn{11}{c}{\cellcolor[HTML]{FFD966}Is the road surface wet from rain?} \\ \hline
\textbf{Ground Truth} &
  \multicolumn{11}{c}{\cellcolor[HTML]{FFFFFF}No the road looks dry} \\ \hline
\cellcolor[HTML]{D9EAD3}\textbf{Best Answer} &
  \multicolumn{1}{c|}{\cellcolor[HTML]{D9EAD3}\begin{tabular}[c]{@{}c@{}}No, the road \\ surface \\ is not wet \\ from rain; \\ it appears \\ completely dry.\end{tabular}} &
  \multicolumn{1}{c|}{\cellcolor[HTML]{D9EAD3}\begin{tabular}[c]{@{}c@{}}No, the road \\ surface\\  appears dry, \\ not wet \\ from rain.\end{tabular}} &
  \multicolumn{1}{c|}{\cellcolor[HTML]{D9EAD3}No} &
  \multicolumn{1}{c|}{\cellcolor[HTML]{D9EAD3}No} &
  \multicolumn{1}{c|}{\cellcolor[HTML]{D9EAD3}\begin{tabular}[c]{@{}c@{}}No, the road \\ surfaceis not wet \\ from rain.\\  The road appears \\ dry and clear of \\ any water.\end{tabular}} &
  \multicolumn{1}{c|}{\cellcolor[HTML]{D9EAD3}\begin{tabular}[c]{@{}c@{}}No, there is no wetness\\  from rain\\  observed {[}...{]}. The road \\ appears \\ to be dry with no visible \\ signs of water {[}...{]}\end{tabular}} &
  \multicolumn{1}{c|}{\cellcolor[HTML]{D9EAD3}\begin{tabular}[c]{@{}c@{}}No, the road \\ surface\\  appears dry.\end{tabular}} &
  \multicolumn{1}{c|}{\cellcolor[HTML]{D9EAD3}\begin{tabular}[c]{@{}c@{}}No, there is no \\ wetness from rain\\  observed {[}...{]}. \\ The road surface \\ appears\\  to be dry with \\ no visible signs of \\ water or \\ reflections {[}...{]}\end{tabular}} &
  \multicolumn{1}{c|}{\cellcolor[HTML]{D9EAD3}\begin{tabular}[c]{@{}c@{}}The road surface \\ appears dry,\\  with no visible\\  signs\\  of rain or \\ wetness.\end{tabular}} &
  \multicolumn{1}{c|}{\cellcolor[HTML]{D9EAD3}\begin{tabular}[c]{@{}c@{}}No, the road surface\\  is not wet \\ from rain.\end{tabular}} &
  \cellcolor[HTML]{D9EAD3}\begin{tabular}[c]{@{}c@{}}No, there is no \\ wetness from rain\\  observed {[}...{]}. \\ The road surface\\  appears to be dry \\ and clear with \\ it not raining \\ at the time {[}...{]}\end{tabular} \\ \midrule \midrule
\cellcolor[HTML]{FFD966}\textbf{Question} &
\multicolumn{11}{c}{\cellcolor[HTML]{FFD966}What large marking is painted in the central lane near the camera?} \\ \hline
\textbf{Ground Truth} &
  \multicolumn{11}{c}{\cellcolor[HTML]{FFFFFF}A white ‘X’ road marking} \\ \hline
\cellcolor[HTML]{F4CCCC}\textbf{Worst Answer} &
  \multicolumn{1}{c|}{\cellcolor[HTML]{F4CCCC}\begin{tabular}[c]{@{}c@{}}A white bicycle\\  symbol.\end{tabular}} &
  \multicolumn{1}{c|}{\cellcolor[HTML]{F4CCCC}\begin{tabular}[c]{@{}c@{}}A large diamond \\ shape \\ (?) is painted in \\ the central lane.\end{tabular}} &
  \multicolumn{1}{c|}{\cellcolor[HTML]{F4CCCC}\begin{tabular}[c]{@{}c@{}}Straight \\ white arrow\end{tabular}} &
  \multicolumn{1}{c|}{\cellcolor[HTML]{F4CCCC}\begin{tabular}[c]{@{}c@{}}A large \\ white \\ marking\\  is painted \\ in the \\ central lane.\end{tabular}} &
  \multicolumn{1}{c|}{\cellcolor[HTML]{F4CCCC}\begin{tabular}[c]{@{}c@{}}The large marking \\ painted\\  in the central lane \\ near the camera \\ is a \\ white arrow \\ pointing \\ to the right\end{tabular}} &
  \multicolumn{1}{c|}{\cellcolor[HTML]{F4CCCC}\begin{tabular}[c]{@{}c@{}}The large marking \\ appears to be a \\ white arrow pointing \\ to the right\\  with \\ it indicating a \\ turn lane {[}...{]}\end{tabular}} &
  \multicolumn{1}{c|}{\cellcolor[HTML]{F4CCCC}\begin{tabular}[c]{@{}c@{}}The large marking \\ painted in\\  the central \\ lane near \\ the camera \\ is a 'no parking' \\ sign\end{tabular}} &
  \multicolumn{1}{c|}{\cellcolor[HTML]{F4CCCC}\begin{tabular}[c]{@{}c@{}}The large marking\\  appears to be a \\ "Bus Only" or "HOV" \\ lane marking\\  with \\ this being indicated by \\ the red color {[}...{]}\end{tabular}} &
  \multicolumn{1}{c|}{\cellcolor[HTML]{F4CCCC}\begin{tabular}[c]{@{}c@{}}The large marking \\ appears to be a \\ white arrow \\ pointing to the\\  right with \\ it indicating a \\ lane change or \\ turn\\  direction {[}...{]}\end{tabular}} &
  \multicolumn{1}{c|}{\cellcolor[HTML]{F4CCCC}\begin{tabular}[c]{@{}c@{}}The large marking\\  painted\\  in the central lane \\ near the camera \\ is a white arrow.\end{tabular}} &
  \cellcolor[HTML]{F4CCCC}\begin{tabular}[c]{@{}c@{}}The large marking\\  painted\\  in the central lane\\  near the camera\\  is a "No Parking"\\  sign.\end{tabular} \\ \bottomrule
\end{tabular}%
}
\caption{Representative examples of best and worst model-generated answers under the \textit{afternoon} scenario.}
\label{tab:video_qa_morning}
\end{table*}

\begin{table*}[ht]
\centering
\resizebox{\textwidth}{!}{%
\begin{tabular}{
>{\columncolor[HTML]{FFFFFF}}c |
>{\columncolor[HTML]{FFFFFF}}c 
>{\columncolor[HTML]{FFFFFF}}c 
>{\columncolor[HTML]{FFFFFF}}c 
>{\columncolor[HTML]{FFFFFF}}c 
>{\columncolor[HTML]{FFFFFF}}c 
>{\columncolor[HTML]{FFFFFF}}c 
>{\columncolor[HTML]{FFFFFF}}c 
>{\columncolor[HTML]{FFFFFF}}c 
>{\columncolor[HTML]{FFFFFF}}c 
>{\columncolor[HTML]{FFFFFF}}c 
>{\columncolor[HTML]{FFFFFF}}c }
\hline
\textbf{Model} &
  \multicolumn{1}{c|}{\cellcolor[HTML]{FFFFFF}\textbf{Gemini 2.5 Pro}} &
  \multicolumn{1}{c|}{\cellcolor[HTML]{FFFFFF}\textbf{\begin{tabular}[c]{@{}c@{}}Gemini 2.5 \\ Flash\end{tabular}}} &
  \multicolumn{1}{c|}{\cellcolor[HTML]{FFFFFF}\textbf{GPT-4o}} &
  \multicolumn{1}{c|}{\cellcolor[HTML]{FFFFFF}\textbf{GPT-5}} &
  \multicolumn{1}{c|}{\cellcolor[HTML]{FFFFFF}\textbf{\begin{tabular}[c]{@{}c@{}}Qwen 2.5 VL - \\ 7B (Fine Tuned)\end{tabular}}} &
  \multicolumn{1}{c|}{\cellcolor[HTML]{FFFFFF}\textbf{Qwen 2.5 VL - 7B}} &
  \multicolumn{1}{c|}{\cellcolor[HTML]{FFFFFF}\textbf{InternVL3-38B}} &
  \multicolumn{1}{c|}{\cellcolor[HTML]{FFFFFF}\textbf{Qwen 2.5 VL 32B}} &
  \multicolumn{1}{c|}{\cellcolor[HTML]{FFFFFF}\textbf{VideoLlama3}} &
  \multicolumn{1}{c|}{\cellcolor[HTML]{FFFFFF}\textbf{NVILA 8B}} &
  \textbf{Llava-NeXT-Video} \\ \midrule 
\cellcolor[HTML]{FFD966}\textbf{Question}&
  \multicolumn{11}{c}{\cellcolor[HTML]{FFD966}Is the road surface wet from rain?} \\ \hline
\textbf{Ground Truth} &
  \multicolumn{11}{c}{\cellcolor[HTML]{FFFFFF}No the road looks dry} \\ \hline
\cellcolor[HTML]{D9EAD3}\textbf{Best Answer} &
  \multicolumn{1}{c|}{\cellcolor[HTML]{D9EAD3}\begin{tabular}[c]{@{}c@{}}No, the road \\ surface \\ is not wet \\ from rain; \\ it appears \\ completely dry.\end{tabular}} &
  \multicolumn{1}{c|}{\cellcolor[HTML]{D9EAD3}\begin{tabular}[c]{@{}c@{}}No, the road \\ surface\\  appears dry, \\ not wet \\ from rain.\end{tabular}} &
  \multicolumn{1}{c|}{\cellcolor[HTML]{D9EAD3}No} &
  \multicolumn{1}{c|}{\cellcolor[HTML]{D9EAD3}No} &
  \multicolumn{1}{c|}{\cellcolor[HTML]{D9EAD3}\begin{tabular}[c]{@{}c@{}}No, the road \\ surfaceis not wet \\ from rain.\\  The road appears \\ dry and clear of \\ any water.\end{tabular}} &
  \multicolumn{1}{c|}{\cellcolor[HTML]{D9EAD3}\begin{tabular}[c]{@{}c@{}}No, there is no wetness\\  from rain\\  observed {[}...{]}. The road \\ appears \\ to be dry with no visible \\ signs of water {[}...{]}\end{tabular}} &
  \multicolumn{1}{c|}{\cellcolor[HTML]{D9EAD3}\begin{tabular}[c]{@{}c@{}}No, the road \\ surface\\  appears dry.\end{tabular}} &
  \multicolumn{1}{c|}{\cellcolor[HTML]{D9EAD3}\begin{tabular}[c]{@{}c@{}}No, there is no \\ wetness from rain\\  observed {[}...{]}. \\ The road surface \\ appears\\  to be dry with \\ no visible signs of \\ water or \\ reflections {[}...{]}\end{tabular}} &
  \multicolumn{1}{c|}{\cellcolor[HTML]{D9EAD3}\begin{tabular}[c]{@{}c@{}}The road surface \\ appears dry,\\  with no visible\\  signs\\  of rain or \\ wetness.\end{tabular}} &
  \multicolumn{1}{c|}{\cellcolor[HTML]{D9EAD3}\begin{tabular}[c]{@{}c@{}}No, the road surface\\  is not wet \\ from rain.\end{tabular}} &
  \cellcolor[HTML]{D9EAD3}\begin{tabular}[c]{@{}c@{}}No, there is no \\ wetness from rain\\  observed {[}...{]}. \\ The road surface\\  appears to be dry \\ and clear with \\ it not raining \\ at the time {[}...{]}\end{tabular} \\ \midrule \midrule
\cellcolor[HTML]{FFD966}\textbf{Question} &
 \multicolumn{11}{c}{\cellcolor[HTML]{FFD966}What large marking is painted in the central lane near the camera?} \\ \hline
\textbf{Ground Truth} &
  \multicolumn{11}{c}{\cellcolor[HTML]{FFFFFF}A white ‘X’ road marking} \\ \hline
\cellcolor[HTML]{F4CCCC}\textbf{Worst Answer} &
  \multicolumn{1}{c|}{\cellcolor[HTML]{F4CCCC}\begin{tabular}[c]{@{}c@{}}A white bicycle\\  symbol.\end{tabular}} &
  \multicolumn{1}{c|}{\cellcolor[HTML]{F4CCCC}\begin{tabular}[c]{@{}c@{}}A large diamond \\ shape \\ (?) is painted in \\ the central lane.\end{tabular}} &
  \multicolumn{1}{c|}{\cellcolor[HTML]{F4CCCC}\begin{tabular}[c]{@{}c@{}}Straight \\ white arrow\end{tabular}} &
  \multicolumn{1}{c|}{\cellcolor[HTML]{F4CCCC}\begin{tabular}[c]{@{}c@{}}A large \\ white \\ marking\\  is painted \\ in the \\ central lane.\end{tabular}} &
  \multicolumn{1}{c|}{\cellcolor[HTML]{F4CCCC}\begin{tabular}[c]{@{}c@{}}The large marking \\ painted\\  in the central lane \\ near the camera \\ is a \\ white arrow \\ pointing \\ to the right\end{tabular}} &
  \multicolumn{1}{c|}{\cellcolor[HTML]{F4CCCC}\begin{tabular}[c]{@{}c@{}}The large marking \\ appears to be a \\ white arrow pointing \\ to the right\\  with \\ it indicating a \\ turn lane {[}...{]}\end{tabular}} &
  \multicolumn{1}{c|}{\cellcolor[HTML]{F4CCCC}\begin{tabular}[c]{@{}c@{}}The large marking \\ painted in\\  the central \\ lane near \\ the camera \\ is a 'no parking' \\ sign\end{tabular}} &
  \multicolumn{1}{c|}{\cellcolor[HTML]{F4CCCC}\begin{tabular}[c]{@{}c@{}}The large marking\\  appears to be a \\ "Bus Only" or "HOV" \\ lane marking\\  with \\ this being indicated by \\ the red color {[}...{]}\end{tabular}} &
  \multicolumn{1}{c|}{\cellcolor[HTML]{F4CCCC}\begin{tabular}[c]{@{}c@{}}The large marking \\ appears to be a \\ white arrow \\ pointing to the\\  right with \\ it indicating a \\ lane change or \\ turn\\  direction {[}...{]}\end{tabular}} &
  \multicolumn{1}{c|}{\cellcolor[HTML]{F4CCCC}\begin{tabular}[c]{@{}c@{}}The large marking\\  painted\\  in the central lane \\ near the camera \\ is a white arrow.\end{tabular}} &
  \cellcolor[HTML]{F4CCCC}\begin{tabular}[c]{@{}c@{}}The large marking\\  painted\\  in the central lane\\  near the camera\\  is a "No Parking"\\  sign.\end{tabular} \\ \bottomrule
\end{tabular}%
}
\caption{Representative examples of best and worst model-generated answers under the \textit{evening} scenario}
\label{tab:video_qa_evening}
\end{table*}

\paragraph{\textit{Temporal and Premise Reasoning Failures.}}
A second common failure mode concerns temporal localization and premise validation. Models often fail to respect explicit temporal windows or to check the truth of the question’s premise against the video evidence.

\noindent\textbf{\textit{Example (Temporal / Premise Error on Evening Set):}}
\begin{itemize}
  \item \textbf{\textit{Question:}} \textit{``Which happened first: the first car entering the foreground or the second one?''}
  \item \textbf{\textit{Ground Truth:}} \textit{``No car enters the foreground.''}
  \item \textbf{\textit{Prediction (multiple models):}} \textit{``The first car...''}
\end{itemize}

\noindent\textbf{\textit{Analysis.}} Here, models assert an order of events despite the correct answer indicating that no such event occurred within the queried interval. This indicates failures in \textbf{\textit{(a)}} isolating the exact temporal window referenced by the question, and \textbf{\textit{(b)}} verifying whether the question’s premise holds before attempting a comparative judgment.

\paragraph{\textit{Fine-grained Perception Failure.}}
Notably, attribution-style questions that require precise localization or recognition of small, high-frequency visual primitives (e.g., \textit{road markings, signage, small symbols}) are among the weakest for all evaluated models.

\noindent\textbf{\textit{Example (Attribution Failure on Morning Set):}}
\begin{itemize}
  \item \textbf{\textit{Question:}} \textit{``What large marking is painted in the central lane near the camera?''}
  \item \textbf{\textit{Ground Truth:}} \textit{``A white `X' road marking.''}
  \item \textbf{\textit{Prediction:}}\textit{Qwen-7B:} \textit{``A white arrow pointing to the right.''}
       \textit{VideoLLaMA 3:} \textit{``A yellow diamond shape.''}, and 
       \textit{Gemini 2.5 Pro:} \textit{``A white bicycle symbol.''}
\end{itemize}

\noindent\textbf{\textit{Analysis.}} Despite being a visually local and ostensibly simple task, models produce a variety of incorrect but plausible labels. This demonstrates a persistent gap between \textbf{\textit{(i)}} scene-level, multi-agent reasoning and \textbf{\textit{(ii)}} reliable low-level visual recognition. In practice, this gap means that models may reach correct conclusions for the wrong perceptual reasons or fail important downstream decisions because they cannot consistently perceive simple visual cues.

\paragraph{\textit{Implications and Directions.}}
Together, these qualitative failure modes suggest two complementary research priorities for \textit{VideoQA} and multimodal reasoning:
\begin{itemize}
  \item \textbf{\textit{Tighter Visual–Verbal Grounding:}} Improve models’ ability to verify perceptual premises (\textit{``is X present?''}), and to condition any inference on an explicit check of visual evidence rather than on language priors alone.
  \item \textbf{\textit{Temporal Precision and Localized Perception:}} Enhance temporal alignment and small-object recognition through either improved video encoders, targeted supervision (e.g., \textit{contrastive/object-centric objectives}), or hybrid symbolic checks that force premise validation before inference.
\end{itemize}

\begin{figure*}[t!]
\centering

\includegraphics[width=1.0\linewidth]{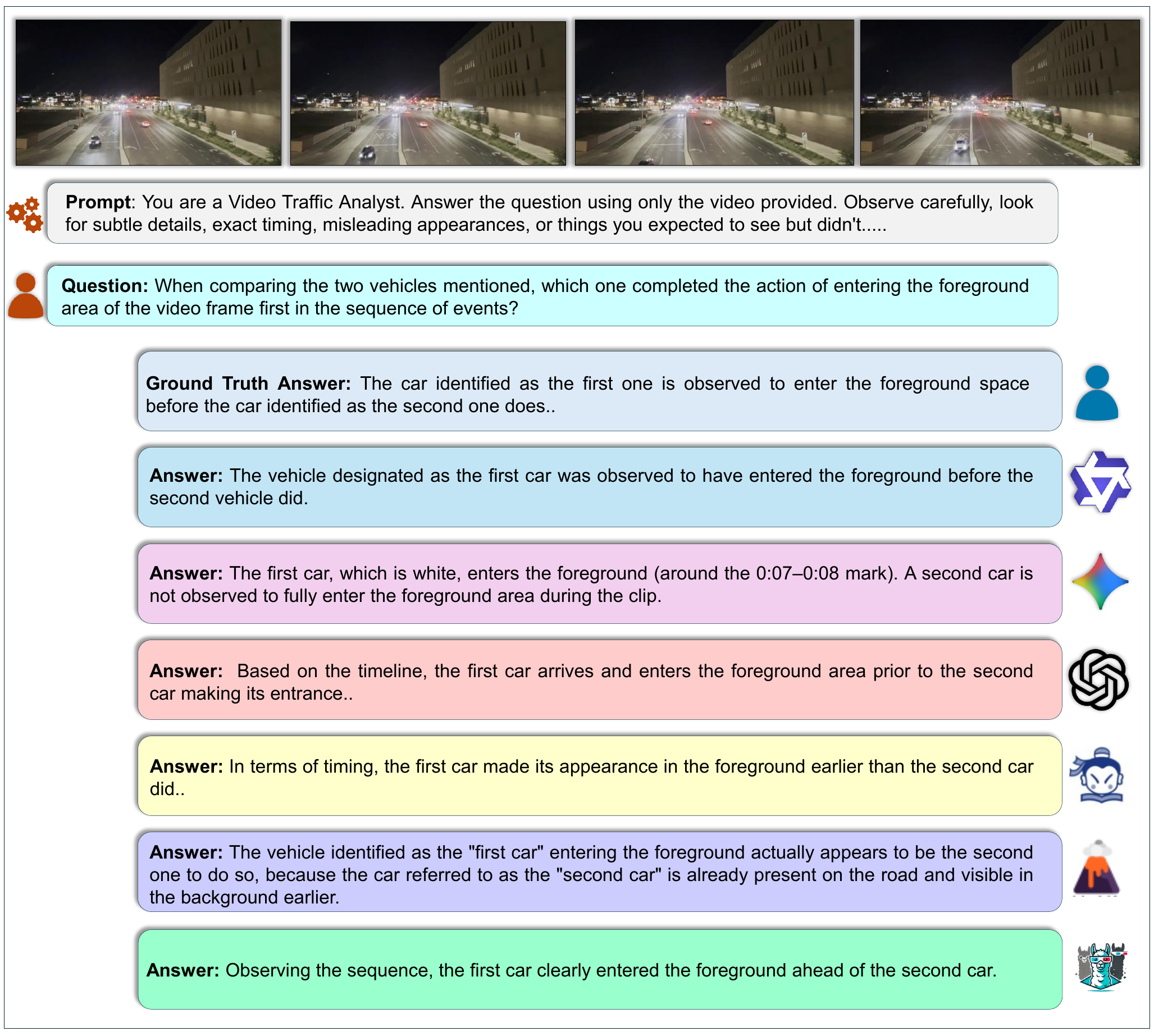}
\caption{Example of a \textit{reverse reasoning} question evaluating fine-grained temporal understanding. Models must track two distinct objects and determine their correct sequential order. \textit{Gemini 2.5 Pro} provides the most accurate response by focusing on the temporal relationships within the sequence.}
\label{fig:evening_Clips} 
\end{figure*}

\begin{figure*}[t!]
\centering

\includegraphics[width=1.0\linewidth]{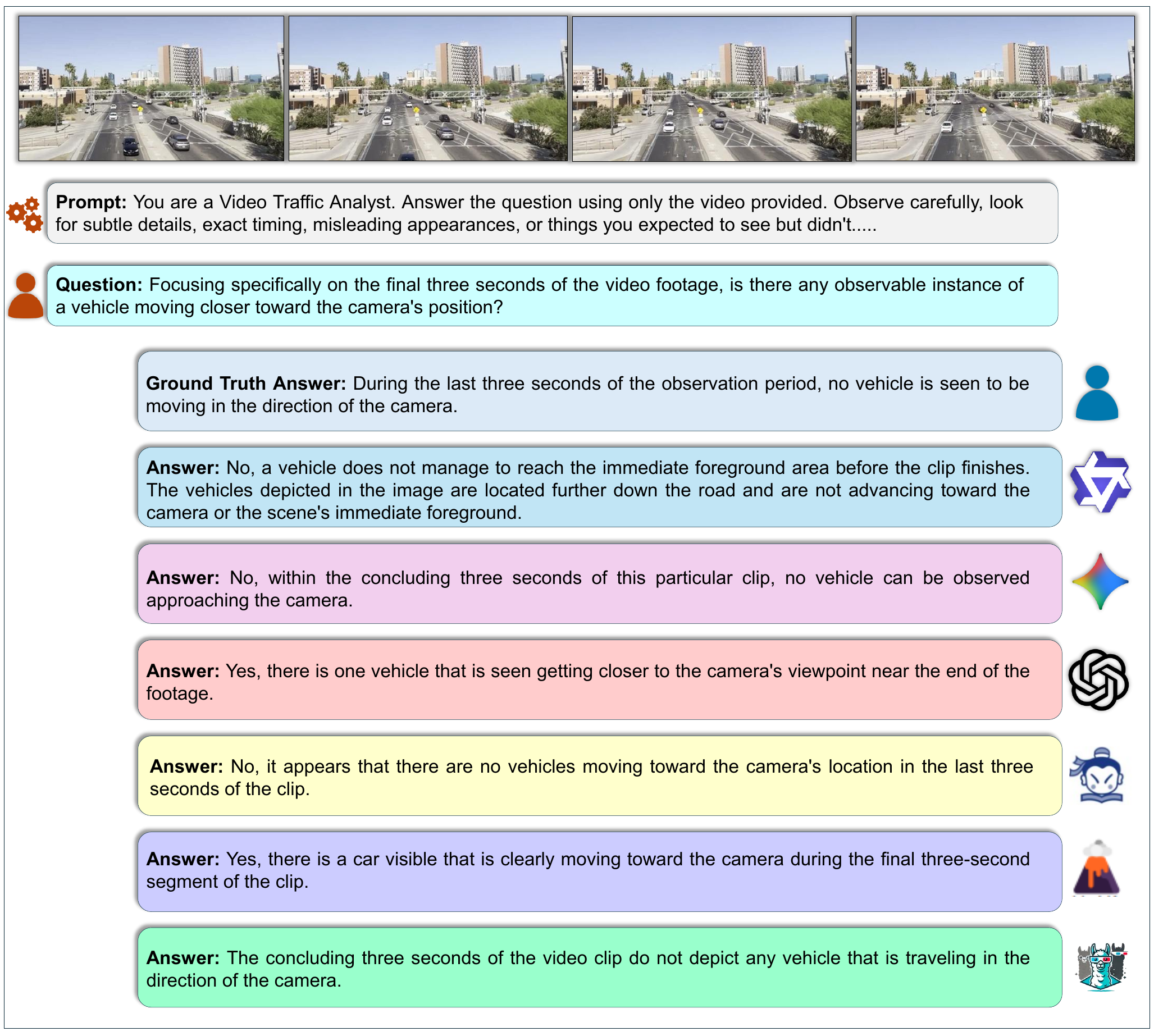}
\caption{. Example of an \textit{event reasoning} challenge illustrating model failures in temporal grounding. Although the ground truth indicates that no vehicle approaches the camera during the final three seconds, several models (e.g., \textit{GPT} and \textit{LLaVA-NeXT-Video}) incorrectly respond ``Yes”, revealing hallucinated motion and poor temporal localization.}
\label{fig:morning_clips} 
\end{figure*}

\begin{figure*}[t!]
\centering

\includegraphics[width=1.0\textwidth]{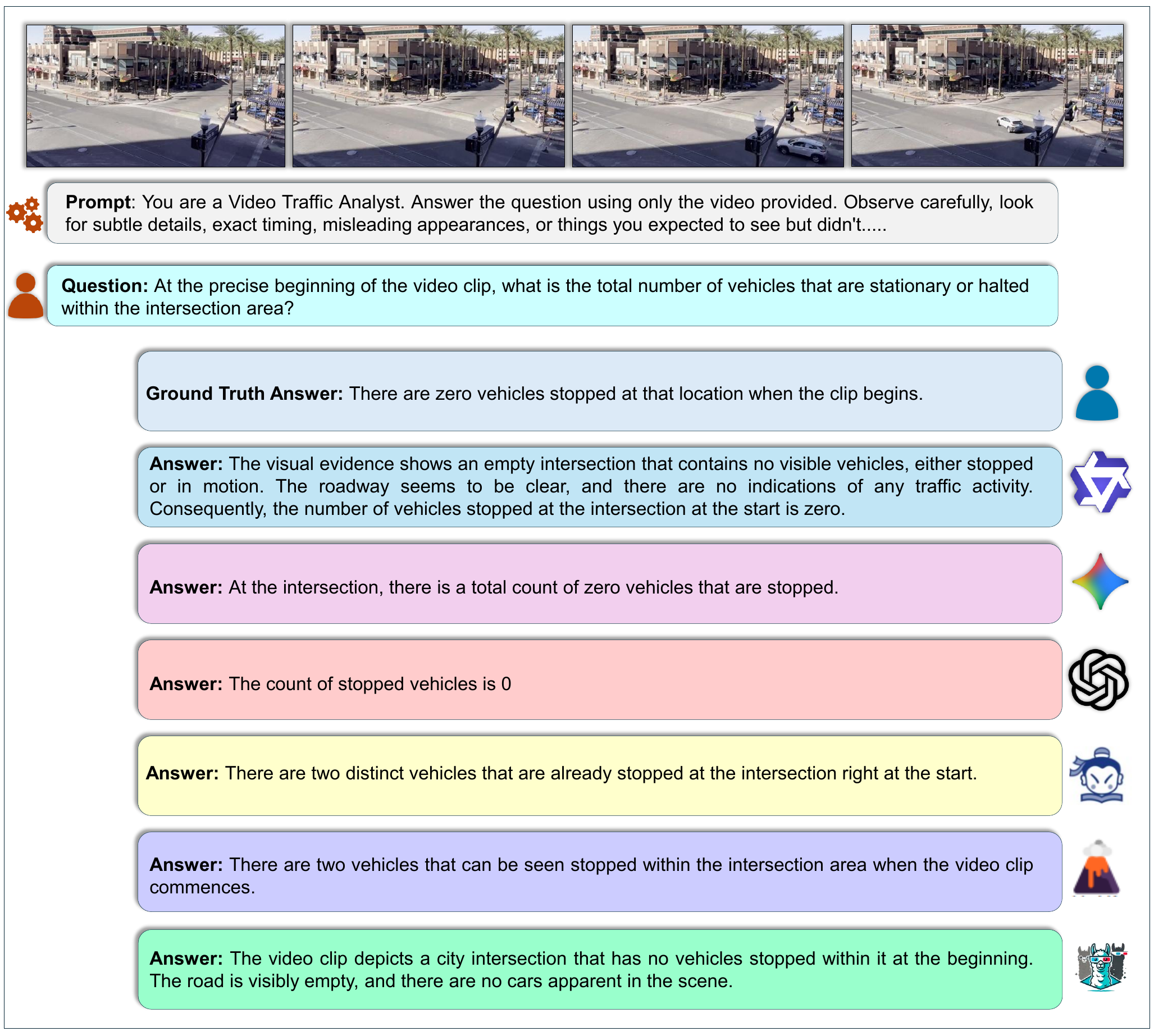}
\caption{Example of an \textit{attribution} question assessing a specific visual attribute, the number of stopped vehicles at the start of the clip. Most models correctly identify zero vehicles, consistent with the ground truth.}
\label{fig:attribution_lol} 
\end{figure*}

\begin{figure*}[t!]
\centering

\includegraphics[width=1.0\textwidth]{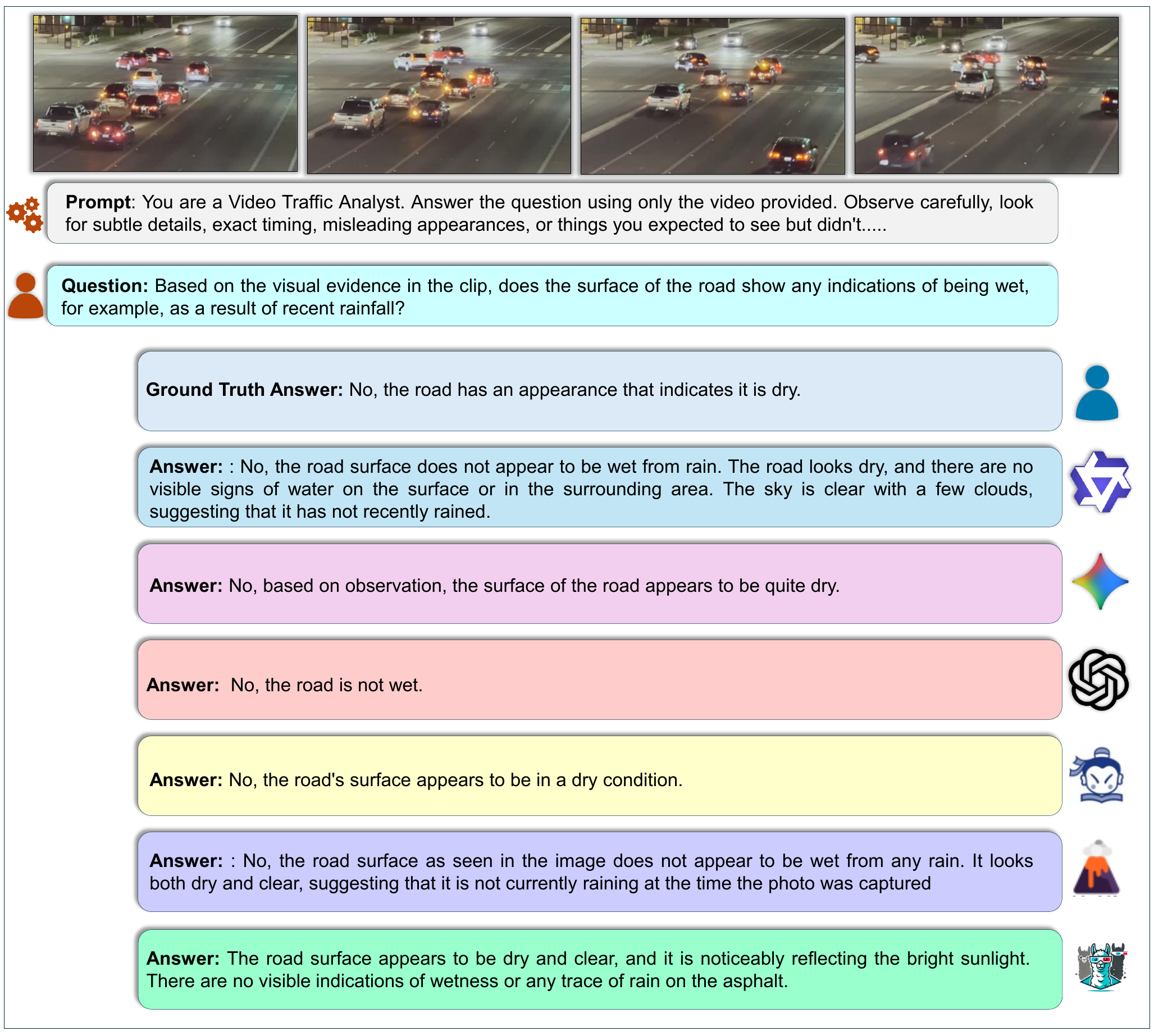}
\caption{Example of a \textit{basic understanding} question evaluating simple environmental perception, specifically, road surface wetness. All models correctly identify the road as dry, demonstrating high accuracy on this foundational reasoning task.}
\label{fig:basics} 
\end{figure*}

\begin{figure*}[t!]
\centering

\includegraphics[width=1.0\textwidth]{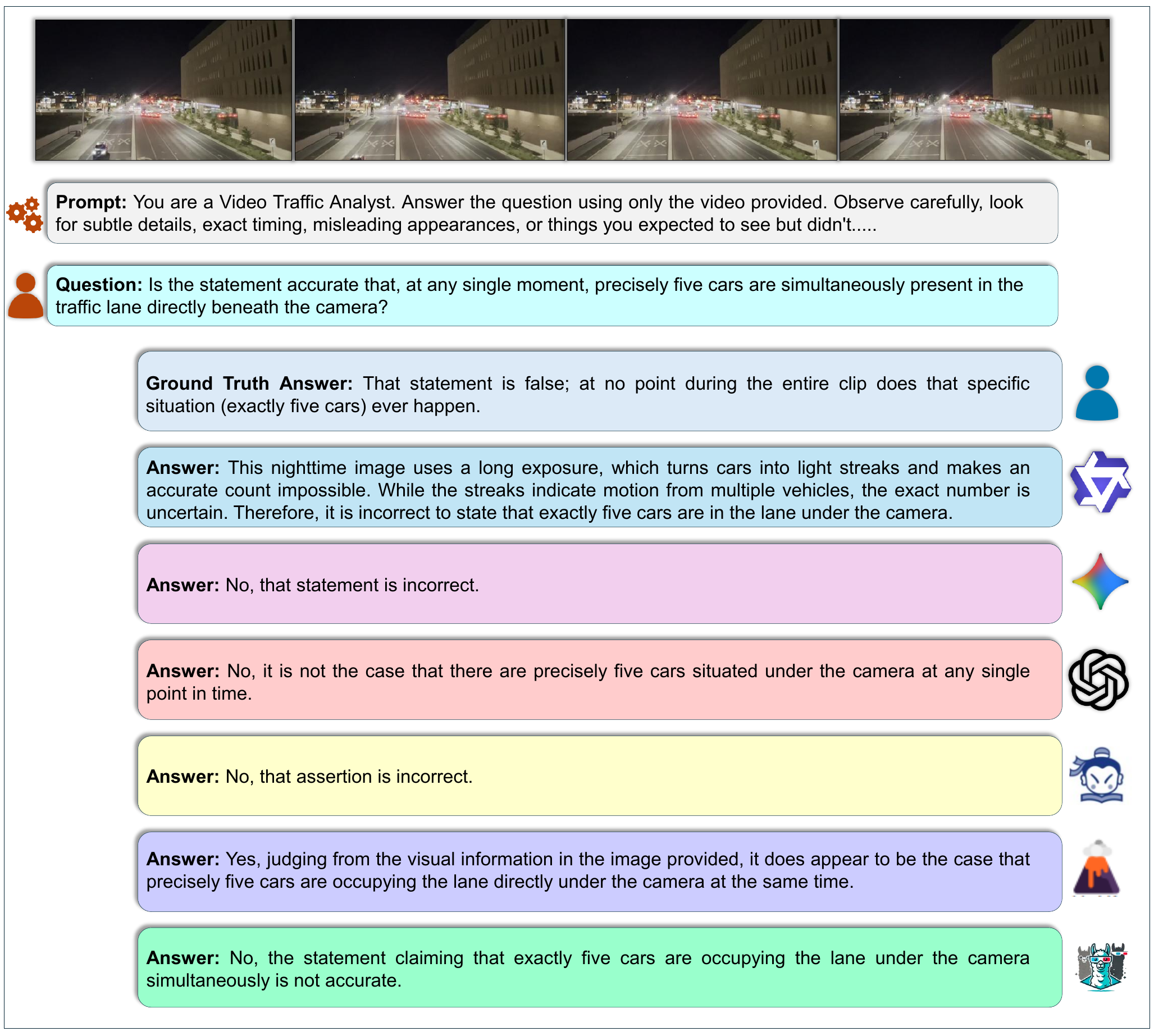}
\caption{Example of a \textit{counterfactual inference} question evaluating a model’s ability to reject a false premise. While most models correctly classify the statement as false, \textit{LLaVA-NeXT} incorrectly confirms the non-existent event, revealing a failure in precise factual verification.}
\label{fig:Counters_Factusla} 
\end{figure*}

Addressing these issues will reduce hallucination, improve the fidelity of counterfactual and event-reasoning evaluations, and close the gap between \textit{``reasoning''} and \textit{``seeing''} that we observe across current SOTA models.
Examples for \textit{RR} are shown in Figure \ref{fig:evening_Clips}, \textit{ER} in Figure \ref{fig:morning_clips}, \textit{RR} in Figure \ref{fig:attribution_lol}, \textit{ER} in Figure \ref{fig:basics}, and \textit{CF} in Figure \ref{fig:Counters_Factusla}, respectively.

\begin{figure*}[t!]
\centering
\includegraphics[width=0.90\textwidth]{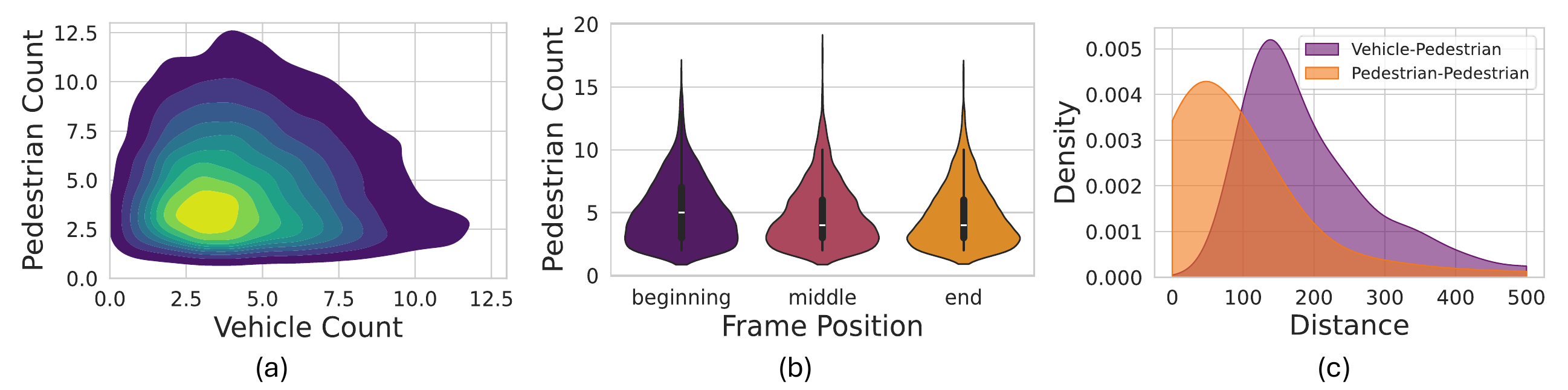}
\caption{Analysis of multi-agent dynamics: (a) most frequent multi-agent scenario observed in the dataset, (b) temporal stability of pedestrian activity across video segments, and (c) proximity density plots highlighting frequent close interactions between vehicle–pedestrian and pedestrian–pedestrian pairs.}
\label{fig:Traffic_density} 
\end{figure*}

\subsection{Multi-Agent Behavioral Analysis}
\label{sec:multi_agent_analysis}

Traffic intersections represent highly dynamic environments where diverse agents-vehicles, pedestrians, and micro-mobility users-share limited space and interact under varying contextual constraints. We hypothesize that such intersections naturally promote frequent, observable multi-agent interactions that make them ideal for studying spatio-temporal reasoning. Our empirical analysis of the \textit{UDVideoQA} dataset supports this hypothesis through three key observations: \textbf{\textit{(1)}} strong agent co-occurrence, \textbf{\textit{(2)}} temporal consistency of activity, and \textbf{\textit{(3)}} measurable proximity-based interaction potential.

\paragraph{\textit{Agent Co-occurrence}}
A fundamental prerequisite for studying interactions is the concurrent presence of multiple agents within the same frame. Figure~\ref{fig:Traffic_density}(a) shows a density heatmap of vehicle and pedestrian counts across all annotated frames. The highest-density region (brightest yellow) lies not near the origin, representing empty or single-agent scenes, but around \textit{$2$-$5$ vehicles} and \textit{$2$–$4$ pedestrians}. This distribution confirms that the majority of footage captures scenes with concurrent multi-agent activity rather than sparse or unoccupied intersections. Such consistent co-occurrence validates the choice of intersections as natural laboratories for multi-agent reasoning, where spatial dependencies and interaction opportunities emerge continuously.

\paragraph{\textit{Temporal Consistency of Activity}}
To verify that these interactions are temporally stable rather than transient, we analyzed pedestrian counts across three temporal segments of each $10$s clip \textit{beginning}, \textit{middle}, and \textit{end}. The violin plots in Figure~\ref{fig:Traffic_density}(b) reveal near-identical distributions across these intervals: the median counts (white dots), interquartile ranges (black bars), and overall distribution shapes remain remarkably consistent. This indicates that the level of pedestrian activity and, by extension, the overall scene dynamics remain stable throughout the clips. The dataset, therefore, provides a balanced temporal sampling of urban traffic behavior rather than isolated or event-biased moments.

\paragraph{\textit{Quantifying Proximity and Interaction Potential}}
Beyond co-occurrence, we measure physical proximity to estimate interaction potential. Figure~\ref{fig:Traffic_density} (c) shows pedestrian–pedestrian distances peaking near $50$–$75$ units, while vehicle–pedestrian distances peak around $150$ units, with a significant tail of low-distance events. These \textit{high-interaction zones} confirm that intersections force multi-agent co-location via traffic constraints like lights and crosswalks. This proximity generates rich behaviors like yielding and conflict, demonstrating that intersections are \textit{optimal real-world settings} for the complex, causally entangled reasoning \textit{UDVideoQA} aims to benchmark.

\begin{figure}[t]
\centering
\includegraphics[width=0.9\linewidth]{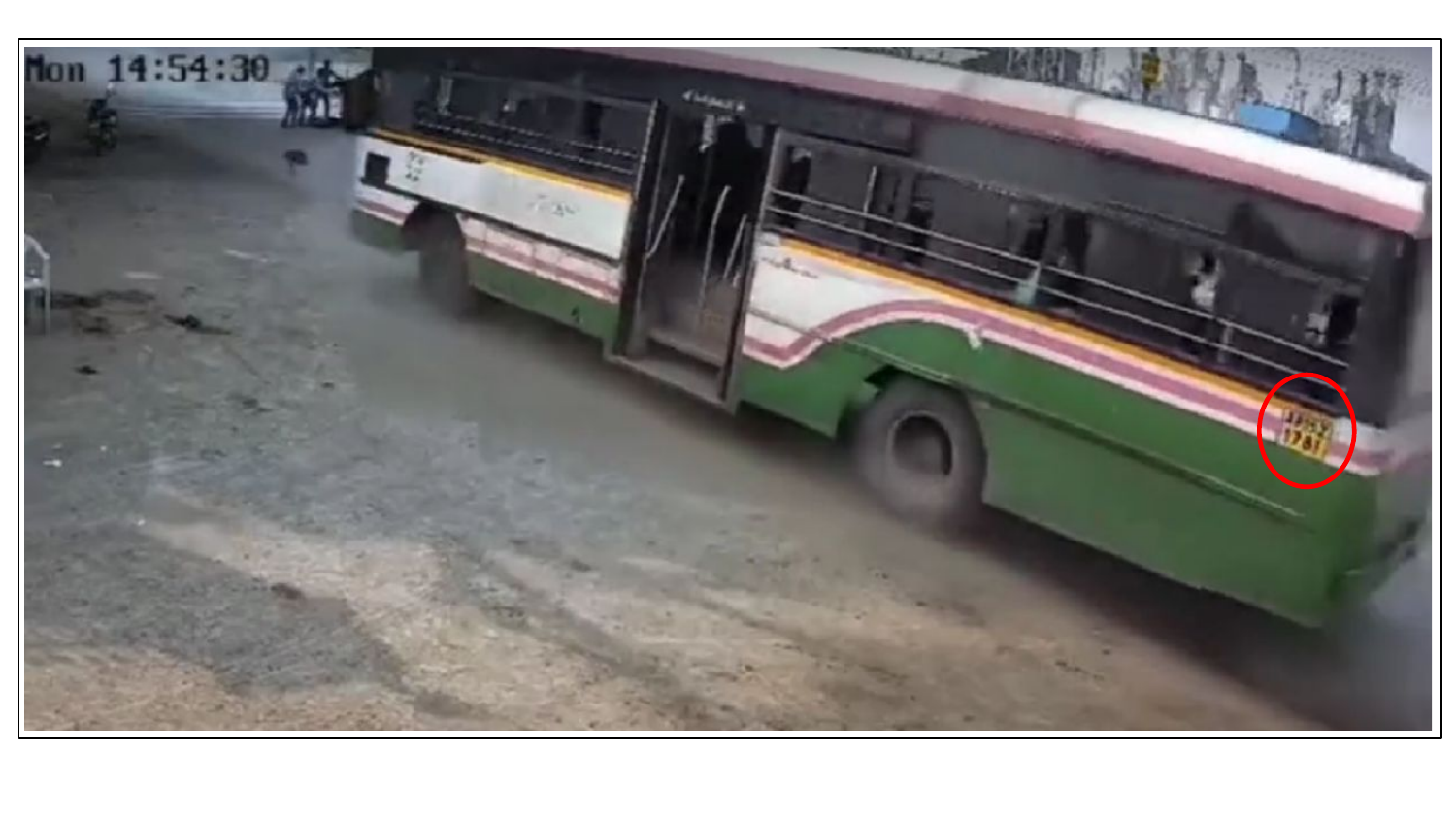}
\caption{Example frame illustrating the model’s use of visual evidence, specifically the ``AP” license plate prefix-to support its prediction in the \textit{RoadSocial} dataset.} 
\label{fig:road_social} 
\end{figure}

\subsection{Generalization Tests on available Datasets}
\label{sec:generalization}

To evaluate cross-domain transferability, we tested the fine-tuned \textit{Qwen 2.5-VL-7B} model on $500$ \textit{QA} pairs each from the validation splits of \textit{RoadSocial}~\cite{RoadSocial2025} and \textit{SUTDTrafficQA}~\cite{Xu2021SUTDTrafficQA}. These datasets were chosen for their structural similarity to \textit{UDVideoQA} in terms of question format and reasoning taxonomy, allowing for a controlled assessment of generalization across different video sources and annotation protocols.
The evaluation produced a few noteworthy observations (see Table~\ref{tab:road_socail_generlization} and Table~\ref {table:SUTD_generlization}). On the \textit{RoadSocial} benchmark, which includes deliberately adversarial questions targeting perception robustness, the model demonstrated strong visual grounding even when its prediction diverged from the labeled ground truth. For example:

\begin{table}[t]
\resizebox{0.95\linewidth}{!}{%
\begin{tabular}{@{}c|cccc|c@{}}
\toprule
                              & \multicolumn{4}{c|}{\textbf{QA-Group}} &               \\ \cmidrule(lr){2-5}
\multirow{-2}{*}{\textbf{\begin{tabular}[c]{@{}c@{}}Training \\ Configuration\end{tabular}}} &
  \textbf{Factual} &
  \textbf{Complex} &
  \textbf{Imaginative} &
  \textbf{Adversarial} &
  \multirow{-2}{*}{\textbf{\begin{tabular}[c]{@{}c@{}}Overall \\ Score\end{tabular}}} \\ \midrule
\rowcolor[HTML]{FFFFC7} 
\textbf{Fine Tuned Model}     & 68.2     & 41.8     & 74.4     & 55    & \textbf{60} \\
\textbf{Baseline} & 66.2       & 3      & 83.7     & 58.1   & \textbf{52.75} \\ \bottomrule
\end{tabular}%
}
\caption{Evaluation of the \textit{UDVideoQA} fine-tuned model’s generalization performance on the \textit{RoadSocial} dataset across four \textit{QA} categories - \textit{factual}, \textit{complex}, \textit{imaginative}, and \textit{adversarial}.}
\label{tab:road_socail_generlization}
\end{table}

\begin{table}[t]
\centering
\resizebox{0.75\linewidth}{!}{%
\begin{tabular}{@{}c|cccc|c@{}}
\toprule
                          & \multicolumn{4}{c|}{\textbf{QA-Group}} &                \\ \cmidrule(lr){2-5}
\multirow{-2}{*}{\textbf{\begin{tabular}[c]{@{}c@{}}Training \\ Configuration\end{tabular}}} &
  \textbf{BU} &
  \textbf{Att} &
  \textbf{CI} &
  \textbf{ER} &
  \multirow{-2}{*}{\textbf{\begin{tabular}[c]{@{}c@{}}Overall \\ Score\end{tabular}}} \\ \midrule
\rowcolor[HTML]{FFFFC7} 
\textbf{Fine Tuned Model} & 54.5     & 73.3     & 75       & 48    & \textbf{62.7}  \\
\textbf{Baseline}         & 48.9     & 69.6     & 73.4     & 44    & \textbf{59.05} \\ \bottomrule
\end{tabular}%
}
\caption{Evaluation of the \textit{UDVideoQA} fine-tuned model’s generalization performance on the \textit{SUTDTraffic} dataset across four \textit{QA} categories \textit{factual}, \textit{complex}, \textit{imaginative}, and \textit{adversarial}.}

\label{table:SUTD_generlization}
\end{table}

\begin{itemize}
\item \textbf{Question:} \textit{“In which Indian state did this road event likely occur?”}
\item \textbf{\textit{Ground Truth:}} \textit{“The incident likely occurred in Telangana, India.”}
\item \textbf{\textit{Model Prediction:}} \textit{“The bus in the video has a license plate that starts with ‘AP,’ which stands for Andhra Pradesh… therefore, it is likely that this road event occurred in Andhra Pradesh.”}
\end{itemize}

While the prediction (\textit{“Andhra Pradesh”}) does not match the ground-truth label (\textit{“Telangana”}), visual inspection confirmed that the bus’s license plate indeed displayed the prefix \textit{“AP”}. This outcome highlights the model’s capacity for \textit{visual evidence-based reasoning}, deriving logically coherent answers grounded in observed cues. Figure~\ref{fig:road_social} illustrates the corresponding frame used in this reasoning process.
Beyond individual examples, the fine-tuned \textit{Qwen} model achieved consistent quantitative improvements across both benchmarks. Under the weighted semantic-scoring framework introduced in Section~\ref{sec:eval_method}, it recorded a \textit{$7.3$\%} gain on \textit{RoadSocial} and a \textit{$3.61$\%} gain on \textit{SUTDTrafficQA} over its baseline. These improvements demonstrate that exposure to high-density, multi-agent intersection data in \textit{UDVideoQA} enhances a model’s ability to generalize causal and perceptual reasoning across unseen traffic domains. In summary, fine-tuning on \textit{UDVideoQA} not only boosts in-domain accuracy but also strengthens cross-domain robustness, particularly in tasks requiring grounded visual interpretation and resistance to hallucination.





\clearpage
\clearpage



\end{document}